\documentclass[sn-apa]{sn-jnl}

\usepackage{graphicx}%
\usepackage{multirow}%
\usepackage{amsmath,amssymb,amsfonts}%
\usepackage{amsthm}%
\usepackage{mathrsfs}%
\usepackage[title]{appendix}%
\usepackage{xcolor}%
\usepackage{textcomp}%
\usepackage{manyfoot}%
\usepackage{booktabs}%
\usepackage{algorithm}%
\usepackage{algorithmicx}%
\usepackage{algpseudocode}%
\usepackage{listings}%

\usepackage{pdflscape}
\usepackage{afterpage}
\usepackage{tabularx}
\usepackage{xltabular}
\usepackage{longtable}
\usepackage{hyperref}

\theoremstyle{thmstyleone}%
\theoremstyle{thmstyletwo}%

\theoremstyle{thmstylethree}%

\begin{document}

\title[PoTeC]{PoTeC: A German Naturalistic Eye-tracking-while-reading Corpus} 


\author*[1]{\fnm{Deborah N.} \sur{Jakobi}}\email{jakobi@cl.uzh.ch}

\author[2]{\fnm{Thomas} \sur{Kern}}

\author[2]{\fnm{David R.} \sur{Reich}}\email{david.reich@uni-potsdam.de}

\author[1]{\fnm{Patrick} \sur{Haller}}\email{haller@cl.uzh.ch}

\author*[1,2]{\fnm{Lena A.} \sur{Jäger}}\email{jaeger@cl.uzh.ch}

\affil[1]{\orgdiv{Department of Computational Linguistics}, \orgname{University of Zurich}, \orgaddress{\street{Andreasstrasse 15}, \city{Zurich}, \postcode{8050}, \country{Switzerland}}}

\affil[2]{\orgdiv{Department of Computer Science}, \orgname{University of Potsdam}, \orgaddress{\street{An der Bahn 2}, \city{Potsdam}, \postcode{14476}, \country{Germany}}}

\abstract{The \textbf{Po}tsdam \textbf{Te}xtbook \textbf{C}orpus (PoTeC) is a naturalistic eye-tracking-while-reading corpus containing data from 75 participants reading 12 scientific texts.  PoTeC is the first naturalistic eye-tracking-while-reading corpus that contains eye-movements from domain-experts as well as novices in a within-participant manipulation: It is based on a 2$\times$2$\times$2 fully-crossed factorial design which includes the participants' level of study and the participants' discipline of study as between-subject factors and the text domain as a within-subject factor. The participants' reading comprehension was assessed by a series of text comprehension questions and their domain knowledge was tested by text-independent background questions for each of the texts. The materials are annotated for a variety of linguistic features at different levels. We envision PoTeC to be used for a wide range of studies including but not limited to analyses of expert and non-expert reading strategies. The corpus and all the accompanying data \textit{at all stages of the preprocessing pipeline} and \textit{all} code used to preprocess the data are made available via GitHub: \href{https://github.com/DiLi-Lab/PoTeC}{https://github.com/DiLi-Lab/PoTeC}.}




\maketitle

\section{Introduction}
\label{sec:intro}
Eye-tracking-while-reading data is widely used in different areas of research including linguistics, cognitive psychology and computer science. In pyscholinguistic reading research it is considered a gold standard dependent variable for investigating the cognitive processes involved in language comprehension \citep{Rayner1998, raynerCarroll2018}. 
An overwhelming majority of eye-tracking studies in psycholinguistic research are based on \textit{controlled} (also referred to as \textit{planned}) \textit{experiments} with hand-crafted stimuli that constitute minimal pairs. However, as recent research has shown, it is crucial to (also) study language processing in ecologically valid settings using naturalistic, real-world stimuli \citep{demberg2019cognitive, DembergKeller2008, NASTASE2020117254}. 

 The stimulus materials of planned experiments, comprising minimal pairs (and relevant filler items), are specifically designed to test a small number of predefined, typically theoretically motivated, hypotheses in a controlled linguistic environment, and thus cover only a small range of linguistic constructions. For example, the hypothesis that one syntactic construction is more difficult to process than another one can be investigated by designing minimal pair stimuli that contain the experimental manipulation and certain words or constituents, where the difference in processing difficulty is expected. This experimental approach is crucial to make statements about specific phenomena which are often highly infrequent in naturally occurring text. Studying infrequent, complex phenomena can be theoretically relevant as it often allows for teasing apart competing accounts that in many cases make very similar predictions for frequently occurring, simple constructions. Only such a controlled experimental approach allows for drawing conclusions about the causality of observed effects --- although one needs to keep in mind that linguistic manipulations are almost always only quasi-experimental in nature. 
 
Conversely, naturalistic reading corpora, where participants read naturally occurring text (e.g., newspaper articles) rather than  minimal pair stimuli, cover a wider range of linguistic constructions. This allows for studying a broad and representative range of phenomena without the need to design new experiments and collect new data. The phenomena can not only be studied within a single dataset but also within-subject, which is not possible in planned experiments where typically one study is limited to a single phenomenon. The coverage of existing theories can be evaluated on these datasets and the theories can be modified accordingly. Additionally, exploratory data analyses can lead to new theories or inspire new research questions \citep{hamilton2020REvolution}.
While planned experiments using minimal pair stimuli are essential for closely examining and comparing different theories and testing hypotheses, naturalistic corpora permit observational studies in ecologically valid settings, enhancing generalizability, and inspiring new theories based on exploratory data analyses.

Besides the broad categorization into planned experiments with minimal pair stimuli and naturalistic reading corpora, there are more fine-grained differences within naturalistic eye movement corpora. One crucial dimension for characterizing naturalistic eye movement data revolves around the nature of the stimuli which spans from entirely naturalistic to partially constructed, where, for instance, specific target words within the stimulus sentences are hand-picked based on their lexical frequency. Seminal work on eye-tracking-while-reading corpora, such as the work by~\citet{PSC}, has adopted this approach by using partially constructed stimuli which enables the inclusion of a broad range of linguistic phenomena that are highly infrequent in natural text (but relevant test cases for linguistic theories) while still utilizing sentences that maintain a higher degree of naturalness than typical planned experiments.
Other naturalistic reading corpora leverage naturally occurring text as stimuli which are not specifically designed for the experiment but simply being reused from an existing source. Often, these texts span multiple sentences which allows for analyses that go beyond the sentence-level. In the most stringent cases, alterations of the texts are entirely avoided. However, in many cases, texts need to be edited to, for example, compensate for missing context or exclude figures and tables. Leveraging texts that closely resemble or are identical to real-world reading material enables to study cognitive processes involved in everyday reading at both sentence and text-level, while, as a natural consequence, specific phenomenon (e.g., long-distance dependencies) might occur rarely or never within an entire dataset.

In addition to psycholinguistic hypothesis testing, eye-tracking-while-reading data has become increasingly relevant for other areas of research such as Natural Language Processing (NLP). NLP research has been leveraging eye movements in reading for a wide range of tasks such as sentiment analysis \citep{long-etal-2017-sentiment, mishra-2017-sentiment}, named entity recognition \citep{hollenstein-zhang-2019-entity}, part-of-speech tagging \citep{barrett-etal-2016-pos-tag} or generating image captions \citep{takmaz-etal-2020-img-caption} among other tasks. 
In recent NLP research, eye-tracking-while-reading data has been used to analyze computational language models, for example by investigating their cognitive plausibility \citep{beinborn2023cognitive, keller-2010-cogn-plausibility}. \citet{sood2020interpreting} analyzed attention weights learned by transformer language models and compared those to human attention implicitly encoded in human gaze data. More than that, cognitive signals can be used to improve language models \citep{hollenstein-etal-2020-practices}. Naturalistic reading data can be used to cognitively enhance and augment language models with human gaze data \citep{yang2023plm, deng-etal-2023-pre, prasse2023spygan-review}. 
For all of these research objectives, it is crucial to have large and diverse amounts of cognitive data available. Many NLP tasks rely on long text passages and thus require cognitive data for entire paragraphs of natural texts rather than just single sentences. 

In this work, we present the Potsdam Textbook Corpus (PoTeC), a naturalistic eye-tracking-while-reading corpus containing eye-tracking data of German native speakers reading German textbook passages from two different domains. PoTeC is the first corpus to include the level of expertise of each participant as a within-subject variable and thus allows for analyzing reading strategies used by experts and non-experts. The naturalistic experimental setting of PoTeC encourages various kinds of analyses which are not restricted to test one specific hypothesis and thus has valuable properties that complement other existing corpora.

\paragraph{New standard for data publication}
In addition to the publication of the eye-tracking data, our aim is to foster transparency and re-usability of the data and facilitate leveraging PoTeC for a variety of different use cases (e.g., psycholinguistic research, NLP research, development of new preprocessing algorithms, eye-movement-based biometrics, and many more). In order to achieve this goal, we are presenting a new standard to publish eye-tracking datasets. Following the FAIR principles (\textbf{F}indable, \textbf{A}ccessible, \textbf{I}nteroperable, \textbf{R}eusable), our data is shared in the following way:
\begin{itemize}
    \item The data is shared via two channels suitable for the respective data type (large data files and code; see Section \ref{sec:data-access}).
    \item The eye-tracking data is made available \textit{at all stages} (e.g., raw sample data, reading measures data, etc.; see Table \ref{tab:preprocessing}). 
    \item All the code that was used for preprocessing and analyzing the data is made publicly available in reproducible formats.
    \item All data and code is accompanied by extensive documentation that not only makes the process transparent and reproducible, but also makes it maximally easy for  users with different backgrounds to reuse the scripts for their own purposes.
\end{itemize} 

In addition to following the general FAIR principles, there are two very specific implementations of these principles: As a unique feature of PoTeC, the eye-tracking fixation data is made available in both the original version as it was collected and as a version where the horizontal calibration drift was manually corrected post-hoc. The availability of this data not only makes the process transparent but also allows to study and compare the original data with the manually corrected data which enables the development of new tools, including machine-learning-based methods, to automatically detect and/or correct horizontal drift. The data is furthermore integrated into the open-source Python package \texttt{pymovements}\footnote{\url{https://pymovements.readthedocs.io/en/stable/}}. The package enables to build easy machine learning and psycholinguistic pipelines for the processing of eye movement data and can be used in Python and R, which increases re-usability \citep{pymovements}.

\section{Related Work}
 
\subsection{Naturalistic text passage corpora for German}
Reading completely naturalistic text passages consisting of multiple sentences shown at the same time as in PoTeC exist for several languages, however, only very few for German (see Table \ref{tab:datasets-full-texts-de}). A very recently created corpus is the Multilingual Eye-Movements Corpus (MECO-L1) which is a 13-language reading corpus. The German subset contains data from 45 participants reading 12 encyclopedic texts on various topics specifically chosen not to require an academic background \citep{Siegelman2022}. The PopSci Corpus includes 17 participants reading 16 popular science texts from different sources in German \citep{popsci}. WebQAmGaze is another natural reading dataset which includes a German subset \citep{ribeiro2023webqamgaze}. However, the data was recorded using a webcam which results in substantially lower data quality compared to a high-precision eye tracker typically used in eye-tracking-while-reading experiments, and is therefore limited to a small subset of use cases compared to other reading corpora.

\begin{table}[h]
\caption{\label{tab:datasets-full-texts-de}Naturalistic eye-tracking-while-reading corpora for German text passages}
\begin{tabularx}{\textwidth}{>{\raggedright\hsize=.35\hsize}X>{\raggedright\hsize=.24\hsize}X>{\raggedright\hsize=.41\hsize}X}
\toprule
\textbf{Stimuli} & \textbf{Participants} & \textbf{Additional characteristics}   \\\midrule

\multicolumn{3}{l}{\textbf{PoTeC (present study)}} \vspace{0.08cm}\\
12 texts from undergraduate textbooks on physics or biology ($\scriptstyle\# \textstyle w$: 1,896, $\overline{\scriptstyle\# \textstyle w}$: 158) & 75 graduate or undergraduate biology or physics students (native speakers) \newline ($\bar{a}$: $24.2$ ($4.2$)) 
& The corpus is designed to specifically enable comparison of expert and non-expert reading. \newline \textbf{Eye-tracker}: EyeLink 1000 \newline \textbf{Sampl. Freq.}: 1,000\,Hz
\\\cmidrule{1-3}

\multicolumn{3}{l}{\textbf{MECO-L1} \citep{Siegelman2022}}\vspace{0.08cm} \\
12 Wikipedia-style texts on various topics not requiring an academic background \newline ($\scriptstyle\# \textstyle w$: 2,028, $\overline{\scriptstyle\# \textstyle w}$: 169) & 45 native speakers \newline ($\bar{a}$: 23.76) 
& Data is available for 12 other languages (nl, en, el, de, he, it, ru, es, tr, ko, no, et and fi) for a total of 535 participants reading text in their native language. \newline \textbf{Eye-tracker}: EyeLink Portable Duo, 1000 or 1000+ \newline \textbf{Sampl. Freq.}: 1,000\,Hz\\\cmidrule{1-3}

\multicolumn{3}{l}{\textbf{PopSci Corpus} \citep{popsci}}\vspace{0.08cm} \\
 16 popular science texts on natural and applied science ($\scriptstyle\# \textstyle w$: 20,000, $\overline{\scriptstyle\# \textstyle w}$: 1,250) & 17 participants & \textbf{Eye-tracker}: EyeLink 1000 \newline \textbf{Sampl. Freq.}: 1,000\,Hz \\\cmidrule{1-3}

\multicolumn{3}{l}{\textbf{WebQAmGaze} \citep{ribeiro2023webqamgaze}}\vspace{0.08cm}\\
38 Wikipedia-style texts: 2 long texts ($\scriptstyle\# \textstyle w$: 370, $\overline{\scriptstyle\# \textstyle w}$: 185) and 36 shorter texts ($\scriptstyle\# \textstyle w$: 3,010, $\overline{\scriptstyle\# \textstyle w}$: 83.6) with questions and annotated answer spans  & 19 native speakers &  The data can be used to study reading during question answering, it was collected using a Webcam, and is available in two other languages (en, es). \newline \textbf{Eye-tracker}: Webcam \newline \textbf{Sampl. Freq. (mean)}: 24.93\,Hz\\
\botrule
\end{tabularx}
\footnotetext{Abbr.: $\scriptstyle\# \textstyle w$, $\overline{\scriptstyle\# \textstyle w}$ = (mean) number of words of stimuli; $\bar{a}$~=~mean age of participants (SD)}
\end{table}

\subsection{Naturalistic text passage corpora for languages other than German}
The vast majority of eye-tracking-while-reading datasets exists for English stimulus texts (see Tables \ref{tab:datasets-full-texts-en-1} and \ref{tab:datasets-full-texts-en-2}). The Provo Corpus contains data collected from 84 participants reading 55 short texts from various sources and includes human predictability norms \citep{Luke2017provo}. \citet{yaneve2016asd} collected a dataset containing data from 108 participants of which some are diagnosed with Autism Spectrum Disorder (ASD) and some are part of a control group, and each participant is reading a subset of in total 27 text. For the GazeBase reading task, a total of 322 participants read up to 18 passages of a poem over a three year period \citep{Griffith2021gazebase}. \citet{malmaud-etal-2020-bridging} collected eye-tracking data for the OneStopQA dataset which is designed to study reading comprehension. They collected data from 296 participants each reading 10 different articles out of 30 articles with the questions presented either before or after reading each text. A corpus constructed to study parafoveal viewing by placing target words inside or outside of the parafoveal view contains 48 participants reading 40 text passages \citep{parker2017passage-reading}.

There exist datasets that were specifically designed to automatically assess reader or text properties or use the data for NLP purposes. SB-SAT is a 95-participant dataset for which participants were asked to judge the subjective difficulty of 4 passages taken from practice tests for the Scholastic Assessment Test (SAT) \citep{ahn2020SAT}. The MAQ-RC corpus contains 23 participants reading 32 movie plots and answering comprehension questions which were also answered by computational models and thus allows for comparing human and machine reading comprehension \citep{sood2020interpreting}. The CFILT coreference dataset contains 14 participants (including 2 linguistic experts) reading 22 texts and at same time the participants were asked to annotate coreferences in the texts \citep{cheri-etal-2016-coreference}. A similar dataset is the CFILT scanpaths dataset that contains data from 16 participants (including 3 linguistic experts) reading 32 paragraphs from (simple) Wikipedia articles. The participants annotated the texts for the effort to read them such that scanpath complexity can be studied, as more effort to read presumably results is more complex scanpaths \citep{mishra2017scanpath}. The CFILT text quality dataset includes data from 20 fluent English speakers reading 30 texts from different sources which were asked to rate the text quality given three properties (organization, coherence and cohesion) \citep{mathias-etal-2018-eyes}. The CFILT essay grading dataset includes 8 fluent English speakers reading and grading 48 essays while their eye movements were being tracked \citep{mathias-etal-2020-happy}.

\begin{table}[h!]
\caption{\label{tab:datasets-full-texts-en-1}Naturalistic eye-tracking-while-reading corpora for text passages in \textbf{English} (part I)}
\begin{footnotesize}
\begin{tabularx}{\textwidth}{>{\raggedright\hsize=.35\hsize}X>{\raggedright\hsize=.23\hsize}X>{\raggedright\hsize=.42\hsize}X}
\toprule
 \textbf{Stimuli} & \textbf{Participants} & \textbf{Additional characteristics} \\\midrule

\multicolumn{3}{l}{\textbf{SB-SAT} \citep{ahn2020SAT}}\vspace{0.08cm}\\
 4 SAT passages taken from practice tests for reading comprehension & 95 undergraduate students & Subjective difficulty ratings \newline \textbf{Eye-tracker}: EyeLink 1000 \newline \textbf{Sampl. Freq.}: 1,000\,Hz \\\cmidrule{1-3}

\multicolumn{3}{l}{\textbf{Passage reading} \citep{parker2017passage-reading}}\vspace{0.08cm}\\
 40 text passages constructed around target words which will either appear in parafovea or not   & 48 native speakers \newline ($\bar{a}$: 26.48 (14.83)) & Constructed specifically to study parafoveal viewing. \newline \textbf{Eye-tracker}: EyeLink 1000 \newline \textbf{Sampl. Freq.}: 1,000\,Hz \\\cmidrule{1-3}

\multicolumn{3}{l}{\textbf{Provo Corpus} \citep{Luke2017provo}}\vspace{0.08cm} \\
 55 short texts from various different sources ($\scriptstyle\# \textstyle w$: 2,750, $\overline{\scriptstyle\# \textstyle w}$: 50) & 84 native speakers &  Includes human predictability norms \newline \textbf{Eye-tracker}: EyeLink 1000+ \newline \textbf{Sampl. Freq.}: 1,000\,Hz \\\cmidrule{1-3}

\multicolumn{3}{l}{\textbf{MQA-RC} \citep{sood2020interpreting}}\vspace{0.08cm} \\
 32 movie plots with comprehension questions including results from computational models answering the questions (200–250 words) &  23 native speakers & The participants had to answer comprehension questions in different settings (e.g. with/without plot). \newline \textbf{Eye-tracker}: Tobii \newline \textbf{Sampl. Freq.}: 600\,Hz \\\cmidrule{1-3}

\multicolumn{3}{l}{\textbf{ASD Data} \citep{yaneve2016asd}}\vspace{0.08cm}\\
 27 texts from various sources ($\scriptstyle\# \textstyle w$: 4,212, $\overline{\scriptstyle\# \textstyle w}$: 156 (49.94))  & 108 native speakers, 56 diagnosed with ASD and a control group ($\bar{a}$: 33.73 (8.36))& Groups of participants read disjoint subsets of the texts. \newline \textbf{Eye-tracker}: Gazepoint GP3 \newline \textbf{Sampl. Freq.}: 60\,Hz \\\cmidrule{1-3}

\multicolumn{3}{l}{\textbf{GazeBase - Reading Task} \citep{Griffith2021gazebase}}\vspace{0.08cm} \\
 18 passages from the same poem with a maximum of 60s to read two texts&  322 participants, not all of them read all passages &  Reading was spread over 3 years \newline \textbf{Eye-tracker}: EyeLink 1000 \newline \textbf{Sampl. Freq.}: 1,000\,Hz \\

\botrule
\end{tabularx}
\footnotetext{Abbr.: $\scriptstyle\# \textstyle w$, $\overline{\scriptstyle\# \textstyle w}$ = (mean) number of words of stimuli (SD); $\bar{a}$~=~mean age of participants (SD)}
\end{footnotesize}
\end{table}

\begin{table}[h!]
\caption{\label{tab:datasets-full-texts-en-2}Naturalistic eye-tracking-while-reading corpora for text passages in \textbf{English} (part II)}
\begin{footnotesize}
\begin{tabularx}{\textwidth}{>{\raggedright\hsize=.30\hsize}X>{\raggedright\hsize=.28\hsize}X>{\raggedright\hsize=.42\hsize}X}
\toprule
 \textbf{Stimuli} & \textbf{Participants} & \textbf{Additional characteristics} \\\midrule

\multicolumn{3}{l}{\textbf{OneStopQA Eye-Tracking} \citep{malmaud-etal-2020-bridging}}\vspace{0.08cm} \\
 30 Guardian articles, each in an advanced difficulty version ($\scriptstyle\# \textstyle w$: 3,858, $\overline{\scriptstyle\# \textstyle w}$: 128.6) and an elementary version ($\scriptstyle\# \textstyle w$: 3,369, $\overline{\scriptstyle\# \textstyle w}$: 112.3) &  269 participants, each participant read 10 articles & One question per text was shown prior to reading or after reading the text \newline \textbf{Eye-tracker}: EyeLink 1000+ \newline \textbf{Sampl. Freq.}: 1,000\,Hz \\\cmidrule{1-3}

\multicolumn{3}{l}{\textbf{CFILT - Coreference} \citep{cheri-etal-2016-coreference}}\vspace{0.08cm} \\
 22 texts (less than 10 sentences each text) & 14 non-native participants: \newline 2 are expert linguists (age: 47–50), 12 (post-)graduates (age: 20–30) & Participants annotated  coreferences in the text. \newline \textbf{Eye-tracker}: EyeLink 1000+ \newline \textbf{Sampl. Freq.}: 500\,Hz \\\cmidrule{1-3}
 
 \multicolumn{3}{l}{\textbf{CFILT - Scanpath} \citep{mishra2017scanpath}}\vspace{0.08cm} \\
  32 paragraphs from (simple) Wikipedia (50–200 words each) & 16 non-native participants: 3 are expert linguists (age: 47–50), 13 (post-)graduates (age: 20–30)& Participants annotated texts for the effort to read them. \newline \textbf{Eye-tracker}: EyeLink 1000+ \newline \textbf{Sampl. Freq.}: --\\\cmidrule{1-3}

   \multicolumn{3}{l}{\textbf{CFILT - Essay Grading} \citep{mathias-etal-2020-happy}}\vspace{0.08cm} \\
   48 essays (max. words per essay: 250)& 8 fluent English speakers ($\bar{a}$: 25) & Participants graded the essays after reading. \newline \textbf{Eye-tracker}: EyeLink 1000 \newline \textbf{Sampl. Freq.}: 500\,Hz\\\cmidrule{1-3}

    \multicolumn{3}{l}{\textbf{CFILT - Text Quality} \citep{mathias-etal-2018-eyes}}\vspace{0.08cm} \\
   30 texts from different sources (ca. 200 words in each text) & 20 fluent English speakers (age: 20–25) & Participants annotated the text quality based on three given properties (organization, coherence and cohesion). \newline \textbf{Eye-tracker}: EyeLink 1000 \newline \textbf{Sampl. Freq.}: 500\,Hz\\\cmidrule{1-3}
   
   \multicolumn{3}{l}{\textbf{MECO-L2} \citep{kuperman-2023-mecol2}}\vspace{0.08cm} \\
12 encyclopedic texts originally designed for English language testing ($\scriptstyle\# \textstyle w$: 1,653, $\overline{\scriptstyle\# \textstyle w}$: 137.8 (25.3)) & 543 non-native participants ($\bar{a}$: 23.4), \newline L1: nl, en, el, de, he, it, ru, es, tr, no, et, fi (ISO-639-1)  & \textbf{Eye-tracker}: EyeLink Portable Duo, 1000 or 1000+ \newline \textbf{Sampl. Freq.}: 1,000\,Hz \\
\botrule
\end{tabularx}
\footnotetext{Abbr.: $\scriptstyle\# \textstyle w$, $\overline{\scriptstyle\# \textstyle w}$ = (mean) number of words of stimuli (SD); $\bar{a}$~=~mean age of participants (SD)}
\end{footnotesize}
\end{table}

Probably the earliest eye-tracking-while-reading corpus for text passages in English and French is the Dundee corpus containing data from 10 English and 10 French native speakers reading newspaper extracts in their native language \citep{dundee, kennedy2013frequency}. MECO-L2 includes 543 non-native speakers of 12 different L1 backgrounds (nl, en, el, de, he, it, ru, es, tr, no, et and fi) reading 12 English texts which is the largest existing corpus of this type \citep{kuperman-2023-mecol2}. Another corpus containing L2 data is GECO which contains data from 14 English monolinguals reading an entire novel in English, and 19 bilinguals (Dutch: L1, English: L2) reading half of of the novel in Dutch and the other half in English \citep{Cop2016geco}. For the same stimulus text, data was also collected from 30 Chinese native speakers reading half of the novel in Chinese and the other half in English \citep[GECO-CN]{Sui2022gecocn}. To study gaze behaviour for summarization tasks, 50 participants' eye movements were tracked while reading 100 Chinese articles with the specific task to summarize the articles after reading \citep{yi2020gaze-sum}. For another corpus in Chinese, 29 participants were instructed to judge the relevance of a given document for a given query for a total of 60 document-query pairs \citep{li2018reading-attention}. The first corpus to study eye movements while reading text passages in Danish is CopCo \citep{hollenstein-etal-2022-copenhagen}. The dataset contains data from 22 native, 19 dyslexic and 10 Danish L2 speakers reading 20 speech manuscripts in Danish.
RastrOS contains eye movement data from 37 participants reading 50 paragraphs from different sources in Portugese and includes human predictability norms \citep{Leal2022rastros}. 
The Mental Simulation Corpus includes data from 102 participants reading 3 literary short stories in Dutch and were subsequently asked to answer questions that allow for studying their mental simulation during reading \citep{mak2019mentalsimulation}.

\begin{table}[h!]
\caption{\label{tab:datasets-full-texts}Naturalistic eye-tracking-while-reading corpora on text passages in different languages (except German and English), or on multilingual text passage corpora}
\begin{footnotesize}
\begin{tabularx}{\textwidth}{>{\raggedright\hsize=.01\hsize}X>{\raggedright\hsize=.31\hsize}X>{\raggedright\hsize=.31\hsize}X>{\raggedright\hsize=.37\hsize}X}
\toprule
\textbf\textbf{L} & \textbf{Stimuli} & \textbf{Participants} & \textbf{Additional characteristics} \\\midrule

\multicolumn{4}{l}{\textbf{GECO} \citep{Cop2016geco}} \vspace{0.08cm}\\
 en, nl & Complete novel which is easy to read in English and Dutch ($\scriptstyle\# \textstyle w$ en: 54,364, $\scriptstyle\# \textstyle w$ nl: 59,716) & 33 participants: 19 bilinguals (L1: nl, L2: en) ($\bar{a}$: 21.2 (2.2)) and 14 monolinguals (en) ($\bar{a}$: 21.8 (5.6))& Bilinguals read one half in Dutch, the other in English.  \newline \textbf{Eye-tracker}: EyeLink 1000 \newline \textbf{Sampl. Freq.}: 1,000\,Hz \\\cmidrule{1-4}
 
 \multicolumn{4}{l}{\textbf{GECO-CN} \citep{Sui2022gecocn}} \vspace{0.08cm}\\
 zh, en & Complete novel in Chinese and English ($\scriptstyle\# \textstyle w$ en: 54,364, $\scriptstyle\# \textstyle w$ zh: 59,403)& 30 bilinguals (L1: zh, L2: en) ($\bar{a}$: 25.3 (2.6))&  Participants read one half in Chinese, the other in English. \newline \textbf{Eye-tracker}: EyeLink 1000+ \newline \textbf{Sampl. Freq.}: 1,000\,Hz \\\cmidrule{1-4}

\multicolumn{4}{l}{\textbf{Gaze Behavior in Text Summarization} \citep{yi2020gaze-sum}} \vspace{0.08cm}\\
 zh & 100 articles from public news websites each belonging to one out of ten categories ($\overline{\scriptstyle\# \textstyle c}$: 502) & 50 participants ($\bar{a}$: 23.1 (1.1)) &  Participants were asked to summarize each text after reading. \newline \textbf{Eye-tracker}: Tobii EyeTracking 4C \newline \textbf{Sampl. Freq.}: 100\,Hz \\\cmidrule{1-4}
 
 \multicolumn{4}{l}{\textbf{CopCo} \citep{hollenstein-etal-2022-copenhagen}}\vspace{0.08cm}\\
 da & 20 speech manuscripts ($\scriptstyle\# \textstyle w$: 34,897, $\overline{\scriptstyle\# \textstyle w}$: 1,745) & 51 participants: 22 native speakers, 19 dislexic native speakers and 10 L2 speakers (age: 21–62)&  \textbf{Eye-tracker}: EyeLink 1000+\newline \textbf{Sampl. Freq.}: 1,000\,Hz  \\\cmidrule{1-4}

\multicolumn{4}{l}{\textbf{Reading Attention} \citep{li2018reading-attention}}\vspace{0.08cm} \\
 zh & 60 documents, each belongs to one out of 15 search queries & 29 native speakers (age: 17–28) \newline each participant read 15 documents (one for each query) & The participants were asked to judge the relevance of the document for the query. \newline \textbf{Eye-tracker}: Tobii X2-30 \newline \textbf{Sampl. Freq.}: --\,Hz \\\cmidrule{1-4}

\multicolumn{4}{l}{\textbf{Mental Simulation Corpus} \citep{mak2019mentalsimulation}} \vspace{0.08cm}\\
 nl &  3 literary short stories ($\scriptstyle\# \textstyle w$: 10,800, $\overline{\scriptstyle\# \textstyle w}$: 2600)  & 102 participants ($\bar{a}$: 23) & Participants answered questions that allow for studying their mental simulation during reading. \newline \textbf{Eye-tracker}: EyeLink 1000+ \newline \textbf{Sampl. Freq.}: 500\,Hz \\\cmidrule{1-4}

\multicolumn{4}{l}{\textbf{Dundee} \citep{dundee, kennedy2013frequency}}\vspace{0.08cm} \\
 en, fr & Extracts of newspaper articles in French and English & 20 participants, 10 each language &  Participants read the texts written in their native language. \newline \textbf{Eye-tracker}: Dr Bouis Oculometer Eyetracker \newline \textbf{Sampl. Freq.}: 1,000\,Hz \\\cmidrule{1-4}

\multicolumn{4}{l}{\textbf{RastrOS} \citep{Leal2022rastros, vieira2020rastro}}\vspace{0.08cm} \\
 pt & 50 paragraphs from journalistic, literary and popular science texts ($\overline{\scriptstyle\# \textstyle w}$: 2,494, $\overline{\scriptstyle\# \textstyle w}$: 49 (7.47))& 37 native speakers ($\bar{a}$: 22.2 (4.7)) & Includes human predictability norms. \newline \textbf{Eye-tracker}: EyeLink 1000 \newline \textbf{Sampl. Freq.}: 1,000\,Hz \\
 
\botrule
\end{tabularx}
\footnotetext{Abbr.: L = language of stimuli (ISO-639-1); $\scriptstyle\# \textstyle w$, $\overline{\scriptstyle\# \textstyle w}$ = (mean) number of words of stimuli (SD); $\overline{\scriptstyle\# \textstyle c}$~=~mean number of characters per stimulus text; $\bar{a}$~=~mean age of participants (SD)}
\end{footnotesize}
\end{table}

\subsection{Single-sentence corpora with partially constructed stimuli for different languages}
Naturalistic single-sentence-experiments have the advantage that the resulting data is easier to analyze as, for example, it does not necessarily involve multiple lines or even multiple pages per stimulus item. On the other hand, their use cases are limited to analyses of reading patterns at the sentence level. 
Many single-sentence naturalistic reading corpora leverage partially or entirely constructed sentences, however, the stimuli are often specifically constructed to contain a wide range of constructions. The Potsdam Sentence Corpus (PSC) is such a corpus which includes 33 young and 32 older adults reading 144 individual German sentences \citep{PSC, kliegl2004psc}. The sentences were constructed to cover a wide range of predefined linguistic constructions by selecting specific target words. There are more such corpora in different languages that follow the same or a similar procedure to create the stimuli based on a target word: The Russian Sentence Corpus (RSC) contains data from 96 participants reading 144 sentences \citep{Laurinavichyute2018RSC}. \cite{Zang2018zh-word-len} collected a Chinese dataset used to study Chinese word length effects based on data from 30 participants reading 90 sentences. The Potsdam-Allahabad Hindi Eyetracking Corpus contains data from 30 participants reading 153 sentences in Hindi and Urdu \citep{Husain-Vasishth-Srinivasan-2014-hindi}. TURead is a Turkish reading corpus including data from 196 participants reading 192 stimulus texts each consisting of 1 to 3 sentences \citep{Acartrk2023}. See Table \ref{tab:corpora-single-sent} for an overview.

\subsection{Naturalistic single-sentence corpora for different languages}
Single-sentence corpora with more naturalistic stimuli exist for several languages (see Table \ref{tab:corpora-single-sent-en} and Table \ref{tab:corpora-single-sent}). For the UCL Corpus, 205 sentences were sampled from English novels, and eye-tracking data is available for 43 participants. A very recent corpus is RaCCooNS \citep{Frank2023} which includes data from 37 participants reading 200 narrative sentences. The largest (in terms of participants) single-sentence corpus with completely naturalistic stimuli is the Corpus of Eye Movements in L1 and L2 English Reading (CELER) which is a 365-participant corpus with data from 69 native and 296 non-native speakers reading 156 English sentences from the Wall Street Journal. The non-native speakers have five different native languages (Japanese, Arabic, Spanish, Portuguese and Chinese) and half of the sentences in the stimulus corpus are uniquely read by one participant. The Hong Kong Corpus of Chinese Sentence and Passage Reading includes 96 participants reading 300 single-line sentences and 7 multi-line passages in Chinese \citep{Wu2023HKC}. \citet{Zhang2022chinese} collected data from 1,718 participants reading in total 7,577 different Chinese sentences. The Beijing Sentence Corpus (BSC) is a 60-participant Chinese reading corpus with 150 stimulus sentences chosen from the People's Daily where strong political tones had been removed \citep{Pan2021BSC}.
For the CFILT sarcasm dataset, 7 non-native speakers of English read 1,000 sentences of which 350 had been previously labeled as sarcastic and 650 as non-sarcastic \citep{Mishra-Kanojia-Bhattacharyya-2016}. Participants had to rate the sentences as either positive or negative while their eye movements were tracked. A similar dataset is the CFILT sentiment complexity dataset where 5 participants were asked to rate the sentiment (positive, negative or objective) of 1,059 English sentences  \citep{joshi-etal-2014-measuring}.

\begin{table}[h!]
\caption{\label{tab:corpora-single-sent-en}Naturalistic eye-tracking-while-reading corpora for single sentences in \textbf{English}}
\begin{footnotesize}
\begin{tabularx}{\textwidth}{>{\raggedright\hsize=.35\hsize}X>{\raggedright\hsize=.3\hsize}X>{\raggedright\hsize=.35\hsize}X}
\toprule
\textbf{Stimuli}  & \textbf{Participants} & \textbf{Additional characteristics}   \\\midrule

\multicolumn{3}{l}{\textbf{CELER} \citep{celer2022}}\vspace{0.08cm} \\
156 sentences from the Wall Street Journal ($\scriptstyle\# \textstyle w$: 900, $\overline{\scriptstyle\# \textstyle w}$:~11.3) & 365 participants: 69 English L1 and 296 English L2 ($\bar{a}$: 27.3 (6.8))
& The L2 readers have five different L2 backgrounds and half of the stimuli are uniquely read by one participant. \newline \textbf{Eye-tracker}: EyeLink 1000+ \newline \textbf{Sampl. Freq.}: 1,000\,Hz \\
\cmidrule{1-3}

\multicolumn{3}{l}{\textbf{CFILT Sarcasm} \citep{Mishra-Kanojia-Bhattacharyya-2016}} \vspace{0.08cm}\\
1000 sentences, 350 are labeled as sarcastic and 650 as non-sarcastic & 7 non-native English speakers & Participants were asked to label the sentences as either positive or negative. \newline \textbf{Eye-tracker}: EyeLink 1000 \newline \textbf{Sampl. Freq.}: 500\,Hz  \\\cmidrule{1-3}

\multicolumn{3}{l}{\textbf{CFILT Sentiment} \citep{joshi-etal-2014-measuring}} \vspace{0.08cm}\\
1059 movie reviews or twitter posts & 5 participants & Participants were asked to label the sentiment of the sentences as either positive, negative or objective. \newline \textbf{Eye-tracker}: Tobii TX 300 \newline \textbf{Sampl. Freq.}: 300\,Hz \\\cmidrule{1-3}

\multicolumn{3}{l}{\textbf{UCL Corpus} \citep{frank2013reading}}\vspace{0.08cm} \\
205 sentences selected from free online novels ($\scriptstyle\# \textstyle w$: 2,399, $\overline{\scriptstyle\# \textstyle w}$: 13.7 (6.36)) & 43 native speakers ($\bar{a}$: 25.8) & The dataset contains self-paced-reading data as well. \newline \textbf{Eye-tracker}: EyeLink II \newline \textbf{Sampl. Freq.}: 500\,Hz  \\\cmidrule{1-3}

\multicolumn{3}{l}{\textbf{ZuCo 1} \citep{Hollenstein2018zuco1}} \vspace{0.08cm}\\
1107 Wikipedia sentences and movie reviews ($\overline{\scriptstyle\# \textstyle w}$: 19.54 (9.72))& 12 native speakers \newline ($\bar{a}$: 38 (9.8)) & Co-registration of EEG data and each sentence was part of a block with a specific task. \newline \textbf{Eye-tracker}: EyeLink 1000+ \newline \textbf{Sampl. Freq.}: 500\,Hz \\\cmidrule{1-3}

\multicolumn{3}{l}{\textbf{ZuCo 2} \citep{hollenstein-etal-2020-zuco}} \vspace{0.08cm}\\
 739 Wikipedia sentences ($\scriptstyle\# \textstyle w$: 15,138, $\overline{\scriptstyle\# \textstyle w}$: 20.5 (9.2)) & 18 native speakers \newline ($\bar{a}$: 34 (8.3)) & Co-registration of EEG data and each sentence was part of a block with a specific task. \newline \textbf{Eye-tracker}: EyeLink 1000+ \newline \textbf{Sampl. Freq.}: 500\,Hz \\
\botrule
\end{tabularx}
\footnotetext{Abbr.: $\scriptstyle\# \textstyle w$, $\overline{\scriptstyle\# \textstyle w}$ = (mean) number of words of stimuli (SD); $\bar{a}$~=~mean age of participants (SD)}
\end{footnotesize}
\end{table}

\begin{table}[h!]
\caption{\label{tab:corpora-single-sent}Naturalistic eye-tracking-while-reading corpora for single sentences in different languages other than English}
\begin{footnotesize}
\begin{tabularx}{\textwidth}{>{\raggedright\hsize=.01\hsize}X>{\raggedright\hsize=.34\hsize}X>{\raggedright\hsize=.3\hsize}X>{\raggedright\hsize=.35\hsize}X}
\toprule
\textbf{L} & \textbf{Stimuli}  & \textbf{Participants} & \textbf{Additional characteristics}   \\\midrule

\multicolumn{4}{l}{\textbf{Beijing Sentence Corpus} \citep{Pan2021BSC}} \vspace{0.08cm}\\
zh & 150 from the People's Daily newspaper ($\overline{\scriptstyle\# \textstyle w}$:~11.2~(1.6))  & 60 native speakers ($\bar{a}$:~22.0~(2.6)) &  The dataset includes human predictability norms.  \newline \textbf{Eye-tracker}: EyeLink II \newline \textbf{Sampl. Freq.}: 500\,Hz \\\cmidrule{1-4}

\multicolumn{4}{l}{\textbf{Potsdam Sentence Corpus} \citep{kliegl2004psc, PSC}} \vspace{0.08cm}\\
de & 144 sentences constructed around target words ($\overline{\scriptstyle\# \textstyle w}$:~7.9) & 222 native speakers (age: 16–84) & The dataset includes human predictability norms.  \newline \textbf{Eye-tracker}: EyeLink I / EyeLink II \newline \textbf{Sampl. Freq.}: 250 / 500\,Hz \\\cmidrule{1-4}

\multicolumn{4}{l}{\textbf{Potsdam-Allahabad Hindi Eyetracking Corpus} \citep{Husain-Vasishth-Srinivasan-2014-hindi}}\vspace{0.08cm} \\
 hi, ur & 153 sentences from the Hindi-Urdu treebank ($\scriptstyle\# \textstyle w$: 2,610, $\overline{\scriptstyle\# \textstyle w}$:~17.0) & 30 participants & The same sentences were read in Hindi and Urdu script in two separate sessions.   \newline \textbf{Eye-tracker}: SMI iView X HED \newline \textbf{Sampl. Freq.}: 500\,Hz \\\cmidrule{1-4}

\multicolumn{4}{l}{\textbf{Chinese Word Length Effect} \citep{Zang2018zh-word-len}} \vspace{0.08cm}\\
 zh & 90 constructed sentences ($\overline{\scriptstyle\# \textstyle c}$:~19.0~(2.0))  & 30 native speakers ($\bar{a}$: 24.0 (2.0)) & Sentences are rated for their naturalness; includes human predictability norms  \newline \textbf{Eye-tracker}: EyeLink 1000\newline \textbf{Sampl. Freq.}: --\,Hz \\\cmidrule{1-4}

\multicolumn{4}{l}{\textbf{Russian Sentence Corpus} \citep{Laurinavichyute2018RSC}}\vspace{0.08cm} \\
 ru &  144 selected sentences based on target words from the Russian National Corpus ($\scriptstyle\# \textstyle w$: 1,362) & 96 native speakers ($\bar{a}$:~24.0) & The dataset includes human predictability norms.  \newline \textbf{Eye-tracker}: EyeLink 1000+ \newline \textbf{Sampl. Freq.}: 1,000\,Hz \\\cmidrule{1-4}

\multicolumn{4}{l}{\textbf{RaCCooNS} \citep{Frank2023}}\vspace{0.08cm} \\
 nl & 200 narrative sentences ($\scriptstyle\# \textstyle w$: 2,783)  & 37 native speakers ($\bar{a}$: 26.2) &   Co-registration of EEG data.  \newline \textbf{Eye-tracker}: EyeLink 1000+ \newline \textbf{Sampl. Freq.}: 1,000\,Hz \\\cmidrule{1-4}
 
 \multicolumn{4}{l}{\textbf{Hong Kong Corpus of Chinese Sentence and Passage Reading} \citep{Wu2023HKC}}\vspace{0.08cm} \\
 zh & 300 single-line sentences ($\scriptstyle\# \textstyle w$: 5,250) and 7 multi-line passages ($\scriptstyle\# \textstyle w$: 4,967) from newspaper articles & 96 native speakers ($\bar{a}$: 26 (3.64)) &  \textbf{Eye-tracker}: EyeLink 1000 \newline \textbf{Sampl. Freq.}: 1,000\,Hz \\\cmidrule{1-4}

\multicolumn{4}{l}{\textbf{Eye-movement Measures on Words in Chinese Reading} \citep{Zhang2022chinese}}\vspace{0.08cm} \\
 zh & 7,577 sentences ($\scriptstyle\# \textstyle w$: 8,551, $\overline{\scriptstyle\# \textstyle c}$: 22.48) & 1,718 native speakers & \textbf{Eye-tracker}: EyeLink 1000 \newline \textbf{Sampl. Freq.}: 1,000\,Hz \\\cmidrule{1-4}

\multicolumn{4}{l}{\textbf{TURead} \citep{Acartrk2023}} \vspace{0.08cm}\\
 tr & 192 sentences selected from existing Turkish corpora based on target words ($\overline{\scriptstyle\# \textstyle w}$:~15.3~(2.9)); 37 stimuli consist of 2 or 3 sentences & 196 native speakers ($\bar{a}$:~22.7~(2.6)) & Includes human predictability norms; both silent and aloud reading.  \newline \textbf{Eye-tracker}: EyeLink 1000 \newline \textbf{Sampl. Freq.}: 1,000\,Hz \\
\botrule
\end{tabularx}
\footnotetext{Abbr.: L = language of stimuli (ISO-639-1); $\scriptstyle\# \textstyle w$, $\overline{\scriptstyle\# \textstyle w}$ = (mean) number of words of stimuli (SD); $\overline{\scriptstyle\# \textstyle c}$~=~mean number of characters of stimuli (SD); $\bar{a}$~=~mean age of participants (SD)}
\end{footnotesize}
\end{table}

\subsection{Naturalistic self-paced reading corpora for different languages}
Self-paced reading (SPR) is another method to study natural reading where participants read a text word by word and typically control the reading speed by proceeding to the next word by, e.g., pressing a button. There exist a few such corpora in multiple languages (see Table \ref{tab:spr}). The UCL corpus contains SPR data in addition to eye-tracking data \citep{frank2013reading}. 117 psychology students read 361 sentences from English novels. The Natural Stories Corpus \citep{futrell2017natural-stories} consists of SPR data from 181 participants reading (a subset of) 10 English stories that have been edited to contain hard-to-process constructions.

\subsection{Variations of naturalistic eye-tracking-while-reading corpora}
There exist variations of naturalistic reading corpora that study reading in other settings such as the Multimodal Duolingo Bio-Signal Dataset, which contains data from participants navigating a German language learning website \citep{notaro2018duolingo}. Another area where eye-tracking data can provide interesting insights is source code reading \citep{Obaidellah-2019-code-reading}. For example, the EMIP data set contains data of 216 participants reading source code in different programming languages \citep{emip2020}. 

\begin{table}[h]
\begin{footnotesize}
\caption{\label{tab:spr}Self-paced naturalistic reading corpora for different languages for both single sentences and text passages}
\begin{tabularx}{\textwidth}{>{\raggedright\hsize=.01\hsize}X>{\raggedright\hsize=.35\hsize}X>{\raggedright\hsize=.29\hsize}X>{\raggedright\hsize=.35\hsize}X}
\toprule
\textbf{L} & \textbf{Stimuli} & \textbf{Participants} & \textbf{Additional characteristics}  \\\midrule
\multicolumn{4}{l}{\textbf{UCL} \citep{frank2013reading}} \\
  en & 361 sentences selected from free online novels ($\scriptstyle\# \textstyle w$: 4,946, $\overline{\scriptstyle\# \textstyle w}$: 13.7 (6.36))  & 117 psychology students ($\bar{a}$: 18.9) & The dataset contains eye-tracking data as well.  \\
  \cmidrule{1-4}
\multicolumn{4}{l}{\textbf{Natural Stories Corpus} \citep{futrell2017natural-stories}} \\
  en & 10 stories with a total of 485 sentences ($\overline{\scriptstyle\# \textstyle w}$ per sentence: 22.38), the texts have been edited to contain many hard-to-process constructions& 181 participants, not all of them read all 10 stories &  \\
\botrule
\end{tabularx}
\footnotetext{Abbr.: L = language of stimuli (ISO-639-1); $\scriptstyle\# \textstyle w$, $\overline{\scriptstyle\# \textstyle w}$ = (mean) number of words of stimuli (SD); $\bar{a}$~=~mean age of participants (SD)}
\end{footnotesize}
\end{table}

\subsection{Differences of PoTeC to existing corpora}
In sum, the overwhelming majority of naturalistic eye-tracking-while-reading datasets uses English stimulus items with either native or non-native speakers. Moreover, the stimulus materials of the majority of the datasets (be it English or another language) are single (i.e., isolated) sentences rather than paragraphs or texts. 
 Apart from these differences in language and stimulus length, PoTeC differs from already existing datasets in various ways. First, PoTeC is larger than any other existing dataset for naturalistic reading in German in terms of number of participants (for other languages, a few larger datasets do exist). Second, the stimulus texts of PoTeC are relatively demanding and presumably require more cognitive effort to process than, for example, Wikipedia excerpts as used in previous work. This higher difficulty level of the texts is presumably reflected in more complex eye movement patterns. Third, the readers' domain expertise about the topic presented in the stimulus texts is experimentally controlled and assessed by asking specific background questions on the topics covered in the texts that are not directly answered in the texts and thus cannot be derived from the read text.

\section{Methods}
\label{sec:methods}
\begin{figure}[h]%
\centering
\includegraphics[width=0.65\textwidth]{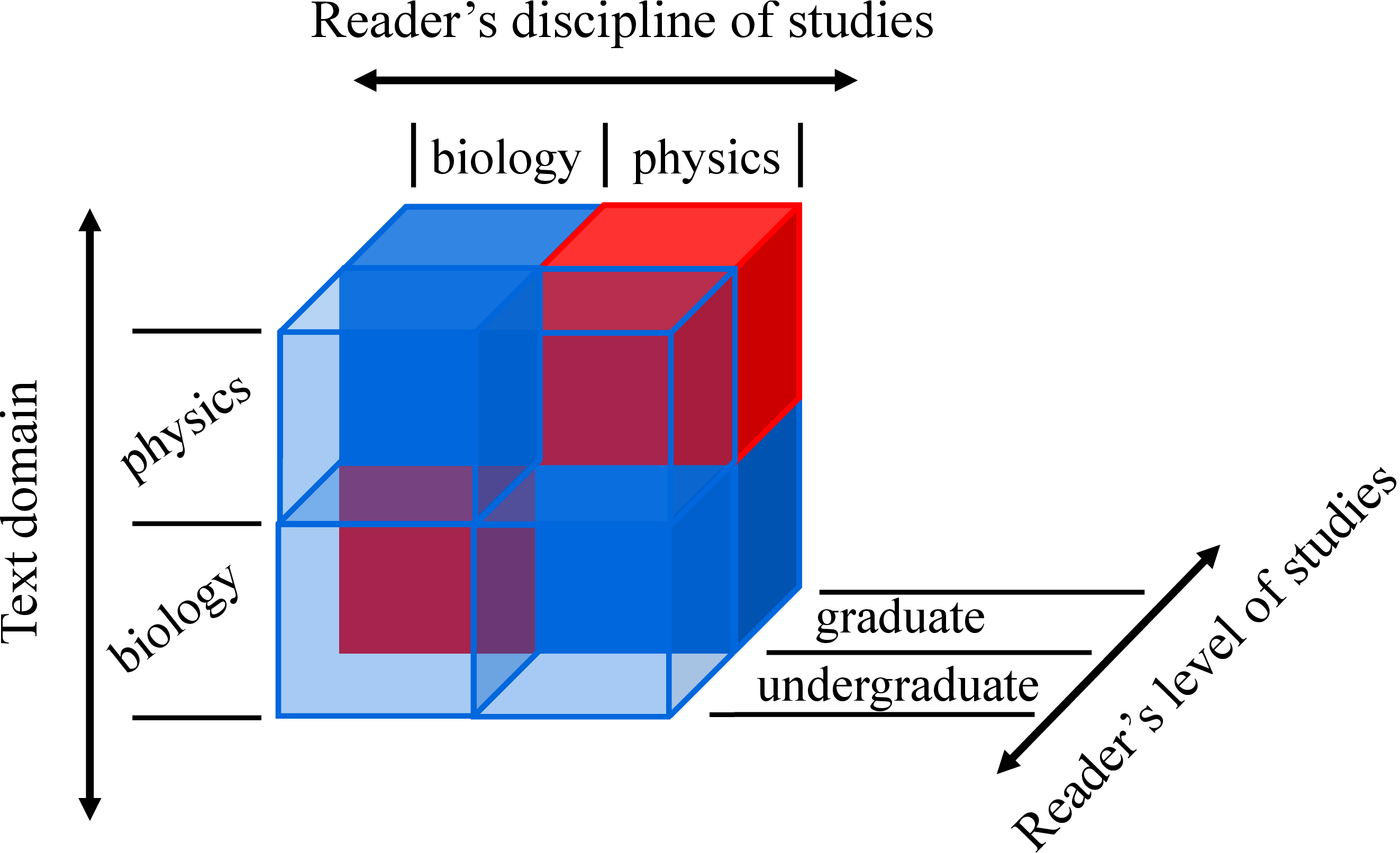}
\caption{\label{fig:study-design}The 2$\times$2$\times$2 fully-crossed factorial study design of PoTeC. The red cubes denote \textit{expert reading}, that is, participants having the \textbf{level of studies} \textit{graduate} who are reading a text, whose \textbf{text domain} is equal to the reader's \textbf{discipline of studies}.}
\end{figure}

PoTeC is an eye-tracking-while-reading dataset using stimulus materials adapted from German university-level textbooks on either physics or biology (see Section~\ref{sec:materials}).
The data collection follows a 2$\times$2$\times$2 fully-crossed factorial design. The three (quasi-experimental) factors are: 1) The reader’s \textbf{discipline of studies} with the levels \textit{physics} and \textit{biology}, which is manipulated between-subjects and within-items: The selection criteria specify that each participant studies exactly one of the two disciplines; and each item (i.e. each text from one of the domains) is read by the participants from both disciplines. 2) The \textbf{text domain} (i.e. the domain of the stimulus texts) with the levels \textit{physics} and \textit{biology} which is manipulated  within-subjects and between-items: all readers read both the \textit{physics} and the \textit{biology} texts, and each text belongs unambiguously to exactly one domain. 3) The reader's \textbf{level of studies} with levels \textit{graduate} and \textit{undergraduate} which is manipulated between-subjects and within-items: each reader belongs to exactly one of the groups and each item is read by each of the groups. The level of studies is defined by the semester or program the students are enrolled in at the time of the experiment: \textit{undergraduate} is defined as \textit{first semester BSc}, while \textit{graduate} is defined as being enrolled in an \textit{MSc} or \textit{PhD} program of the respective discipline of studies. 
See Figure \ref{fig:study-design} for an overview on the study design. 

In addition to the eye-tracking data, the reader's text comprehension and text-independent background knowledge was assessed via comprehension and background questions (see Section~\ref{sec:comp-questions}), demographic information was collected from each of the participants (see Section~\ref{sec:questionnaire}), and comprehensive linguistic annotations of the stimulus texts are provided (see Section~\ref{sec:materials}). The eye-tracking data is pre-processed to obtain fixations and reading measures (among other data formats) as described in more detail in Section \ref{sec:preprocessing}.

\subsection{Materials}
\label{sec:materials}
The present section provides an overview of the materials used and how the materials were annotated. Table \ref{tab:annotation} summarizes the different types of annotations and the tools that were used for the different steps of the stimulus annotation process.

\begin{table}[h]
\begin{footnotesize}
    \caption{\label{tab:annotation}Overview of the tools used to annotate the materials}
    \begin{tabularx}{\textwidth}{>{\raggedright\hsize=.25\hsize}X>{\raggedright\hsize=.4\hsize}X>{\raggedright\hsize=.35\hsize}X}
    \toprule
    \textbf{Annotation} & \textbf{Data Description} & \textbf{Tools} \\\midrule
     
    Manual annotation & \textbf{Word features (manual)}: stored in one file for each text with one row for each word together with all word level tags (see Section \ref{sec:manual-annotation}). & Manual \newline STTS tag set \citep{STTSguidelines} \\\cmidrule{1-3}
      Corpus-based annotation & \textbf{Word features (extracted)}: stored together with the other word level tags (see Section \ref{sec:corpus-level}). & dlexDB online interface\footnotemark[1] \newline dlexDB \citep{Heister2011} \\\cmidrule{1-3}
     Language-model-based annotation & \textbf{Surprisal}: The estimated surprisal values are directly added to the files containing the other word level features (see Section \ref{sec:surprisal}). & Pre-trained language models \newline Python scripts: \texttt{surprisal.py} and \texttt{get\_surprisal.py} \\\cmidrule{1-3}
     Semi-automatic annotation  & \textbf{Dependency \& constituency trees}: The trees are created and stored in separate files per tree type and text each containing all trees for one text (see Section \ref{sec:trees}). & Python script: \texttt{add\_syntax\_trees.py}\\
    \botrule
    \end{tabularx}
    \footnotetext[1]{\url{http://www.dlexdb.de/query/kern/typposlem/}}
\end{footnotesize}
\end{table}

\subsubsection{Stimulus texts}
\begin{figure}[h]
\centering
\includegraphics[width=\textwidth]{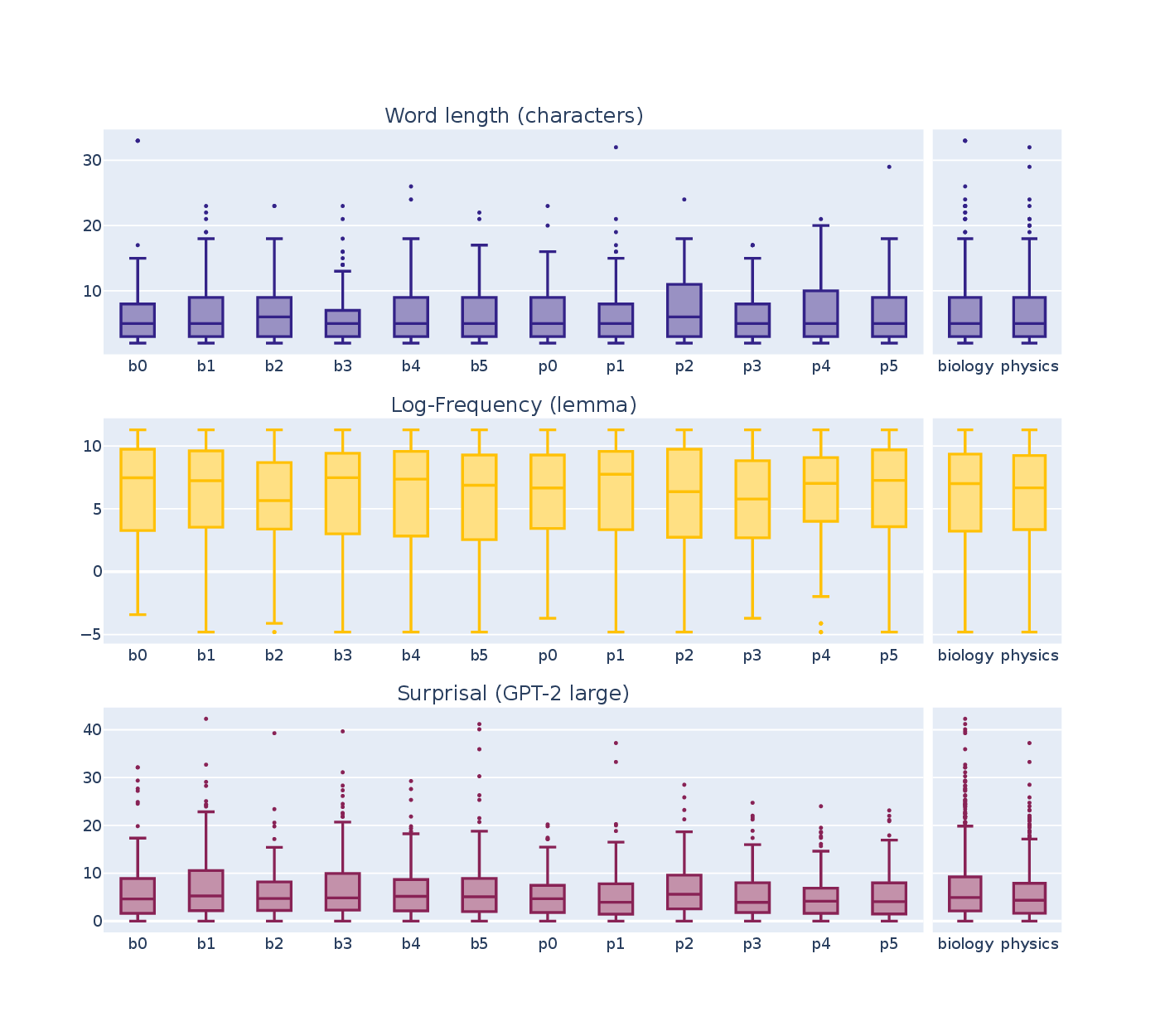}
\caption{\label{fig:text-chars}Domain-specific and text-specific summary of the word length in characters, the log-lexical lemma frequency, and surprisal (as estimated by GPT-2 large).}
\end{figure}

We selected a total of twelve texts from various German university-level physics (six texts) and biology textbooks (six texts) \citep{Demtroeder2, Demtroeder3, Demtroeder4, MolBio, Oekologie, Genetik, ZellMolbio}. The original texts were chosen in a way that each text was approximately 150 words long (min: 126 words; max: 180 words; mean: 158 words). If necessary, the text's content was adjusted to account for the deletion of mathematical formulas, figures and tables. The resulting texts were self-contained, and consisted only of plain text. The texts enable cross-domain comparisons as they are very similar in terms of text features such as lexical frequency (see Figure \ref{fig:text-chars}). Their main difference lies in their content and not in, for example, texts in one domain simply containing less frequent words.

\subsubsection{Comprehension \& background questions}
\label{sec:comp-questions}
For each text, three text comprehension questions and three background questions were created. The text comprehension questions required a thorough understanding of the text, but did not require any additional background knowledge. The background questions, in contrast, tested the general knowledge in the topic presented in the text and hence required background knowledge. The background questions could not be answered with knowledge only acquired through the respective text. All questions had been designed by experts in the respective field. 

\subsubsection{Manual word level annotation}
\label{sec:manual-annotation}

Each stimulus text was manually part-of-speech (PoS) tagged according to the Stuttgart-T\"ubingen-Tagset (STTS) \citep{STTSguidelines}. In addition to the PoS-tag of the word itself, we provide the PoS-tag of \textit{punctuation marks that directly} (i.e., without a white space) \textit{precede} or \textit{follow the word}, as well as hand-crafted tags to indicate whether a word was contained in a constituent that is \textit{in quotes} or \textit{parentheses}. Furthermore, the words were manually tagged for other lexical and orthographic features that arguably affect eye movement behavior in reading, namely whether the word is an \textit{expert} or \textit{general technical term}, with \textit{expert technical term} being a term that is typically only understood by experts in that area (e.g., biology: ``homolog'', physics: ``phasenrichtig'') and \textit{general technical term} being a term that is generally understood (e.g., biology: ``Proteine'', physics: ``Kristalle''). Other tags are whether the word \textit{is} (e.g., ``DNA'') or \textit{contains an abbreviation} (e.g., ``DNA-Fragment''), \textit{contains a non-Latin character or symbol} (e.g., ``$\beta$-D-Glucose''), or \textit{contains a hyphen} (e.g., ``z-Richtung''). Finally, we added various ordinal or binary tags to encode positional information for each word (ordinal \textit{position of the word in the text} and \textit{in the sentence}, and whether it is the \textit{first} or \textit{last word of a clause} or \textit{sentence}). See Table~\ref{tab:manual-features} for the precise definitions of the different hand-crafted word tags. 

 \begin{table}[h!]
 \begin{center}
\caption{\label{tab:manual-features}Definitions of the manually created lexical and sentence-level features}
 \begin{minipage}{\columnwidth}
 \begin{tabularx}{\columnwidth}{>{\raggedright\hsize=.3\hsize}X>{\hsize=.40\hsize}X >{\raggedright\hsize=.3\hsize}X}
 \toprule
 \textbf{Feature}&\textbf{Definition}&\textbf{Example}\\
 \midrule
  \multicolumn{3}{l}{\textit{Lexical and orthographic features}}\\\cmidrule{1-3}
 expert technical term & The word is a technical term typically only understood by experts in the field & homolog (biology), phasenrichtig (physics)\\
 general technical term & The word is a technical term but generally understood without specific expertise required & Proteine (biology), Kristalle (physics)\\
 is abbreviation& The \textit{entire} word is an abbreviation& DNA\\
 contains abbreviation & The word contains an abbreviation &DNA-Fragment\\
 contains symbol& Word contains a symbol &$+$Ende; $\beta$-D-Glucose\\
 contains hyphen & Word contains at least one hyphen that is \textit{not} STTS-tagged as TRUNC &DNA-Fragment; z-Richtung;  $\beta$-D-Glucose\\\cmidrule{1-3}
  \multicolumn{3}{l}{\textit{Linear position information}}\\\cmidrule{1-3}
 word index in text&Position of the word within the current text, irrespective of sentences coded as integer&\\
 word index in sentence& Position of the word within the current sentence coded as integer&\\
  sentence index&Position of the sentence to which the word belongs within the current text coded as integer& \\\cmidrule{1-3}

  \multicolumn{3}{l}{\textit{Punctuation}}\\\cmidrule{1-3}
 STTS punctuation before&STTS-tag of the punctuation that precedes the word if applicable \citep{STTSguidelines}&\\
 STTS punctuation after&STTS-tag of the punctuation that follows the word if applicable \citep{STTSguidelines}&\\
 quote&Word is (part of an expression that is) in quotes& ``Gleisen''\\
 parentheses&	Word is (part of an expression that is) in parentheses& (z.B. dielektrische ... )\\\cmidrule{1-3}
  \multicolumn{3}{l}{\textit{Syntactic features}}\\\cmidrule{1-3}
  clause begin/end& Marks the first/last word of a new clause&\\
  sentence begin/end & Marks the first/last word of a new sentence&\\
   dependency \& constituency trees& Dependency trees for each sentence created semi-automatically& \\
\botrule
 \end{tabularx}
 \end{minipage}
 \end{center}
 \end{table}

\subsubsection{Corpus-based word level annotation}
\label{sec:corpus-level}
Moreover, for each word, several word length measures, lexical frequency measures, and lexical neighborhood measures commonly used in reading research were extracted from the lexical database dlexDB  \citep{Heister2011}, which is based on the reference corpus underlying the  Digital Dictionary of the German Language (DWDS) \citep{dwds}. All extracted values were manually corrected by a linguistic expert labeller. In particular, the type-to-lemma mapping was disambiguated and incorrect database entries (e.g., incorrect lemmatization) were corrected. If the lemma has been manually added, the lemma frequency was coded as missing value. Overviews of all features extracted from dLexDB are provided in Tables~\ref{table:linguistic-repr-dblexdb}, \ref{table:freq-dblexdb}, and \ref{table:neighborhood-dblexdb}.

\begin{table}[h]
\begin{center}
\caption{Definition of the corpus-based features extracted from dlexDB: Linguistic representations \& word length features}
  \label{table:linguistic-repr-dblexdb}
 \begin{minipage}{\columnwidth}
 \begin{tabularx}{\columnwidth}{l X}
 \textbf{Feature}&\textbf{Definition}\\
 \toprule
Type& Orthographic representation of a word as found in the stimulus text (case sensitive)\\
 Lemma&Headword, i.e., an uninflected form that may or may not occur in the stimulus corpus itself\\
 Syllables&Syllables of which the word consists\\\\
  Type length char. & Number of characters of a type\\
 Type length syll. & Number of syllables of a type\\
Lemma length &Number of characters of a lemma\\
  \botrule
 \end{tabularx}
 \end{minipage}
 \end{center}
 \end{table}

\begin{table}[h!]
\caption{Definition of the corpus-based features extracted from dlexDB: Frequency measures}
 \label{table:freq-dblexdb}
 \begin{tabularx}{\columnwidth}{>{\raggedright\hsize=.35\hsize}X >{\hsize=.65\hsize}X}
 \textbf{Feature}&\textbf{Definition}\\
 \toprule
Annotated type frequency&  Number of occurrences of a  unique combination of a type, its STTS tag and its lemma in the corpus (per mio tokens)\\
Type frequency& Number of occurrences of a type in the corpus (per mio tokens) \\ 
  Lemma frequency& Total number of occurrences of types associated with this lemma in the corpus (per mio tokens)\\ 
  Document frequency& The number of documents with at least one occurrence of this type (per 10.000 documents)\\ 
  Sentence frequency& Number of sentences with at least one occurrence of this type (per 100.000 sentences)\\
  Cumulative syllable corpus frequency& Cumulative frequency of the individual syllables of the word in the corpus (per mio tokens)\\
  Cumulative syllable lexicon frequency& Cumulative frequency of the individual syllables of the word as listed in the lexicon (per mio types)\\
 Cumulative character corpus frequency& Cumulative corpus frequency of all characters contained in this type (per mio tokens)\\
 Cumulative character lexicon frequency& Cumulative lexicon frequency of all characters contained in this type (per mio types)\\	
 Cumulative character bigram corpus frequency& Cumulative corpus frequency of all character bigrams contained in this type (per mio tokens)\\
 Cumulative character bigram lexicon frequency&Cumulative lexicon frequency of all character bigrams contained in this type  (per mio types)\\
 Cumulative character trigram corpus frequency&Cumulative corpus frequency of all character trigrams contained in this type  (per mio tokens)\\
Cumulative character trigram lexicon frequency&Cumulative lexicon frequency of all character trigrams contained in this type  (per mio types)\\
 Initial letter frequency& Cumulative frequency of all types sharing the same initial letter (per mio tokens)\\
 Initial bigram frequency&Cumulative frequency of all types sharing the same initial character bigram (per mio tokens)\\
 Initial trigram frequency&Cumulative frequency of all types sharing the same initial character trigram (per mio tokens)\\
 Average conditional probability in bigrams & Conditional probability of the respective word being the second component in word bigrams, given the occurrence of its first component, averaged across all possible bigrams with that word as second component (computed on the basis of the annotated type information)\\
 Average conditional probability in trigrams & Conditional probability of the respective word being the third component in word trigrams, given the occurrence of its first and second component, averaged across all possible trigrams with that word as third component (computed on the basis of the annotated type information)\\
 Familiarity & Cumulative frequency of all types of the same length sharing the same initial trigram\\
 Regularity &  The number of types of the same length sharing the same initial trigram\\
  \botrule
 \end{tabularx}
 \end{table}

 \begin{table}[h]
\begin{center}
\caption{Definition of the corpus-based features extracted from dlexDB: Neighborhood measures}
  \label{table:neighborhood-dblexdb}
 \begin{minipage}{\columnwidth}
 \begin{tabularx}{\columnwidth}{>{\raggedright\hsize=.35\hsize}X >{\hsize=.65\hsize}X}
 \textbf{Feature}&\textbf{Definition}\\
 \toprule
 Cumulative frequency of higher frequency neighbors (Coltheart)&Cumulative frequency of all higher frequency orthographic neighbors according to the definition of \citet{Coltheart1977}: Neighbors need to be of the same length and differ at one character position from each other. E.g., ``Hans'' has higher frequency Coltheart neighbor ``Haus''.\\\\
 Count of of higher frequency neighbors &Number of higher frequency orthographic neighbors according to the definition of \citet{Coltheart1977}\\\\
 Cumulative frequency of all neighbors (Coltheart)&Cumulative frequency of all orthographic neighbors according to the definition of \citet{Coltheart1977} \\\\
 Count of all neighbors (Coltheart)& Number of orthographic neighbors according to the definition of \citet{Coltheart1977} \\\\
 Cumulative frequency of higher frequency neighbors  (Levenshtein) & Cumulative frequency of all higher frequency orthographic neighbors according to the definition of \citet{Levenshtein1966}: Words are neighbors if they differ by one change operation (inserting, deleting or exchanging a character).\\\\
 Count of of higher frequency neighbors (Levenshtein) &Number of higher frequency orthographic neighbors according to the definition of \citet{Levenshtein1966}\\\\
 Cumulative frequency of all neighbors (Levenshtein)&Cumulative frequency of all orthographic neighbors according to the definition of \citet{Levenshtein1966}\\\\
 Count of all neighbors (Levenshtein) &Number of orthographic neighbors according to the definition of  \citet{Levenshtein1966}\\\\
  \botrule
 \end{tabularx}
 \end{minipage}
 \end{center}
 \end{table}

\subsubsection{Language-model-based word level annotation}
\label{sec:surprisal}
In addition to the manual tagging and the dlexDB tags, all words in the texts were annotated with surprisal values obtained from different language models. As has been shown by previous research, surprisal values and their predictive power differ depending on the language model that was used for their estimation \citep{goodkind-bicknell-2018-predictive, wilcox-etal:2020-on-the-predictive-power, wilcox-etal-2023-language}. Therefore, the surprisal values were estimated by three models differing in model architectures and size. Two models had been trained in an auto-regressive manner which takes only the left-hand side context of a word into account to predict the next word, which mimics the incremental nature of human language processing. The first is GerPT2 \citep{Minixhofer_GerPT2_German_large_2020}, a language model that was trained on German data but initialized from the English GPT-2 model \citep{radford2019language}. Both GerPT2 Large\footnote{\url{https://huggingface.co/benjamin/gerpt2-large}} (876M parameters) and Base\footnote{\url{https://huggingface.co/benjamin/gerpt2}} (176M parameters) were used. The second model is LeoLM, a German language model built on Llama 2 \citep{touvron2023llama} and is fine-tuned on German data. We used the 7b\footnote{\url{https://huggingface.co/LeoLM/leo-hessianai-7b}} and 13b\footnote{\url{https://huggingface.co/LeoLM/leo-hessianai-13b}} versions of LeoLM. 
The third model used is a German variant\footnote{\url{https://huggingface.co/bert-base-german-cased}} of BERT \citep{devlin-etal-2019-bert} with 109M parameters. BERT is not an auto-regressive model as it takes both the left- and right-hand context into account. 

In order to obtain the surprisal values, all texts were sub-word-tokenized for the respective model and the tokenized input sequence was processed by the model to obtain the log-probabilities of the individual sub-word-tokens which were then added up to get the surprisal for each word. All surprisal values were estimated once with the respective (left-hand) sentence as context and once with the entire (left-hand) text as context as the participants had also seen the entire text on one page during reading.

  \begin{table}[h!]
\caption{Language-model-based word level features estimated by different model types}
  \label{table:surprisal}
 \begin{tabularx}{\columnwidth}{>{\raggedright\hsize=.3\hsize}X>{\hsize=.4\hsize}X >{\raggedright\hsize=.3\hsize}X}
 \textbf{Feature}&\textbf{Model type}&\textbf{Size}\\
 \toprule
 & \multicolumn{2}{l}{\textbf{Auto-regressive}} \\\cmidrule{2-3}
\multirow{5}{0.3\columnwidth}{sentence context surprisal \newline \& \newline text context surprisal} & GerPT2 base (GPT-based) & 176M parameters \\
 & GerPT2 large (GPT-based) & 876M parameters \\
 & LeoLM 7b (Llama-2-based) & 7B parameters \\
 & LeoLM 13b (Llama-2-based) & 13B parameters \\\cmidrule{2-3}
  & \multicolumn{2}{l}{\textbf{Non-auto-regressive}} \\\cmidrule{2-3}
 & BERT & 190M parameters \\
  \botrule
 \end{tabularx}
 \footnotetext{For all model types, surprisal has been estimated once with the sentence and once with the entire text as context. As the surprisal is estimated on sub-word-token-level, the resulting surprisal values have been added up to obtain the surprisal for one word.} 
 \end{table}

\subsubsection{Semi-automatically created dependency and constituency trees}
\label{sec:trees}

To facilitate future analysis of the data, dependency and constituency trees were added to all sentences. In order to create the dependency trees, the Python library \texttt{spaCy}\footnote{\url{https://spacy.io/}} was used. The tool to create the dependency trees was trained on the TIGER corpus and it therefore uses the TIGER annotation scheme \citep{tiger2004corpus, tiger2003schema}. The tool automatically parses and annotates each word with its respective dependencies. The dependency trees were manually corrected by a German linguistic expert.
The constituency trees were created using the open-source tool benepar \citep{kitaev-etal-2019-multilingual, kitaev-klein-2018-constituency}. The tool receives each text separately as input, splits it into sentences, tokenizes the sentences, annotates the words with PoS-tags and groups the resulting annotated words into sentence constituents and annotates them accordingly. As the entire pipeline relies on pre-trained models for the different tasks, it might result in different PoS-tag annotation compared to our manual tags. The resulting constituency trees were consequently manually corrected such that the PoS-tags correspond with our manual tags and to account for any other errors.

\subsection{Participants}
\label{sec:participants}
75 students\footnote{Data for 76 participants was collected, however, for one participant only the data for 5 out of 12 trials was available which is why this participant's data was eventually excluded.} of the University of Potsdam all of whom were native speakers of German with normal or corrected-to-normal
vision participated in the experiment. They were either students of \textit{biology} or of \textit{physics} in either their first semester of the BSc program (\textit{undergraduate}) or graduate students currently attending an MSc or PhD program (\textit{graduate}). In total there were 12 undergraduate physics students, 20 graduate physics students, 16 undergraduate biology students and 27 graduate biology students.
Participants were requested to not  have consumed any alcohol the day of the experiment and come to the experiment well rested.
Participants received a compensation of 20~EUR. A short overview of the participants mean age and a selection of other characteristics is found in Table \ref{tab:participants}. 

\begin{table}[h]
\caption{Overview over the participants mean age and a selection of other characteristics}
  \label{tab:participants}
 \begin{tabularx}{\columnwidth}{>{\raggedright\hsize=.13\hsize}X>{\raggedright\hsize=.13\hsize}X|>{\raggedright\hsize=.18\hsize}X>{\raggedright\hsize=.13\hsize}X>{\raggedright\hsize=.16\hsize}X>{\raggedright\hsize=.28\hsize}X}
\toprule
\textbf{Discipline of studies} & \textbf{Level of studies} & \textbf{Number of participants} & \textbf{Mean age (SD)} & \textbf{Mean hours of sleep (SD)} & \textbf{Glasses}\\
\midrule
Biology & undergrad. & 16 & 21.5 (3.2) & 7.3 (0.9) & no: 14, yes: 1, N/A: 1\\
 Biology & graduate & 27 & 26.2 (4.1) & 7.3 (1.0) & no: 17, yes: 10\\
 Physics & undergrad. & 12 & 20.5 (3.3) & 6.6 (2.3) & no: 8, yes: 4\\
 Physics & graduate & 20 & 25.7 (2.8) & 7.4 (1.2) & no: 15, yes: 5 \\\cmidrule{1-6}
\multicolumn{2}{l|}{\textbf{Overall}}  & 75 & 24.2 (4.2) & 7.2 (1.3) & no: 54, yes: 20, N/A: 1\\
\bottomrule
\end{tabularx}
\end{table}

\subsection{Experiment Procedure and Technical Set-up}
\label{sec:procedure}
The data collection was carried out in accordance with the Helsinki Declaration \citep{helsinki-declaration2013}. Informed consent was obtained from all participants prior to starting the experiment. 
Participants were instructed about the procedure of the experiment in written form and clarified their questions with the experimenter orally. 
The experiment started with the recording of the eye movements (see Section~\ref{sec:eyetracking}) which was followed by a short demographic questionnaire (see Section~\ref{sec:questionnaire}).  The total duration of the experiment including instructions, camera-setup, breaks, calibrations and questionnaire was approximately 90 minutes.

\subsubsection{Eye-tracking-while-reading}
\label{sec:eyetracking}
 Participants' eye movements (right eye monocular tracking) were recorded at a sampling rate of 1,000\,Hz using an Eyelink 1000 eye tracker manufactured by SR Research with a desktop mounted camera system with a 35\,mm lens. A Cedrus button box was used as a response pad. The experimental presentation and the communication between the presentation computer and the eye tracker was implemented using the Experiment Builder software provided by SR Research. 

The participant was seated at a height-adjustable table to ensure a constant eye-to-screen distance across participants. The participant's head was stabilized using a chin- and forehead rest which should increase the quality of the recorded data. The texts were presented on a 22~inch monitor with a resolution of 1680$\times$1050~pixels and a screen size of 47.5$\times$30\,cm. The eye-to-screen distance measured 61\,cm and the eye-to-camera distance was 65\,cm. Both distance measures were measured as specified in the EyeLink Installation Guide \citep[p.\,15, p.\,70]{eyelink-install-guide}. 

 The texts were presented in a mono-spaced white font (Courier, font size 18) on a black background. The reason for choosing a black background was the rather long duration of the experiment. A bright background color would strain the participants' eyes and potentially lead to wet eyes which has a negative impact on calibration accuracy.

After the set-up and initial calibration (9-point calibration) of the camera and the validation, the participant first read one practice text followed by six practice questions to get familiar with the experimental procedure. The twelve experimental texts were presented in randomized order (separate randomization for each participant). 
  Each experimental trial began with the presentation of the header of the following text on an individual screen. The participant had to press a button to continue to the text which was shown on a new screen. Each text fit onto a single screen. There were no restrictions regarding the time spent on reading each text. After having finished reading the text, the participant had to look at a green sticker that was placed on the bottom right corner of the monitor and at the same time press a button to continue to the questions. This procedure helps to avoid random fixations on the text after the participant has finished reading as they are fixating a specified target away from the areas of interest.  
 Each  question was presented on a separate screen together with four answer options of which the participant had to select one by pressing the respective button on the response pad. It was not possible to go back to the text or previous questions nor was it possible to undo an answer. The order of answer options was randomly shuffled for each participant. The three text comprehension questions always preceded the three background questions in order to minimize memory effects on the response accuracy; the order of the three questions within each type was randomized  for each participant. Participants were informed that some of the questions required background knowledge, however they were not informed which ones. 

 If necessary, re-calibrations were performed before the beginning of a new trial followed by another validation. For all of the participants, re-calibrations were performed throughout the experiment (see Table \ref{tab:avg-validations} for an overview on the calibrations and validations performed). Appendix \ref{sec:A3} presents a table where all the average validation scores for each session are listed together with the number of validations and calibrations performed in that session.
 Participants were allowed to take a break before the beginning of a new trial. 

\begin{table}[h]
\caption{\label{tab:avg-validations} Statistics over the number of calibrations and validations performed in the experiment and the validation scores.}
\begin{tabularx}{\textwidth}{X|XXXX|XXXX|XXXX}\toprule
\textbf{Type} & \multicolumn{4}{l|}{\textbf{Validation score (avg)}} & \multicolumn{4}{l|}{\textbf{Validation score (max)}} & \multicolumn{4}{l}{\textbf{\# per session}} \\
 & \textit{mean} & \textit{std} & \textit{max} & \textit{min} & \textit{mean} & \textit{std} & \textit{max} & \textit{min} & \textit{mean} & \textit{std} & \textit{max} & \textit{min} \\\cmidrule{1-13}
cal &  &  &  &  &  &  &  &  & 14.418 & 4.73 & 27 & 2 \\
val & 0.421 & 0.253 & 2.37 & 0.188 & 0.84 & 0.567 & 5.045 & 0.422 & 8.899 & 3.07 & 15 & 2 \\\botrule
\end{tabularx}%
\footnotetext{Note that the validation scores (avg and max) are both averages over all the validations performed in one session. See Appendix \ref{sec:A3} for a session overview. SR research recommends an average validation score below $0.5$.}
\end{table}

\subsubsection{Demographic questionnaire}
\label{sec:questionnaire}
 After the eye-tracking experiment was concluded, participants had to fill in a short demographic questionnaire. The following information was collected: the field of studies (including  area of specialization if applicable) and the current semester of studies in order to verify the expert status, gender, age, handedness, whether the participant was wearing (soft or hard) contact lenses or glasses, hours of sleep the night before the experiment, alcohol consumption within 24h hours before the experiment, whether or not the participant had grown up bilingually, and the state (``Bundesland'') where the German language was acquired. 

\subsection{Data Preprocessing}
\label{sec:preprocessing}
The eye-tracking data for each participant and text were pre-processed and are made available in different formats. Whenever the tools and scripts are created by ourselves and not protected by a license or copyright, the tools used to complete the different steps are made publicly available. Table~\ref{tab:preprocessing} provides an overview of the different preprocessing steps and the tools used. 

\begin{table}[h!]
\begin{footnotesize}
    
    \centering
    \caption{\label{tab:preprocessing}Data processing pipeline including the tools that were used}
    \begin{tabularx}{\textwidth}{>{\raggedright\hsize=.25\hsize}X>{\raggedright\hsize=.5\hsize}X>{\raggedright\hsize=.25\hsize}X}
    \toprule
    \textbf{Pre-processing Step} & \textbf{Data Description} & \textbf{Tools} \\\midrule
    
     \multicolumn{3}{l}{\textbf{Pre-processing of eye movement data}}\\\cmidrule{1-3}
     
     \textit{Collect}: \newline raw data & \textbf{Raw data (as-is)}: Raw data including metadata \textit{for each session} in one file. At experiment-time non-human-readable \texttt{edf} files are written which are directly converted to human-readable \texttt{asc} files. & EyeLink 1000, Experiment Builder software (SR Research), SR Research \texttt{edf2asc} \\\cmidrule{1-3}

    \textit{Pre-process}: \newline raw data & \textbf{Raw data (pre-processed)}: Raw data as collected is pre-processed to only contain the relevant samples (right eye $x$ and $y$ coordinates, pupil diameter, timestamp) \textit{for each trial}. The files contains one data sample per line (\texttt{tsv} format). & \texttt{asc\_to\_csv.py},\newline SR Research \texttt{asc2csv} \\\cmidrule{1-3}
     
     \textit{Compute}: \newline raw data $\rightarrow$ fixation data & \textbf{Uncorrected fixation data}: From the pre-processed raw data, fixations are computed. Contains the character-level area of interest for each fixation and fixation and saccade features.  & Eyelink Data Viewer \citep{dataviewer}, default parameter settings  \\\cmidrule{1-3}

     \textit{Correct}: \newline fixation data $\rightarrow$ corrected fixation data & \textbf{Fixations}: Manually corrected fixation data. Contains information about whether the fixation was corrected or not and if so, information on the original fixation (see Section \ref{sec:manual-data-cor}). & Python script: \texttt{correct\_fixations.py} and \texttt{split\_fixation\_} \texttt{report.py}\\\cmidrule{1-3}

    \multicolumn{3}{l}{\textbf{Pre-processing of stimulus material}}\\\cmidrule{1-3}
          
     \textit{Compute}:\newline character index $\rightarrow$ word index & \textbf{Character to word mapping}: The indices of the character-level areas of interest are mapped to the word indices in the respective text. & Python script: \texttt{char\_index\_to\_} \texttt{word\_index.py}\\\cmidrule{1-3}
     
     \textit{Compute}: \newline text $\rightarrow$ word/sentence limits & \textbf{Word limits and sentence limits}: Contains information about the first and the last character index (i.e., aois) of each word and each sentence in each text. & Python script: \texttt{create\_word\_} \texttt{aoi\_limits.py}\\\cmidrule{1-3}
     
    \multicolumn{3}{l}{\textbf{Pre-processing of stimulus material and eye movement data combined}}\\\cmidrule{1-3}
     
     \textit{Compute}: \newline corrected fixation data $\rightarrow$ word level reading measures & \textbf{Reading measures}: For each trial, word level reading measures are computed that are written to a separate file for each reader and text (see Section \ref{sec:rm}). & Python script: \texttt{compute\_reading\_} \texttt{measures.py}\\\cmidrule{1-3}
     
     \textit{Merge}: \newline corrected fixation data with character and words & \textbf{Scanpaths}: Each character-level fixation in all scanpaths was merged with the fixated character and word, and information on trial-level like the text identifier is added (see Section \ref{sec:structure}). & Python script: \texttt{generate\_scanpaths.py} \\\cmidrule{1-3}
     
     \textit{Merge}: \newline reading measures with stimulus corpus data  & \textbf{Reading measures merged}: Merge reading measures with all features at all levels that have not been included yet (see Table \ref{tab:feature-levels} for an overview) & Python script: \texttt{merge\_reading\_} \texttt{measures.py} \\\cmidrule{1-3}
     
     \textit{Merge}: \newline scanpaths with stimulus corpus data & \textbf{Scanpaths merged}: Merge scanpaths with all features at all levels that have not been included yet (see Table \ref{tab:feature-levels} for an overview). & Python script: \texttt{merge\_scanpaths.py}\\

    \botrule
    \end{tabularx}
\end{footnotesize}
    
\end{table}

\subsubsection{Pre-processing of raw data}
The data files originally written by the eye-tracking device are non-human readable \texttt{edf} files which were directly converted to a human-readable \texttt{asc} format using the SR Research \texttt{edf2asc} tool. The \texttt{asc }files contain the data for one session including metadata and need to be parsed to extract the relevant samples for each trial and store it in a separate \texttt{tsv} file which then contains one sample per line for one trial. One sample consists of the $x$ and $y$ coordinates for the tracked eye and the timestamp. 

\subsubsection{Computation of fixations}

Fixations and saccades were computed from the pre-processed raw data using the EyeLink Data Viewer software provided by SR Research with the default parameter settings \citep{dataviewer}. Subsequently, each fixation was mapped to the character index in the text that was fixated and annotated with the line index and the character index in that line. Fixations on the white space between two words were mapped to the closest character. 

\subsubsection{Manual correction of the fixation data}
\label{sec:manual-data-cor}
Visual inspection of the fixation data revealed that in certain fixation sequences, vertical calibration error gradually increased over time. This measurement error was manually corrected by adjusting the fixation-to-character mapping (i.e., re-mapping a fixation to the character in the line above or below the currently mapped character). 
Horizontal calibration drift could not be corrected as there is no way to infer the magnitude of the measurement error from the data. An exception are fixations on the first or last word of a line. If the fixation was just next to the line and not on the word, it was corrected to be on the first, respectively last word of the line. In any case, horizontal measurement error is less dramatic for the purpose of this study (and reading experiments in general) as fixations are eventually mapped to words, thus a horizontal measurement error will only result in a wrong fixation-to-word mapping when the measurement error is larger than the distance of the real fixation location to the word boundary. In contrast, even small measurement error on the vertical axis can lead to incorrectly mapping the fixation to be on a word in the line above or below. 
In some trials, the participant did not only read the text but also scanned the screen without reading. Such sequences of fixations can be easily distinguished from eye movements reflecting reading by visual inspection (automatized approaches exist, but are less accurate, see \citet{Biedert2012}). Whenever such sequences of non-reading fixations occurred at the beginning or the end of a trial, they were deleted within the process of correcting the fixation locations. In addition, very short fixations that looked like optical artifacts were deleted. All three stages of the data up to this point, the pre-processed raw data, the originally computed uncorrected fixations and the manually corrected fixation data (corrected fixation location and deleted non-reading sequences) are made available in the data repository.

\subsubsection{Computation of word-level reading measures}
\label{sec:rm}
From the fixation data, various reading measures commonly used in reading research were computed for each text and reader. Each word (defined by the surrounding white spaces) was considered one area of interest for all measures except for \textit{landing position} which was based on the characters within a word. Fixations on the punctuation marks were considered to belong to the preceding word by default and to the following word in case of opening parentheses or opening quotation marks. Definitions of the various measures are provided in Table~\ref{tab:rm}. Note that several subsets of these measures are linearly dependent. 

\begin{table}[h!]
 \caption{Definition of the word-level eye movement measures.}
 \label{tab:rm}
\centering
\begin{footnotesize}
 \begin{tabularx}{\textwidth}{>{\raggedright\hsize=.25\hsize}X>{\raggedright\hsize=.08\hsize}X>{\raggedright\hsize=.67\hsize}X}
 \toprule
 \textbf{Measure}& \textbf{Abbr.}& \textbf{Definition}\\\cmidrule{1-3}
 \multicolumn{3}{l}{\textit{Continuous measures (in ms)}}\\\cmidrule{1-3}
 first-fixation duration&FFD	& duration of the first fixation on a word if this word is fixated in first-pass reading, otherwise 0\\
 first duration&FD&duration of the first fixation on a word (identical to  FFD if not skipped in the first-pass)\\
 first-pass reading time &FPRT&sum of the durations of all first-pass fixations on a word (0 if the word was skipped in the first-pass)\\
 single-fixation duration&SFD &  duration of the only first-pass fixation on a word, 0 if the word was skipped or more than one fixations occurred in the first-pass (equals FFD in case of a single first-pass fixation)\\
first-reading time&FRT&  sum of the duration of all fixations from first fixating the word (independent if the first fixations occurs in first-pass reading) until leaving the word for the first time (equals FPRT in case the word was fixated in the first-pass)\\
 total-fixation time&TFT& sum of all fixations on a word (FPRT$+$RRT)\\
 re-reading time&RRT&sum of the durations of all fixations on a word that do not belong to the first-pass (TFT$-$FPRT)\\
inclusive regression-path duration&RPD\_inc&  sum of all fixation durations starting from the first first-pass fixation on a word until fixating a word to the right of this word (including all regressive fixations on previous words), 0 if the word was not fixated in the first-pass  (RPD\_exc$+$RBRT) 	\\
exclusive regression-path duration&RPD\_exc&  sum of all fixation durations after initiating a first-pass regression from a word until fixating a word to the right of this word, without counting fixations on the word itself (RPD\_inc$-$RBRT)\\  
right-bounded reading time&RBRT&sum of all fixation durations on a word until a word to the right of this word is fixated (RPD\_inc$-$RPD\_exc). \\\cmidrule{1-3}
\multicolumn{3}{l}{\textit{Binary measures}}\\\cmidrule{1-3}
 fixation & Fix   & 1 if the word was fixated, otherwise 0 (FPF \texttt{or} RR)\\
 first-pass fixation & FPF &  1 if the word was fixated in the first-pass, otherwise 0 \\
 first-pass regression & FPReg & 1 if a regression was initiated in the first-pass reading of the word, otherwise 0 (\texttt{sign(}RPD\_exc\texttt{)})\\
 re-reading & RR  & 1 if the word was fixated after the first-pass reading, otherwise 0 (\texttt{sign(}RRT\texttt{)})\\\cmidrule{1-3}

 \multicolumn{3}{l}{\textit{Ordinal measures}}\\\cmidrule{1-3}
 total fixation count & TFC & number of all fixations on a word\\
 landing position & LP &position of the first saccade on the word expressed by ordinal position of the fixated character \\
 incoming saccade length&SL\_in &   length of the saccade that leads to first fixation on a word in number of words; positive sign if the saccade is a progressive one, negative sign if it is a regression\\
 outgoing saccade length & SL\_out &  length of the first saccade that leaves the word in number of words; positive sign if the saccade is a progressive one, negative sign if it is a regression; 0 if the word is never fixated\\
 total count of outgoing regressions & TRC\_out &total number of regressive saccades initiated from this word\\
 total count of incoming regressions & TRC\_in &total number of regressive saccades landing on this word\\
 \botrule
 \end{tabularx}
 \end{footnotesize}
 \end{table}

\subsection{Summary of the Available Features}
\label{sec:structure}
After all the data was annotated and pre-processed, PoTeC has different features available at different levels (word-, text-, reader-, fixation- and trial-level)\footnote{For some textual features like the the discipline of studies or the text domain, a numerical encoding was added for easy computational processing while still keeping an explicitly understandable text-encoded label (e.g., level of studies \textit{physics}). As both features encode the same information, it won't be explicitly stated in the following overview but is defined in the data repository.}. 
\textbf{1) Word-level features}: All word-level features, which include manual and (semi)-automatically computed or estimated tags, are defined in Tables~\ref{tab:manual-features}--\ref{table:surprisal}. 
\textbf{2) Text-level features}: Each text was given a unique identifier encoded as textual feature explicitly encoding the text domain. In addition to the text itself, the text's domain, the comprehension questions, all the answer options, and the encoding of the correct answers to all questions are text-level features.
\textbf{3) Reader-level features}: The data collected through the demographic questionnaire and a unique reader identifier are reader-level features as well as the level of studies and the discipline of studies. The order of the stimuli texts and the order of the answer options for each of the questions are reader-level features and the average response accuracy over all questions for each question type.
\textbf{4) Fixation-level features}: A chronological fixation index, fixation duration, previous saccade duration, next saccade duration and the area of interest are available fixation-level features. In addition, whether or not the fixation has been manually corrected, the original chronological fixation index, the uncorrected fixation location and the uncorrected area of interest are added as features.
\textbf{5) Trial-level features}: For each text read by a participant, the response accuracy for all questions and the mean accuracy for all text and all background questions are provided. In addition, a feature was added to each trial that encodes whether this specific trial is \textit{expert reading} or not (see Figure \ref{fig:study-design}). In addition, trial-level features include the word-level reading measures that have been computed for each text and reader as described in Section~\ref{sec:rm}. 

\begin{table}[h]
    \caption{\label{tab:feature-levels}Available features of PoTeC at different levels}
    \begin{tabularx}{\textwidth}{lX}
    \textbf{Feature level} &  \textbf{Features}\\\toprule
     Reader-level & Unique reader identifier, age, handedness, gender, how many hours of sleep the reader had the night before the experiment, whether of not the reader wears glasses, whether or not the participant consumed alcohol within 24 hours before the experiment start, \textbf{level of studies}, \textbf{discipline of studies}, and \textbf{domain expert level} (e.g., \textit{physics-undergraduate}), the order of texts in this session, and the order of answer options for all question.
     \\\\
     Text-level &  Unique text identifier, text, questions, answer options, the correct answer for all questions, and \textbf{text domain}.  \\\\
     Trial-level & Answer accuracy for all questions for the respective text, mean accuracies for text and background questions for the text, and whether or not the trial is an example of \textbf{expert reading} (i.e., the text is read by an expert in the text domain) and the word level reading measures defined in Table \ref{tab:rm}.   \\\\
     Word-level & All features defined in Tables \ref{tab:manual-features}--\ref{table:surprisal}.  \\\\
     Fixation-level & Chronological fixation index, fixation duration, previous saccade duration, next saccade duration, whether or not the fixation has been manually corrected, the original chronological fixation index, fixated character and word, line index of the fixated character and the character index in the line.  \\
    \botrule
    \end{tabularx}
    \footnotetext{Note: The features marked in bold are either the experimental factors or features based on the experimental factors.}
\end{table}

The fixation data and reading measures are processed further and merged with different features of the corpus material on different levels to facilitate a simple use of the data. Note that these steps do not add any additional information but combine and merge the existing corpus data in such a way that common analyses of eye movement data for both psycholinguistic as well as NLP use cases are simplified and therefore encouraged. 
As the fixation data is originally only associated with the index of the fixated character, it is further processed to explicitly include the fixated character and word. Also included are word-level features of the fixated word as well as most trial-, session-, reader- and text-level features in addition to the fixation-level features. The resulting data representation is then referred to as the scanpath for this reader and text which represents the chronological order of the eye movements on the text. The reading measures are merged with all the remaining reader-, text-, trial-, and session-level features that had not been added before as well as the remaining word-level features (i.e., the various word-level tags).  
Both the reading measures data and the scanpaths are made available in a format including only the most important features and in an extended format that includes all available features merged with the eye movement data (compare Table~\ref{tab:preprocessing}). Note that those two data representations represent the same data in two very different formats with the scanpath data representing one fixation per line in chronological fixation order and the reading measures data representing the eye movement data at the word-level with one line representing one word in the order of the text.

\subsection{Accessing the Data}
\label{sec:data-access}
PoTeC consists of various different types of data which include on the one hand many large data files and on the other hand different Python scripts. To account for the differences in those data formats two channels have been chosen to share the data. The data files are stored in an Open Science Framework (OSF) repository: \url{https://osf.io/dn5hp/}. The code is stored in a GitHub repository: \url{https://github.com/DiLi-Lab/PoTeC}. The GitHub repository can be cloned locally and the data files can be automatically downloaded using a Python script\footnote{\texttt{download\_data\_files.py}, please consult the README files in the repository for further instructions.}. Alternatively, the data files can be manually downloaded from OSF. 

Additionally, PoTeC has been integrated into the Python package \texttt{pymovements}\footnote{\url{https://pymovements.readthedocs.io/en/stable/reference/index.html}} \citep{pymovements}. The package allows for downloading the raw data directly within a Python or R script and can then be further processed using the package. For example, events can be detected or plots can be created to visualize the data\footnote{More instructions are found in our GitHub repository.}.

\section{Usage of the Data}\label{sec:data-use}
The data has already been used for different research purposes. For example, 
\citet{Krakowczyk2023} used PoTeC to develop Explainable Artificial Intelligence (XAI) methods for analysing deep neural networks processing eye-tracking data. 
\citet{hollenstein2021} fine-tuned large language models on different eye-tracking corpora including PoTeC to understand better to what extent the representations learned by large language models are comparable to human reading behavior. \citet{MakowskiECML2018} used PoTeC for biometric identification and the prediction of reading comprehension.  

\section{Analyses of the Data}\label{sec:pla}
We have already illustrated a few past use cases for PoTeC in Section~\ref{sec:data-use}. In the following, we present a series of analyses on a range of reading measures (first-pass reading time, total fixation time, re-reading time, first-pass regressions) to test the effects of the word- and sentence-level features as well as the impact of expertise on eye-movements. More specifically, we deploy hierarchical (generalized) linear-mixed models to:
\begin{enumerate}
    \item explore the reading behavior of experts and non-experts (see Figure~\ref{fig:study-design} for a visual representation of what we refer to as \textit{expert reading}),
    \item test the effect of different word- and sentence-level features such as word length, surprisal, and lexical frequency, as well as whether the word represents an expert term, whether it was read in the expert reading condition (see Table~\ref{tab:feature-levels}), and the reader's discipline of studies.
\end{enumerate}

We model each word-level reading measure $y$ using \texttt{expert reading} (binary; 1=stimulus text is read by a \textit{graduate} student of the \textit{discipline} that constitutes the \textit{text domain}, 0=else; see Figure \ref{fig:study-design}), \texttt{expert technical term}\footnote{Note that in the data, there are two types of technical terms. Please refer to Section \ref{sec:manual-annotation} and Table \ref{tab:manual-features} for an explanation of those terms and all the features used in this present section.} (binary; 1=expert technical term that is not generally understandable, 0=else), \texttt{reader discipline} (binary; 1=physics, 0=biology; reader's discipline of studies), \texttt{word length}, \texttt{log-lemma frequency} and \texttt{lexicalized surprisal}\footnote{We used sentence-based surprisal values extracted from GPT-2 large.} as predictors, formalized as follows:
\begin{equation*}
\begin{split} \label{eq:lmm}
    y_{ij} = & g(\beta_0 + \beta_{0i} + \beta_{1} ~ \texttt{expert reading}_{i} + \beta_{2} ~ \texttt{reader discipline}_{i}+  \\ &\quad    \beta_{3} ~ \texttt{reader discipline}_{i} * \texttt{expert reading}_{i}+ \\ &\quad \beta_{4} ~ \texttt{expert technical term}_{i} + \beta_{5} ~\texttt{word length}_{j} + \\ &\quad \beta_{6} ~ \texttt{expert reading}_{i} * \texttt{word length}_{j}+ \beta_{7} ~\texttt{log-lemma frequency}_{j}  +  \\ &\quad \beta_{8} ~ \texttt{expert reading}_{i} * \texttt{log-lemma frequency}_{j}+ \beta_{9} ~\texttt{surprisal}_{j} + \\ &\quad     \beta_{10} ~ \texttt{expert reading}_{i} * \texttt{surprisal}_{j})
\end{split}
\end{equation*}
where $y_{ij}$ refers to the eye-tracking reading measure of subject $i$ for the $j$th word in the stimulus corpus across all texts. $\beta_0$ represents the global intercept, and $\beta_{0i}$ the random intercept for subject $i$. $g(\cdot)$ denotes the linking function: $g(z)=\ln \frac{z}{1-z}$ for the binary measures (first-pass regression) with $y_{ij}$ following a Bernoulli distribution (i.e., logistic regression); and the identity function for the remaining continuous measures with $y_{ij}$ following a log-normal distribution (i.e., linear regression on a log-transformed dependent variable). We run each model for $6000$ iterations using standard priors. We provide the Stan code including priors in the repository.

\begin{figure}[h]%
\label{fig:effect-sizes}
\centering
\includegraphics[width=\textwidth]{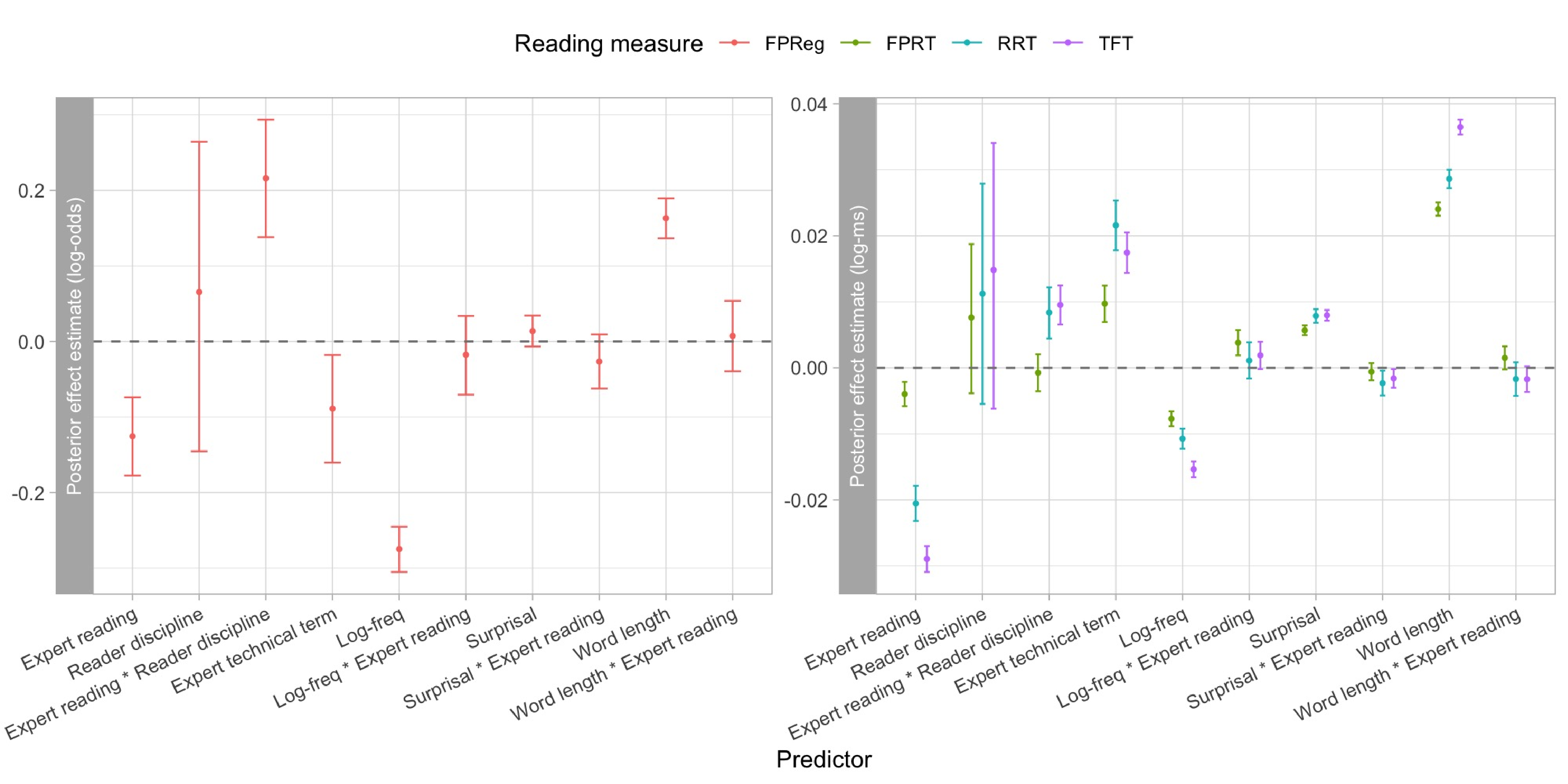}
\caption{Posterior effect estimates for predictors \emph{expert reading}, \emph{reader discipline}, the interaction of \emph{expert reading} with \emph{reader discipline}, \emph{expert technical term} as well as \emph{log-lexical frequency}, \emph{surprisal} and \emph{word length} and the interactions of \emph{log-lexical frequency}, \emph{surprisal}, and \emph{word length} with \emph{expert reading}.}\label{fig:exp-nonexp}
\end{figure}

With respect to the effects of word features on reading times, we find different expected effects. Higher \textit{lemma-frequency} facilitates processing (shorter reading times, fewer regressions) while increased \textit{word length} is associated with increased processing effort (longer reading times, more regressions). Similarly, high \textit{surprisal} was associated with longer reading times. Moreover, participants showed increased reading times on \textit{expert technical terms}. We find that \textit{expert reading} is characterized by shorter reading times, most strongly observed in total fixation times, and fewer regressions. Lastly, while we did not find any effects of \textit{reader discipline} itself on any of the reading measures, the positive interaction between \textit{expert reading} and \textit{reader discipline} suggests that physics graduates (i.e., experts) make more regressions and exhibit higher total-fixation and re-reading times during expert reading.

\section{Summary \& Conclusion}\label{sec:conclusion}
We presented PoTeC, a naturalistic German eye-tracking-while-reading corpus. PoTeC is the first eye-tracking dataset using a novel experiment design where the readers' expertise has been manipulated within-subject. That is, experts and beginners from two different disciplines read texts from both disciplines. The stimulus corpus has been comprehensively linguistically annotated with a wide range of features on different linguistic levels obtained through manual annotation, extracted from databases or computed by state-of-the-art computational models. The eye movement data is made available at \textit{all} pre-processing stages with maximal flexibility for the user to chose which version of the data is useful for their use case (e.g., areas of interest at character- and word-level, uncorrected and corrected fixation data, scanpath data, raw data samples, etc.). The data is released with all the code that has been written for annotating or pre-processing the data which means that the entire pipeline can be fully reproduced and the code can be adapted to other use cases. Extensive demographic data has been collected for the participants and reading comprehension scores have been collected through comprehension questions and, in addition, the readers' background knowledge has been assessed through background questions on the different topics of the stimulus texts. The data and code is made available in a way that creates maximal transparency, maximally increases the re-usability of the data and the code, and makes the data accessible in a user-friendly way using easy to understand data formats. In addition, the data has been integrated into an existing open-source software\footnote{\url{https://pymovements.readthedocs.io/en/stable/reference/index.html}} for the processing of eye-movements that can be used in Python and R.

Given the different features described above, we envision different use cases of the data: 1) PoTeC can be used to study within-subject expert and non-expert reading patterns. As our exploratory analyses have shown, expert reading behavior can be characterized by shorter reading times when compared with non-expert reading. PoTeC is the first eye-tracking-while-reading dataset to allow such analyses. 2) All of these features can be used to train computational models to learn how to distinguish expert readers from non-expert readers. That is, models could be trained, for example, to infer the reader's domain expertise when presented with a scanpath on a particular text. Such models could be used to conduct tests in e-learning scenarios to assess the learners progress. 3) The naturalistic reading data can be leveraged for NLP purposes without specifically studying expert reading. 4) The aim of the corpus is not only to make all the data available but also to publish any scripts and tools that were used to preprocess the data. This should increase transparency and foster high-quality data as it allows for analyzing not only the data but also the data collection and preprocessing pipeline which can help to continuously improve future data collections and algorithms to preprocess eye-tracking data. 5) Making the uncorrected and the manually corrected fixation data available allows for investigating how fixation data can be corrected automatically. PoTeC is the first corpus to make this kind of data available. In particular, the data can be used to train computational models to automatically correct fixation data as the manual correction is a \textit{very} time-consuming endeavor but in practice often necessary and often part of preprocessing pipelines. 

There exists many more use cases such as: creating models based on raw data, analyses of high and low reading comprehension, inference of the reader's reading comprehension, inference of whether a text is difficult for a reader or whether it is a text from their domain of expertise. Our vision is that the corpus can be used for all of the above mentioned use cases and many more that go beyond psycholinguistics, cognitive reading research or NLP. Finally, the work we described aims to encourage further eye-tracking data collections that can complement the existing data.
 
\backmatter

\bmhead{Acknowledgments}
We thank Shravan Vasishth for generously granting access to his eye-tracking lab and for providing the necessary financial support to compensate the participants, enabling the successful execution of the experiment. We further thank Marius Huber for his assistance with setting up the data repository. 

This work was partially funded by the Swiss National Science Foundation under grant 212276 and the German Federal Ministry of Education and Research under grant 01$\vert$ S20043.

\bmhead{Open Practices Statement}
Materials and code are available in a GitHub repository at \url{https://github.com/DiLi-Lab/PoTeC}. The large data files are stored at \url{https://osf.io/dn5hp/}, but can be downloaded via the GitHub repository. None of the reported studies were preregistered.

\begin{appendices}

\section{Text Characteristics}\label{sec:A2}
\begin{figure}[ht!]%
\centering
\includegraphics[width=0.9\textwidth]{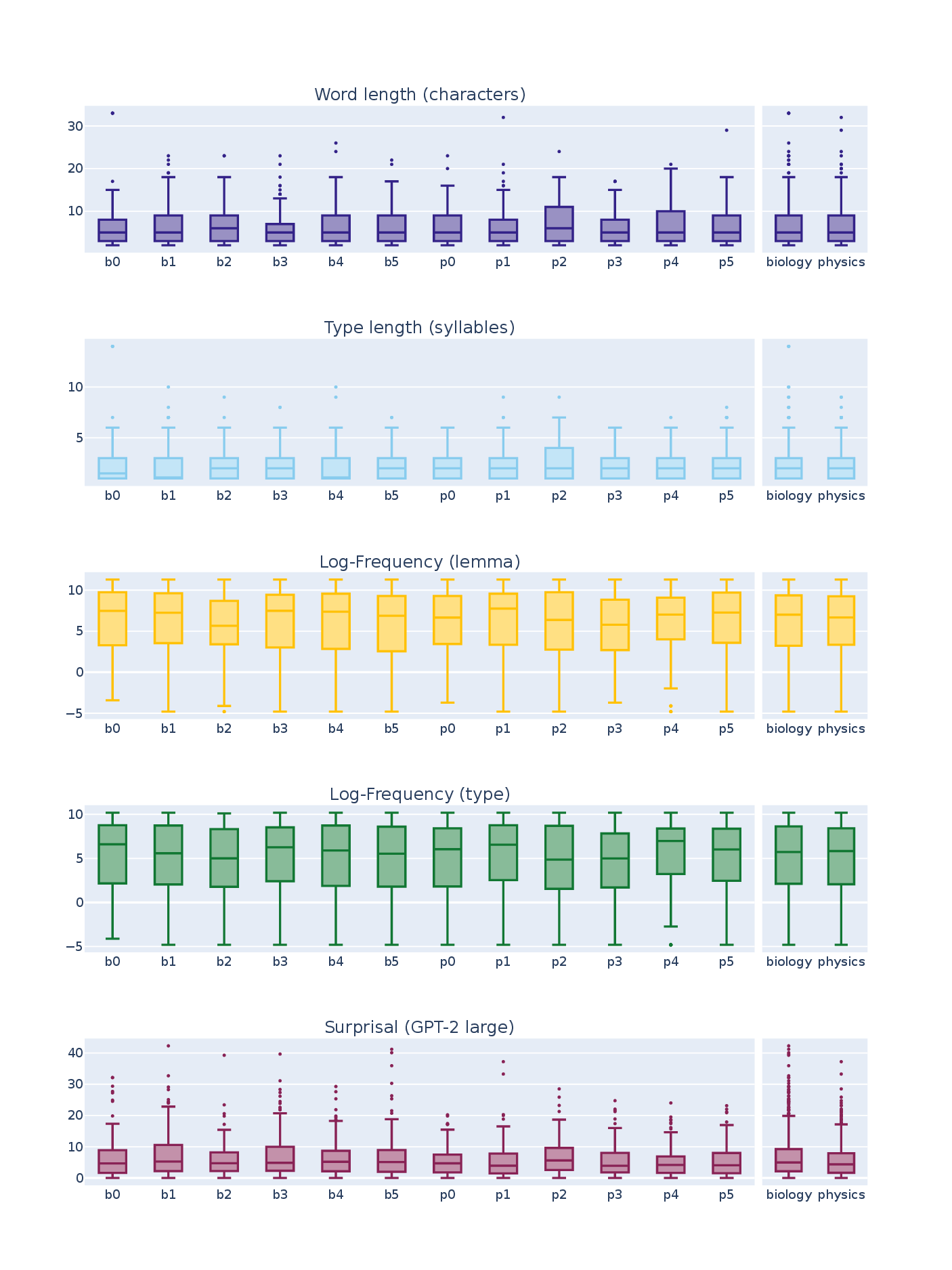}
\caption{Domain-specific and text-specific summary of word length (both character and syllable counts), log-lexical frequency (lemma and type) and surprisal (GPT-2 large estimated with left-hand sentence context) across texts.}
\label{fig:text-chars-long}
\end{figure}

Figure \ref{fig:text-chars-long} presents a more extensive overview over different text characteristics.

\section{Calibration and Validation Overview}\label{sec:A3}

\begin{table}[ht!]
\begin{footnotesize}
\caption{\label{tab:val-scores} Number of 9-point calibrations and validations per session and the averages of the average and the maximal validation score for each validation in each session.}
\begin{tabularx}{\textwidth}{>{\raggedright\hsize=.03\hsize}X>{\raggedright\hsize=.07\hsize}X>{\raggedright\hsize=.05\hsize}X>{\raggedright\hsize=.1\hsize}X>{\raggedright\hsize=.1\hsize}X>{\raggedright\hsize=.03\hsize}X|>{\raggedright\hsize=.03\hsize}X>{\raggedright\hsize=.07\hsize}X>{\raggedright\hsize=.05\hsize}X>{\raggedright\hsize=.1\hsize}X>{\raggedright\hsize=.1\hsize}X>{\raggedright\hsize=.03\hsize}X|>{\raggedright\hsize=.03\hsize}X>{\raggedright\hsize=.07\hsize}X>{\raggedright\hsize=.05\hsize}X>{\raggedright\hsize=.1\hsize}X>{\raggedright\hsize=.1\hsize}X>{\raggedright\hsize=.03\hsize}X}
\textbf{ID }& \textbf{Type} & \textbf{Eye} & \textbf{$\overline{avg}$} & \textbf{$\overline{max}$} & \textbf{\#} & \textbf{ID} & \textbf{Type} & \textbf{Eye} & \textbf{$\overline{avg}$} & \textbf{$\overline{max}$} & \textbf{\#} & \textbf{ID} & \textbf{Type} & \textbf{Eye} & \textbf{$\overline{avg}$} & \textbf{$\overline{max}$} & \textbf{\#}\\\toprule

\multirow{2}{0.07\columnwidth}{0} & cal & R &  &  & 16 & 
\multirow{2}{0.07\columnwidth}{34} & cal & R &  &  & 9 & 
\multirow{2}{0.07\columnwidth}{78} & cal & R &  &  & 14 \\

 & val & R & 0.364 & 0.737 & 9 & 
 & val & R & 0.422 & 0.679 & 8 & 
 & val & R & 0.307 & 0.554 & 11 \\\hline

\multirow{2}{0.07\columnwidth}{1} & cal & R &  &  & 20 &
\multirow{2}{0.07\columnwidth}{35} & cal & R &  &  & 12 & 
 \multirow{2}{0.07\columnwidth}{79} & cal & R &  &  & 14 \\

 & val & R & 0.52 & 1.536 & 7 & 
 & val & R & 0.312 & 0.538 & 8 & 
 & val & R & 0.268 & 0.601 & 8 \\\hline
 
\multirow{2}{0.07\columnwidth}{2} & cal & R &  &  & 25 &
\multirow{2}{0.07\columnwidth}{36} & cal & R &  &  & 15 & 
\multirow{2}{0.07\columnwidth}{80} & cal & R &  &  & 12 \\

 & val & R & 0.597 & 1.068 & 6 & 
 & val & R & 0.283 & 0.431 & 15 & 
 & val & R & 0.341 & 0.655 & 11 \\\hline
 
\multirow{2}{0.07\columnwidth}{3} & cal & R &  &  & 13 & 
\multirow{2}{0.07\columnwidth}{37} & cal & R &  &  & 15 & 
 \multirow{2}{0.07\columnwidth}{81} & cal & R &  &  & 13 \\

 & val & R & 0.306 & 0.65 & 7 & 
 & val & R & 0.445 & 0.828 & 10 & 
 & val & R & 0.437 & 0.78 & 13 \\\hline
 
\multirow{2}{0.07\columnwidth}{4} & cal & R &  &  & 6 &
\multirow{2}{0.07\columnwidth}{38} & cal & R &  &  & 15 & 
\multirow{2}{0.07\columnwidth}{82} & cal & R &  &  & 14 \\

 & val & R & 0.228 & 0.45 & 5 & 
 & val & R & 0.347 & 0.611 & 7 & 
 & val & R & 0.569 & 0.962 & 9 \\\hline
 
\multirow{2}{0.07\columnwidth}{5} & cal & R &  &  & 10 & 
\multirow{2}{0.07\columnwidth}{39} & cal & R &  &  & 15 & 
 \multirow{4}{0.07\columnwidth}{83} & \multirow{2}{0.07\columnwidth}{cal} & L &  &  & 19 \\

 & val & R & 0.293 & 0.677 & 7 & 
 & val & R & 0.355 & 0.681 & 11 & 
  &  & R &  &  & 3 \\\cmidrule{1-12}
 
\multirow{2}{0.07\columnwidth}{6} & cal & R &  &  & 12 & 
\multirow{2}{0.07\columnwidth}{40} & cal & R &  &  & 16 & 
& \multirow{2}{0.07\columnwidth}{val} & L & 0.882 & 1.773 & 9 \\

 & val & R & 0.289 & 0.489 & 9 & 
 & val & R & 0.463 & 0.88 & 10 & 
 &  & R & 2.37 & 5.045 & 2 \\\hline
 
\multirow{2}{0.07\columnwidth}{7} & cal & R &  &  & 10 & 
\multirow{2}{0.07\columnwidth}{41} & cal & R &  &  & 14 & 
\multirow{2}{0.07\columnwidth}{84} & cal & R &  &  & 12 \\

 & val & R & 0.251 & 0.566 & 8 & 
 & val & R & 0.343 & 0.744 & 7 & 
  & val & R & 0.334 & 0.876 & 7 \\\hline
 
\multirow{2}{0.07\columnwidth}{8} & cal & R &  &  & 7 & 
\multirow{2}{0.07\columnwidth}{60} & cal & R &  &  & 9 & 
\multirow{2}{0.07\columnwidth}{85} & cal & R &  &  & 15 \\

 & val & R & 0.287 & 0.528 & 6 & 
 & val & R & 0.475 & 1.135 & 4 & 
 & val & R & 0.364 & 0.701 & 14 \\\hline
 
\multirow{2}{0.07\columnwidth}{9} & cal & R &  &  & 17 & 
\multirow{2}{0.07\columnwidth}{61} & cal & R &  &  & 27 & 
\multirow{2}{0.07\columnwidth}{87} & cal & R &  &  & 11 \\

 & val & R & 0.579 & 1.739 & 7 & 
 & val & R & 0.518 & 0.922 & 8 & 
 & val & R & 0.274 & 0.572 & 8 \\\hline
 
\multirow{2}{0.07\columnwidth}{10} & cal & R &  &  & 9 &
\multirow{2}{0.07\columnwidth}{62} & cal & R &  &  & 22 & 
\multirow{2}{0.07\columnwidth}{90} & cal & R &  &  & 19 \\

 & val & R & 0.294 & 0.508 & 5 & 
 & val & R & 0.34 & 0.58 & 13 & 
 & val & R & 0.688 & 1.693 & 6 \\\hline
 
\multirow{2}{0.07\columnwidth}{12} & cal & R &  &  & 12 &
\multirow{2}{0.07\columnwidth}{63} & cal & R &  &  & 17 & 
\multirow{2}{0.07\columnwidth}{91} & cal & R &  &  & 18 \\

 & val & R & 0.384 & 0.754 & 5 & 
 & val & R & 0.5 & 0.817 & 9 & 
 & val & R & 0.378 & 0.637 & 6 \\\hline
 
\multirow{2}{0.07\columnwidth}{13} & cal & R &  &  & 21 &
\multirow{2}{0.07\columnwidth}{64} & cal & R &  &  & 21 & 
\multirow{2}{0.07\columnwidth}{92} & cal & R &  &  & 14 \\

 & val & R & 0.375 & 0.664 & 8 & 
 & val & R & 0.372 & 0.6 & 14 & 
 & val & R & 0.261 & 0.486 & 12 \\\hline

\multirow{2}{0.07\columnwidth}{14} & cal & R &  &  & 16 & 
\multirow{2}{0.07\columnwidth}{65} & cal & R &  &  & 16 &
 \multirow{2}{0.07\columnwidth}{93} & cal & R &  &  & 14 \\

 & val & R & 0.416 & 0.789 & 10 & 
  & val & R & 0.499 & 0.967 & 14 &
  & val & R & 0.304 & 0.549 & 8 \\\hline
 
\multirow{2}{0.07\columnwidth}{15} & cal & R &  &  & 17 & 
\multirow{2}{0.07\columnwidth}{66} & cal & R &  &  & 21 &
 \multirow{2}{0.07\columnwidth}{94} & cal & R &  &  & 10 \\

 & val & R & 0.407 & 0.981 & 14 & 
& val & R & 0.266 & 0.566 & 13 &
  & val & R & 0.443 & 0.813 & 7 \\\hline
 
\multirow{4}{0.07\columnwidth}{16} & \multirow{2}{0.07\columnwidth}{cal} & L &  &  & 2 & 
\multirow{2}{0.07\columnwidth}{67} & cal & R &  &  & 22 &
\multirow{2}{0.07\columnwidth}{95} & cal & R &  &  & 13 \\

 &  & R &  &  & 20 & 
& val & R & 0.441 & 1.301 & 9 &
 & val & R & 0.188 & 0.422 & 5 \\\cmidrule{7-18}
 
 & \multirow{2}{0.07\columnwidth}{val} & L & 0.515 & 0.92 & 2 & 
 \multirow{2}{0.07\columnwidth}{68} & cal & R &  &  & 11 &
\multirow{2}{0.07\columnwidth}{96} & cal & R &  &  & 21 \\
 
 &  & R & 0.743 & 1.425 & 13 & 
& val & R & 0.285 & 0.493 & 6 &
 & val & R & 0.278 & 0.482 & 14 \\\hline
 
\multirow{2}{0.07\columnwidth}{17} & cal & R &  &  & 14 & 
\multirow{2}{0.07\columnwidth}{69} & cal & R &  &  & 17 &
\multirow{2}{0.07\columnwidth}{97} & cal & R &  &  & 16 \\

 & val & R & 0.315 & 0.61 & 13 & 
 & val & R & 0.508 & 1.021 & 9 &
 & val & R & 0.282 & 0.515 & 12 \\\hline
 
\multirow{2}{0.07\columnwidth}{18} & cal & R &  &  & 13 & 
\multirow{2}{0.07\columnwidth}{70} & cal & R &  &  & 15 &
\multirow{2}{0.07\columnwidth}{98} & cal & R &  &  & 16 \\

 & val & R & 0.409 & 0.774 & 10 & 
  & val & R & 0.305 & 0.528 & 12 &
 & val & R & 0.454 & 0.906 & 7 \\\hline
 
\multirow{2}{0.07\columnwidth}{19} & cal & R &  &  & 14 & 
\multirow{2}{0.07\columnwidth}{71} & cal & R &  &  & 16 &
\multirow{2}{0.07\columnwidth}{99} & cal & R &  &  & 15 \\

 & val & R & 0.409 & 0.756 & 10 & 
 & val & R & 0.291 & 0.52 & 12 &
 & val & R & 0.355 & 0.725 & 6 \\\hline
 
\multirow{2}{0.07\columnwidth}{20} & cal & R &  &  & 14 & 
\multirow{2}{0.07\columnwidth}{72} & cal & R &  &  & 8 &
\multirow{2}{0.07\columnwidth}{100} & cal & R &  &  & 9 \\

 & val & R & 0.369 & 0.67 & 10 & 
& val & R & 0.267 & 0.532 & 6 & 
 & val & R & 0.412 & 0.799 & 9 \\\hline
 
\multirow{2}{0.07\columnwidth}{22} & cal & R &  &  & 16 &
\multirow{2}{0.07\columnwidth}{73} & cal & R &  &  & 8 &
\multirow{2}{0.07\columnwidth}{101} & cal & R &  &  & 17 \\

 & val & R & 0.574 & 1.351 & 9 & 
  & val & R & 0.316 & 0.616 & 7 &
 & val & R & 0.53 & 1.069 & 14 \\\hline
 
\multirow{2}{0.07\columnwidth}{23} & cal & R &  &  & 15 & 
\multirow{2}{0.07\columnwidth}{74} & cal & R &  &  & 14 & 
\multirow{2}{0.07\columnwidth}{102} & cal & R &  &  & 18 \\

 & val & R & 0.426 & 0.784 & 8 & 
 & val & R & 0.353 & 0.581 & 7 &  
 & val & R & 0.522 & 1.163 & 15 \\\hline
 
\multirow{2}{0.07\columnwidth}{30} & cal & R &  &  & 7 & 
\multirow{2}{0.07\columnwidth}{75} & cal & R &  &  & 10 & 
\multirow{2}{0.07\columnwidth}{103} & cal & R &  &  & 15 \\

 & val & R & 0.34 & 0.613 & 3 & 
 & val & R & 0.318 & 0.48 & 6 & 
 & val & R & 0.5 & 0.993 & 10 \\\hline
 
\multirow{2}{0.07\columnwidth}{31} & cal & R &  &  & 21 & 
\multirow{2}{0.07\columnwidth}{76} & cal & R &  &  & 19 & 
\multirow{2}{0.07\columnwidth}{104} & cal & R &  &  & 13 \\

 & val & R & 0.532 & 1.002 & 9 & 
 & val & R & 0.342 & 0.793 & 10 &  
 & val & R & 0.395 & 0.982 & 10 \\\hline
 
\multirow{2}{0.07\columnwidth}{32} & cal & R &  &  & 9 & 
\multirow{2}{0.07\columnwidth}{77} & cal & R &  &  & 8 &
\multirow{2}{0.07\columnwidth}{105} & cal & R &  &  & 14 \\

 & val & R & 0.267 & 0.503 & 6 &
 & val & R & 0.453 & 0.82 & 7  &  
 & val & R & 0.438 & 0.818 & 9 \\
 
\end{tabularx}%
    
\end{footnotesize}
\end{table}

\end{appendices}

\clearpage
\bibliography{PoTeC}

\begin{thebibliography}{}
\renewcommand{\doi}[1]{\url{https://doi.org/#1}}
\bibcommenthead

\bibitem [\protect \citeauthoryear {%
Ableitner%
}{%
Ableitner%
}{%
{\protect \APACyear {2014}}%
}]{%
MolBio}
\APACinsertmetastar {%
MolBio}%
\begin{APACrefauthors}%
Ableitner, O.%
\end{APACrefauthors}%
\unskip\
\newblock
\APACrefYear{2014}.
\newblock
\APACrefbtitle {{Einführung in die Molekularbiologie. Basiswissen für das
  Arbeiten im Labor}} {{Einführung in die Molekularbiologie. Basiswissen für
  das Arbeiten im Labor}}.
\newblock
\APACaddressPublisher{Wiesbaden}{Springer}.
\PrintBackRefs{\CurrentBib}

\bibitem [\protect \citeauthoryear {%
Acart\"{u}rk%
, \"{O}zkan%
, Pek\c{c}etin%
, Ormanoğlu%
\BCBL {}\ \BBA {} Kırkıcı%
}{%
Acart\"{u}rk%
\ \protect \BOthers {.}}{%
{\protect \APACyear {2023}}%
}]{%
Acartrk2023}
\APACinsertmetastar {%
Acartrk2023}%
\begin{APACrefauthors}%
Acart\"{u}rk, C.%
, \"{O}zkan, A.%
, Pek\c{c}etin, T.N.%
, Ormanoğlu, Z.%
\BCBL {} Kırkıcı, B.%
\end{APACrefauthors}%
\unskip\
\newblock
\APACrefYearMonthDay{2023}{}{}.
\newblock
{\BBOQ}\APACrefatitle {{TURead}: An eye movement dataset of {T}urkish reading}
  {{TURead}: An eye movement dataset of {T}urkish reading}.{\BBCQ}
\newblock
\APACjournalVolNumPages{Behavior Research Methods}{}{}{1-14,}
\newblock
\begin{APACrefDOI} \doi{10.3758/s13428-023-02120-6} \end{APACrefDOI}
\newblock

\newblock

\PrintBackRefs{\CurrentBib}

\bibitem [\protect \citeauthoryear {%
Ahn%
, Kelton%
, Balasubramanian%
\BCBL {}\ \BBA {} Zelinsky%
}{%
Ahn%
\ \protect \BOthers {.}}{%
{\protect \APACyear {2020}}%
}]{%
ahn2020SAT}
\APACinsertmetastar {%
ahn2020SAT}%
\begin{APACrefauthors}%
Ahn, S.%
, Kelton, C.%
, Balasubramanian, A.%
\BCBL {} Zelinsky, G.%
\end{APACrefauthors}%
\unskip\
\newblock
\APACrefYearMonthDay{2020}{}{}.
\newblock
{\BBOQ}\APACrefatitle {Towards Predicting Reading Comprehension From Gaze
  Behavior} {Towards predicting reading comprehension from gaze
  behavior}.{\BBCQ}
\newblock
 \APACrefbtitle {{{ACM} Symposium on Eye Tracking Research and Applications}.}
  {{{ACM} Symposium on Eye Tracking Research and Applications}.}
\newblock
\APACaddressPublisher{}{Association for Computing Machinery}.
\PrintBackRefs{\CurrentBib}

\bibitem [\protect \citeauthoryear {%
Barrett%
, Bingel%
, Keller%
\BCBL {}\ \BBA {} S{\o}gaard%
}{%
Barrett%
\ \protect \BOthers {.}}{%
{\protect \APACyear {2016}}%
}]{%
barrett-etal-2016-pos-tag}
\APACinsertmetastar {%
barrett-etal-2016-pos-tag}%
\begin{APACrefauthors}%
Barrett, M.%
, Bingel, J.%
, Keller, F.%
\BCBL {} S{\o}gaard, A.%
\end{APACrefauthors}%
\unskip\
\newblock
\APACrefYearMonthDay{2016}{}{}.
\newblock
{\BBOQ}\APACrefatitle {Weakly Supervised Part-of-speech Tagging Using
  Eye-tracking Data} {Weakly supervised part-of-speech tagging using
  eye-tracking data}.{\BBCQ}
\newblock
 \APACrefbtitle {{Proceedings of the 54th Annual Meeting of the Association for
  Computational Linguistics (Volume 2: Short Papers)}} {{Proceedings of the
  54th Annual Meeting of the Association for Computational Linguistics (Volume
  2: Short Papers)}}\ (\BPGS\ 579--584).
\newblock
\APACaddressPublisher{}{Association for Computational Linguistics}.
\PrintBackRefs{\CurrentBib}

\bibitem [\protect \citeauthoryear {%
Bednarik%
\ \protect \BOthers {.}}{%
Bednarik%
\ \protect \BOthers {.}}{%
{\protect \APACyear {2020}}%
}]{%
emip2020}
\APACinsertmetastar {%
emip2020}%
\begin{APACrefauthors}%
Bednarik, R.%
, Busjahn, T.%
, Gibaldi, A.%
, Ahadi, A.%
, Bielikova, M.%
, Crosby, M.%
\BDBL {}{van der Linde}, I.%
\end{APACrefauthors}%
\unskip\
\newblock
\APACrefYearMonthDay{2020}{}{}.
\newblock
{\BBOQ}\APACrefatitle {{EMIP}: The eye movements in programming dataset}
  {{EMIP}: The eye movements in programming dataset}.{\BBCQ}
\newblock
\APACjournalVolNumPages{Science of Computer Programming}{198}{}{1–11,}
\newblock
\begin{APACrefDOI} \doi{10.1016/j.scico.2020.102520} \end{APACrefDOI}
\newblock

\newblock

\PrintBackRefs{\CurrentBib}

\bibitem [\protect \citeauthoryear {%
Beinborn%
\ \BBA {} Hollenstein%
}{%
Beinborn%
\ \BBA {} Hollenstein%
}{%
{\protect \APACyear {2023}}%
}]{%
beinborn2023cognitive}
\APACinsertmetastar {%
beinborn2023cognitive}%
\begin{APACrefauthors}%
Beinborn, L.%
\BCBT {}\ \BBA {} Hollenstein, N.%
\end{APACrefauthors}%
\unskip\
\newblock
\APACrefYear{2023}.
\newblock
\APACrefbtitle {{Cognitive Plausibility in Natural Language Processing}}
  {{Cognitive Plausibility in Natural Language Processing}}.
\newblock
\APACaddressPublisher{}{Springer Nature}.
\PrintBackRefs{\CurrentBib}

\bibitem [\protect \citeauthoryear {%
Berzak%
\ \protect \BOthers {.}}{%
Berzak%
\ \protect \BOthers {.}}{%
{\protect \APACyear {2022}}%
}]{%
celer2022}
\APACinsertmetastar {%
celer2022}%
\begin{APACrefauthors}%
Berzak, Y.%
, Nakamura, C.%
, Smith, A.%
, Weng, E.%
, Katz, B.%
, Flynn, S.%
\BCBL {} Levy, R.%
\end{APACrefauthors}%
\unskip\
\newblock
\APACrefYearMonthDay{2022}{}{}.
\newblock
{\BBOQ}\APACrefatitle {{CELER: A 365-Participant Corpus of Eye Movements in L1
  and L2 English Reading}} {{CELER: A 365-Participant Corpus of Eye Movements
  in L1 and L2 English Reading}}.{\BBCQ}
\newblock
\APACjournalVolNumPages{Open Mind}{}{}{1-10,}
\newblock
\begin{APACrefDOI} \doi{10.1162/opmi_a_00054} \end{APACrefDOI}
\newblock

\newblock

\PrintBackRefs{\CurrentBib}

\bibitem [\protect \citeauthoryear {%
Biedert%
, Hees%
, Dengel%
\BCBL {}\ \BBA {} Buscher%
}{%
Biedert%
\ \protect \BOthers {.}}{%
{\protect \APACyear {2012}}%
}]{%
Biedert2012}
\APACinsertmetastar {%
Biedert2012}%
\begin{APACrefauthors}%
Biedert, R.%
, Hees, J.%
, Dengel, A.%
\BCBL {} Buscher, G.%
\end{APACrefauthors}%
\unskip\
\newblock
\APACrefYearMonthDay{2012}{}{}.
\newblock
{\BBOQ}\APACrefatitle {A robust realtime reading-skimming classifier} {A robust
  realtime reading-skimming classifier}.{\BBCQ}
\newblock
 \APACrefbtitle {Proceedings of {ETRA} '12 (2012)} {Proceedings of {ETRA} '12
  (2012)}\ (\BPGS\ 123--130).
\PrintBackRefs{\CurrentBib}

\bibitem [\protect \citeauthoryear {%
Boujard%
, Anselme%
, Cullin%
\BCBL {}\ \BBA {} Ragu\'en\`es-Nicol%
}{%
Boujard%
\ \protect \BOthers {.}}{%
{\protect \APACyear {2014}}%
}]{%
ZellMolbio}
\APACinsertmetastar {%
ZellMolbio}%
\begin{APACrefauthors}%
Boujard, D.%
, Anselme, B.%
, Cullin, C.%
\BCBL {} Ragu\'en\`es-Nicol, C.%
\end{APACrefauthors}%
\unskip\
\newblock
\APACrefYear{2014}.
\newblock
\APACrefbtitle {{Zell- und Molekularbiologie im \"Uberblick}} {{Zell- und
  Molekularbiologie im \"Uberblick}}\ (S.~Lechowski, \BTRANS{}).
\newblock
\APACaddressPublisher{Berlin}{Springer}.
\PrintBackRefs{\CurrentBib}

\bibitem [\protect \citeauthoryear {%
Brants%
\ \protect \BOthers {.}}{%
Brants%
\ \protect \BOthers {.}}{%
{\protect \APACyear {2004}}%
}]{%
tiger2004corpus}
\APACinsertmetastar {%
tiger2004corpus}%
\begin{APACrefauthors}%
Brants, S.%
, Dipper, S.%
, Eisenberg, P.%
, Hansen, S.%
, König, E.%
, Lezius, W.%
\BDBL {}Uszkoreit, H.%
\end{APACrefauthors}%
\unskip\
\newblock
\APACrefYearMonthDay{2004}{}{}.
\newblock
{\BBOQ}\APACrefatitle {{TIGER}: Linguistic Interpretation of a {G}erman Corpus}
  {{TIGER}: Linguistic interpretation of a {G}erman corpus}.{\BBCQ}
\newblock
\APACjournalVolNumPages{Journal of Language and Computation}{}{}{597-620,}
\newblock
\begin{APACrefDOI} \doi{10.1007/s11168-004-7431-3} \end{APACrefDOI}
\newblock

\newblock

\PrintBackRefs{\CurrentBib}

\bibitem [\protect \citeauthoryear {%
Cheri%
, Mishra%
\BCBL {}\ \BBA {} Bhattacharyya%
}{%
Cheri%
\ \protect \BOthers {.}}{%
{\protect \APACyear {2016}}%
}]{%
cheri-etal-2016-coreference}
\APACinsertmetastar {%
cheri-etal-2016-coreference}%
\begin{APACrefauthors}%
Cheri, J.%
, Mishra, A.%
\BCBL {} Bhattacharyya, P.%
\end{APACrefauthors}%
\unskip\
\newblock
\APACrefYearMonthDay{2016}{}{}.
\newblock
{\BBOQ}\APACrefatitle {Leveraging Annotators{'} Gaze Behaviour for Coreference
  Resolution} {Leveraging annotators{'} gaze behaviour for coreference
  resolution}.{\BBCQ}
\newblock
 \APACrefbtitle {{Proceedings of the 7th Workshop on Cognitive Aspects of
  Computational Language Learning}} {{Proceedings of the 7th Workshop on
  Cognitive Aspects of Computational Language Learning}}\ (\BPGS\ 22--26).
\newblock
\APACaddressPublisher{}{Association for Computational Linguistics}.
\PrintBackRefs{\CurrentBib}

\bibitem [\protect \citeauthoryear {%
Coltheart%
, Davelaar%
, Jonasson%
\BCBL {}\ \BBA {} Besner%
}{%
Coltheart%
\ \protect \BOthers {.}}{%
{\protect \APACyear {1977}}%
}]{%
Coltheart1977}
\APACinsertmetastar {%
Coltheart1977}%
\begin{APACrefauthors}%
Coltheart, M.%
, Davelaar, E.J.%
, Jonasson, J.T.%
\BCBL {} Besner, D.%
\end{APACrefauthors}%
\unskip\
\newblock
\APACrefYearMonthDay{1977}{}{}.
\newblock
{\BBOQ}\APACrefatitle {Access to the Internal Lexicon} {Access to the internal
  lexicon}.{\BBCQ}
\newblock
 S.~Dornic\ (\BED), \APACrefbtitle {{A}ttention and {P}erformance} {{A}ttention
  and {P}erformance}\ (\BVOL~VI, \BPGS\ 535--555).
\newblock
\APACaddressPublisher{Hillsdale}{Lawrence Erlbaum Associates}.
\PrintBackRefs{\CurrentBib}

\bibitem [\protect \citeauthoryear {%
Cop%
, Dirix%
, Drieghe%
\BCBL {}\ \BBA {} Duyck%
}{%
Cop%
\ \protect \BOthers {.}}{%
{\protect \APACyear {2016}}%
}]{%
Cop2016geco}
\APACinsertmetastar {%
Cop2016geco}%
\begin{APACrefauthors}%
Cop, U.%
, Dirix, N.%
, Drieghe, D.%
\BCBL {} Duyck, W.%
\end{APACrefauthors}%
\unskip\
\newblock
\APACrefYearMonthDay{2016}{}{}.
\newblock
{\BBOQ}\APACrefatitle {Presenting {GECO}: An eyetracking corpus of monolingual
  and bilingual sentence reading} {Presenting {GECO}: An eyetracking corpus of
  monolingual and bilingual sentence reading}.{\BBCQ}
\newblock
\APACjournalVolNumPages{Behavior Research Methods}{49}{2}{602–615,}
\newblock
\begin{APACrefDOI} \doi{10.3758/s13428-016-0734-0} \end{APACrefDOI}
\newblock

\newblock

\PrintBackRefs{\CurrentBib}

\bibitem [\protect \citeauthoryear {%
Das Digitale {W}\"orterbuch der deutschen {S}prache}{%
Das Digitale {W}\"orterbuch der deutschen {S}prache}{%
{\protect \APACyear {2016}}%
}]{%
dwds}
\APACinsertmetastar {%
dwds}%
\APACrefbtitle {Das Digitale {W}\"orterbuch der deutschen {S}prache ({DWDS})}
  {Das digitale {W}\"orterbuch der deutschen {S}prache ({DWDS})}.
\newblock
\APACrefYear{2016}.
\newblock
\APACaddressPublisher{}{Berlin-Brandenburg Academy of Science}.
\newblock
\APAChowpublished {\url{http://www.dwds.de}}.
\PrintBackRefs{\CurrentBib}

\bibitem [\protect \citeauthoryear {%
Demberg%
\ \BBA {} Keller%
}{%
Demberg%
\ \BBA {} Keller%
}{%
{\protect \APACyear {2008}}%
}]{%
DembergKeller2008}
\APACinsertmetastar {%
DembergKeller2008}%
\begin{APACrefauthors}%
Demberg, V.%
\BCBT {}\ \BBA {} Keller, F.%
\end{APACrefauthors}%
\unskip\
\newblock
\APACrefYearMonthDay{2008}{}{}.
\newblock
{\BBOQ}\APACrefatitle {Data from eyetracking corpora as evidence for theories
  of syntactic processing complexity} {Data from eyetracking corpora as
  evidence for theories of syntactic processing complexity}.{\BBCQ}
\newblock
\APACjournalVolNumPages{Cognition}{109}{}{193--210,}
\newblock
\begin{APACrefDOI} \doi{10.1016/j.cognition.2008.07.008} \end{APACrefDOI}
\newblock

\newblock

\PrintBackRefs{\CurrentBib}

\bibitem [\protect \citeauthoryear {%
Demberg%
\ \BBA {} Keller%
}{%
Demberg%
\ \BBA {} Keller%
}{%
{\protect \APACyear {2019}}%
}]{%
demberg2019cognitive}
\APACinsertmetastar {%
demberg2019cognitive}%
\begin{APACrefauthors}%
Demberg, V.%
\BCBT {}\ \BBA {} Keller, F.%
\end{APACrefauthors}%
\unskip\
\newblock
\APACrefYearMonthDay{2019}{}{}.
\newblock
{\BBOQ}\APACrefatitle {Cognitive models of syntax and sentence processing}
  {Cognitive models of syntax and sentence processing}.{\BBCQ}
\newblock
\APACjournalVolNumPages{{Human Language: From Genes and Brains to
  Behavior}}{}{}{293--312,}
\newblock
\begin{APACrefDOI} \doi{10.7551/mitpress/10841.003.0027} \end{APACrefDOI}
\newblock

\newblock

\PrintBackRefs{\CurrentBib}

\bibitem [\protect \citeauthoryear {%
Demtr\"oder%
}{%
Demtr\"oder%
}{%
{\protect \APACyear {2010}}%
}]{%
Demtroeder3}
\APACinsertmetastar {%
Demtroeder3}%
\begin{APACrefauthors}%
Demtr\"oder, W.%
\end{APACrefauthors}%
\unskip\
\newblock
\APACrefYear{2010}.
\newblock
\APACrefbtitle {{Experimentalphysik 3: Atome, Moleküle und Festkörper}}
  {{Experimentalphysik 3: Atome, Moleküle und Festkörper}}\
  (\PrintOrdinal{4th}\ \BEd).
\newblock
\APACaddressPublisher{Berlin}{Springer}.
\PrintBackRefs{\CurrentBib}

\bibitem [\protect \citeauthoryear {%
Demtr\"oder%
}{%
Demtr\"oder%
}{%
{\protect \APACyear {2014}}%
{\protect \APACexlab {{\protect \BCnt {1}}}}}]{%
Demtroeder2}
\APACinsertmetastar {%
Demtroeder2}%
\begin{APACrefauthors}%
Demtr\"oder, W.%
\end{APACrefauthors}%
\unskip\
\newblock
\APACrefYear{2014{\protect \BCnt {1}}}.
\newblock
\APACrefbtitle {{Experimentalphysik 2: Elektrizität und Optik}}
  {{Experimentalphysik 2: Elektrizität und Optik}}\ (\PrintOrdinal{6th}\
  \BEd).
\newblock
\APACaddressPublisher{Berlin}{Springer}.
\PrintBackRefs{\CurrentBib}

\bibitem [\protect \citeauthoryear {%
Demtr\"oder%
}{%
Demtr\"oder%
}{%
{\protect \APACyear {2014}}%
{\protect \APACexlab {{\protect \BCnt {2}}}}}]{%
Demtroeder4}
\APACinsertmetastar {%
Demtroeder4}%
\begin{APACrefauthors}%
Demtr\"oder, W.%
\end{APACrefauthors}%
\unskip\
\newblock
\APACrefYear{2014{\protect \BCnt {2}}}.
\newblock
\APACrefbtitle {{Experimentalphysik 4: Kern-, Teilchen- und Astrophysik}}
  {{Experimentalphysik 4: Kern-, Teilchen- und Astrophysik}}\
  (\PrintOrdinal{4th}\ \BEd).
\newblock
\APACaddressPublisher{Berlin}{Springer}.
\PrintBackRefs{\CurrentBib}

\bibitem [\protect \citeauthoryear {%
Deng%
, Prasse%
, Reich%
, Scheffer%
\BCBL {}\ \BBA {} J{\"a}ger%
}{%
Deng%
\ \protect \BOthers {.}}{%
{\protect \APACyear {2023}}%
}]{%
deng-etal-2023-pre}
\APACinsertmetastar {%
deng-etal-2023-pre}%
\begin{APACrefauthors}%
Deng, S.%
, Prasse, P.%
, Reich, D.%
, Scheffer, T.%
\BCBL {} J{\"a}ger, L.%
\end{APACrefauthors}%
\unskip\
\newblock
\APACrefYearMonthDay{2023}{}{}.
\newblock
{\BBOQ}\APACrefatitle {Pre-Trained Language Models Augmented with Synthetic
  Scanpaths for Natural Language Understanding} {Pre-trained language models
  augmented with synthetic scanpaths for natural language
  understanding}.{\BBCQ}
\newblock
 \APACrefbtitle {{Proceedings of the 2023 Conference on Empirical Methods in
  Natural Language Processing}} {{Proceedings of the 2023 Conference on
  Empirical Methods in Natural Language Processing}}\ (\BPGS\ 6500--6507).
\newblock
\APACaddressPublisher{}{Association for Computational Linguistics}.
\PrintBackRefs{\CurrentBib}

\bibitem [\protect \citeauthoryear {%
Devlin%
, Chang%
, Lee%
\BCBL {}\ \BBA {} Toutanova%
}{%
Devlin%
\ \protect \BOthers {.}}{%
{\protect \APACyear {2019}}%
}]{%
devlin-etal-2019-bert}
\APACinsertmetastar {%
devlin-etal-2019-bert}%
\begin{APACrefauthors}%
Devlin, J.%
, Chang, M\BHBI W.%
, Lee, K.%
\BCBL {} Toutanova, K.%
\end{APACrefauthors}%
\unskip\
\newblock
\APACrefYearMonthDay{2019}{}{}.
\newblock
{\BBOQ}\APACrefatitle {{BERT}: Pre-training of Deep Bidirectional Transformers
  for Language Understanding} {{BERT}: Pre-training of deep bidirectional
  transformers for language understanding}.{\BBCQ}
\newblock
 \APACrefbtitle {Proceedings of the 2019 Conference of the North {A}merican
  Chapter of the Association for Computational Linguistics: Human Language
  Technologies, Volume 1 (Long and Short Papers)} {Proceedings of the 2019
  conference of the north {A}merican chapter of the association for
  computational linguistics: Human language technologies, volume 1 (long and
  short papers)}\ (\BPGS\ 4171--4186).
\newblock
\APACaddressPublisher{}{Association for Computational Linguistics}.
\PrintBackRefs{\CurrentBib}

\bibitem [\protect \citeauthoryear {%
Frank%
\ \BBA {} Aumeistere%
}{%
Frank%
\ \BBA {} Aumeistere%
}{%
{\protect \APACyear {2023}}%
}]{%
Frank2023}
\APACinsertmetastar {%
Frank2023}%
\begin{APACrefauthors}%
Frank, S.L.%
\BCBT {}\ \BBA {} Aumeistere, A.%
\end{APACrefauthors}%
\unskip\
\newblock
\APACrefYearMonthDay{2023}{}{}.
\newblock
{\BBOQ}\APACrefatitle {An eye-tracking-with-{EEG} coregistration corpus of
  narrative sentences} {An eye-tracking-with-{EEG} coregistration corpus of
  narrative sentences}.{\BBCQ}
\newblock
\APACjournalVolNumPages{Language Resources and Evaluation}{}{}{1–17,}
\newblock
\begin{APACrefDOI} \doi{10.1007/s10579-023-09684-x} \end{APACrefDOI}
\newblock

\newblock

\PrintBackRefs{\CurrentBib}

\bibitem [\protect \citeauthoryear {%
Frank%
, Fernandez~Monsalve%
, Thompson%
\BCBL {}\ \BBA {} Vigliocco%
}{%
Frank%
\ \protect \BOthers {.}}{%
{\protect \APACyear {2013}}%
}]{%
frank2013reading}
\APACinsertmetastar {%
frank2013reading}%
\begin{APACrefauthors}%
Frank, S.L.%
, Fernandez~Monsalve, I.%
, Thompson, R.L.%
\BCBL {} Vigliocco, G.%
\end{APACrefauthors}%
\unskip\
\newblock
\APACrefYearMonthDay{2013}{}{}.
\newblock
{\BBOQ}\APACrefatitle {Reading time data for evaluating broad-coverage models
  of {English} sentence processing} {Reading time data for evaluating
  broad-coverage models of {English} sentence processing}.{\BBCQ}
\newblock
\APACjournalVolNumPages{Behavior research methods}{45}{}{1182--1190,}
\newblock
\begin{APACrefDOI} \doi{10.3758/s13428-012-0313-y} \end{APACrefDOI}
\newblock

\newblock

\PrintBackRefs{\CurrentBib}

\bibitem [\protect \citeauthoryear {%
Futrell%
\ \protect \BOthers {.}}{%
Futrell%
\ \protect \BOthers {.}}{%
{\protect \APACyear {2017}}%
}]{%
futrell2017natural-stories}
\APACinsertmetastar {%
futrell2017natural-stories}%
\begin{APACrefauthors}%
Futrell, R.%
, Gibson, E.%
, Tily, H.%
, Blank, I.%
, Vishnevetsky, A.%
, Piantadosi, S.T.%
\BCBL {} Fedorenko, E.%
\end{APACrefauthors}%
\unskip\
\newblock
\APACrefYearMonthDay{2017}{}{}.
\newblock
{\BBOQ}\APACrefatitle {The Natural Stories Corpus} {The natural stories
  corpus}.{\BBCQ}
\newblock
\APACjournalVolNumPages{CoRR}{abs/1708.05763}{}{,}
\newblock
\begin{APACrefDOI} \doi{10.48550/arXiv.1708.05763} \end{APACrefDOI}
\newblock

\newblock

\PrintBackRefs{\CurrentBib}

\bibitem [\protect \citeauthoryear {%
Goodkind%
\ \BBA {} Bicknell%
}{%
Goodkind%
\ \BBA {} Bicknell%
}{%
{\protect \APACyear {2018}}%
}]{%
goodkind-bicknell-2018-predictive}
\APACinsertmetastar {%
goodkind-bicknell-2018-predictive}%
\begin{APACrefauthors}%
Goodkind, A.%
\BCBT {}\ \BBA {} Bicknell, K.%
\end{APACrefauthors}%
\unskip\
\newblock
\APACrefYearMonthDay{2018}{}{}.
\newblock
{\BBOQ}\APACrefatitle {Predictive power of word surprisal for reading times is
  a linear function of language model quality} {Predictive power of word
  surprisal for reading times is a linear function of language model
  quality}.{\BBCQ}
\newblock
 \APACrefbtitle {P{roceedings of the 8th Workshop on Cognitive Modeling and
  Computational Linguistics ({CMCL} 2018)}} {P{roceedings of the 8th Workshop
  on Cognitive Modeling and Computational Linguistics ({CMCL} 2018)}}\ (\BPGS\
  10--18).
\newblock
\APACaddressPublisher{}{Association for Computational Linguistics}.
\PrintBackRefs{\CurrentBib}

\bibitem [\protect \citeauthoryear {%
Graw%
}{%
Graw%
}{%
{\protect \APACyear {2015}}%
}]{%
Genetik}
\APACinsertmetastar {%
Genetik}%
\begin{APACrefauthors}%
Graw, J.%
\end{APACrefauthors}%
\unskip\
\newblock
\APACrefYear{2015}.
\newblock
\APACrefbtitle {Genetik} {Genetik}\ (\PrintOrdinal{6th}\ \BEd).
\newblock
\APACaddressPublisher{Berlin}{Springer}.
\PrintBackRefs{\CurrentBib}

\bibitem [\protect \citeauthoryear {%
Griffith%
, Lohr%
, Abdulin%
\BCBL {}\ \BBA {} Komogortsev%
}{%
Griffith%
\ \protect \BOthers {.}}{%
{\protect \APACyear {2021}}%
}]{%
Griffith2021gazebase}
\APACinsertmetastar {%
Griffith2021gazebase}%
\begin{APACrefauthors}%
Griffith, H.%
, Lohr, D.%
, Abdulin, E.%
\BCBL {} Komogortsev, O.%
\end{APACrefauthors}%
\unskip\
\newblock
\APACrefYearMonthDay{2021}{}{}.
\newblock
{\BBOQ}\APACrefatitle {{GazeBase}, a large-scale, multi-stimulus, longitudinal
  eye movement dataset} {{GazeBase}, a large-scale, multi-stimulus,
  longitudinal eye movement dataset}.{\BBCQ}
\newblock
\APACjournalVolNumPages{Scientific Data}{8}{1}{1–9,}
\newblock
\begin{APACrefDOI} \doi{10.1038/s41597-021-00959-y} \end{APACrefDOI}
\newblock

\newblock

\PrintBackRefs{\CurrentBib}

\bibitem [\protect \citeauthoryear {%
Hamilton%
\ \BBA {} Huth%
}{%
Hamilton%
\ \BBA {} Huth%
}{%
{\protect \APACyear {2020}}%
}]{%
hamilton2020REvolution}
\APACinsertmetastar {%
hamilton2020REvolution}%
\begin{APACrefauthors}%
Hamilton, L.S.%
\BCBT {}\ \BBA {} Huth, A.G.%
\end{APACrefauthors}%
\unskip\
\newblock
\APACrefYearMonthDay{2020}{}{}.
\newblock
{\BBOQ}\APACrefatitle {The revolution will not be controlled: natural stimuli
  in speech neuroscience} {The revolution will not be controlled: natural
  stimuli in speech neuroscience}.{\BBCQ}
\newblock
\APACjournalVolNumPages{Language, Cognition and Neuroscience}{35}{5}{573-582,}
\newblock
\begin{APACrefDOI} \doi{10.1080/23273798.2018.1499946} \end{APACrefDOI}
\newblock

\newblock

\PrintBackRefs{\CurrentBib}

\bibitem [\protect \citeauthoryear {%
Heister%
\ \protect \BOthers {.}}{%
Heister%
\ \protect \BOthers {.}}{%
{\protect \APACyear {2011}}%
}]{%
Heister2011}
\APACinsertmetastar {%
Heister2011}%
\begin{APACrefauthors}%
Heister, J.%
, W\"urzner, K\BHBI M.%
, Bubenzer, J.%
, Pohl, E.%
, Hanneforth, T.%
, Geyken, A.%
\BCBL {} Kliegl, R.%
\end{APACrefauthors}%
\unskip\
\newblock
\APACrefYearMonthDay{2011}{}{}.
\newblock
{\BBOQ}\APACrefatitle {{dlexDB} -- eine lexikalische {D}atenbank f\"ur die
  psychologische und linguistische {F}orschung} {{dlexDB} -- eine lexikalische
  {D}atenbank f\"ur die psychologische und linguistische {F}orschung}.{\BBCQ}
\newblock
\APACjournalVolNumPages{{P}sychologische {R}undschau}{62}{1}{10--20,}
\newblock
\begin{APACrefDOI} \doi{10.1026/0033-3042/a000029} \end{APACrefDOI}
\newblock
\begin{APACrefURL} {\url{http://www.dlexdb.de/}} \end{APACrefURL}
\newblock

\newblock

\PrintBackRefs{\CurrentBib}

\bibitem [\protect \citeauthoryear {%
Hollenstein%
, Barrett%
\BCBL {}\ \BBA {} Beinborn%
}{%
Hollenstein%
, Barrett%
\BCBL {}\ \BBA {} Beinborn%
}{%
{\protect \APACyear {2020}}%
}]{%
hollenstein-etal-2020-practices}
\APACinsertmetastar {%
hollenstein-etal-2020-practices}%
\begin{APACrefauthors}%
Hollenstein, N.%
, Barrett, M.%
\BCBL {} Beinborn, L.%
\end{APACrefauthors}%
\unskip\
\newblock
\APACrefYearMonthDay{2020}{}{}.
\newblock
{\BBOQ}\APACrefatitle {Towards Best Practices for Leveraging Human Language
  Processing Signals for Natural Language Processing} {Towards best practices
  for leveraging human language processing signals for natural language
  processing}.{\BBCQ}
\newblock
 \APACrefbtitle {{Proceedings of the Second Workshop on Linguistic and
  Neurocognitive Resources}} {{Proceedings of the Second Workshop on Linguistic
  and Neurocognitive Resources}}\ (\BPGS\ 15--27).
\newblock
\APACaddressPublisher{}{European Language Resources Association}.
\PrintBackRefs{\CurrentBib}

\bibitem [\protect \citeauthoryear {%
Hollenstein%
, Barrett%
\BCBL {}\ \BBA {} Bj{\"o}rnsd{\'o}ttir%
}{%
Hollenstein%
\ \protect \BOthers {.}}{%
{\protect \APACyear {2022}}%
}]{%
hollenstein-etal-2022-copenhagen}
\APACinsertmetastar {%
hollenstein-etal-2022-copenhagen}%
\begin{APACrefauthors}%
Hollenstein, N.%
, Barrett, M.%
\BCBL {} Bj{\"o}rnsd{\'o}ttir, M.%
\end{APACrefauthors}%
\unskip\
\newblock
\APACrefYearMonthDay{2022}{}{}.
\newblock
{\BBOQ}\APACrefatitle {The Copenhagen Corpus of Eye Tracking Recordings from
  Natural Reading of {D}anish Texts} {The copenhagen corpus of eye tracking
  recordings from natural reading of {D}anish texts}.{\BBCQ}
\newblock
 \APACrefbtitle {{Proceedings of the Thirteenth Language Resources and
  Evaluation Conference}} {{Proceedings of the Thirteenth Language Resources
  and Evaluation Conference}}\ (\BPGS\ 1712--1720).
\newblock
\APACaddressPublisher{}{European Language Resources Association}.
\PrintBackRefs{\CurrentBib}

\bibitem [\protect \citeauthoryear {%
Hollenstein%
, Pirovano%
, Zhang%
, J{\"a}ger%
\BCBL {}\ \BBA {} Beinborn%
}{%
Hollenstein%
\ \protect \BOthers {.}}{%
{\protect \APACyear {2021}}%
}]{%
hollenstein2021}
\APACinsertmetastar {%
hollenstein2021}%
\begin{APACrefauthors}%
Hollenstein, N.%
, Pirovano, F.%
, Zhang, C.%
, J{\"a}ger, L.A.%
\BCBL {} Beinborn, L.%
\end{APACrefauthors}%
\unskip\
\newblock
\APACrefYearMonthDay{2021}{}{}.
\newblock
{\BBOQ}\APACrefatitle {Multilingual Language Models Predict Human Reading
  Behavior} {Multilingual language models predict human reading
  behavior}.{\BBCQ}
\newblock
 \APACrefbtitle {{Proceedings of the 2021 Conference of the North American
  Chapter of the Association for Computational Linguistics: Human Language
  Technologies}} {{Proceedings of the 2021 Conference of the North American
  Chapter of the Association for Computational Linguistics: Human Language
  Technologies}}\ (\BPG~106–123).
\newblock
\APACaddressPublisher{}{Association for Computational Linguistics}.
\PrintBackRefs{\CurrentBib}

\bibitem [\protect \citeauthoryear {%
Hollenstein%
\ \protect \BOthers {.}}{%
Hollenstein%
\ \protect \BOthers {.}}{%
{\protect \APACyear {2018}}%
}]{%
Hollenstein2018zuco1}
\APACinsertmetastar {%
Hollenstein2018zuco1}%
\begin{APACrefauthors}%
Hollenstein, N.%
, Rotsztejn, J.%
, Troendle, M.%
, Pedroni, A.%
, Zhang, C.%
\BCBL {} Langer, N.%
\end{APACrefauthors}%
\unskip\
\newblock
\APACrefYearMonthDay{2018}{}{}.
\newblock
{\BBOQ}\APACrefatitle {{ZuCo}, a simultaneous {EEG} and eye-tracking resource
  for natural sentence reading} {{ZuCo}, a simultaneous {EEG} and eye-tracking
  resource for natural sentence reading}.{\BBCQ}
\newblock
\APACjournalVolNumPages{Scientific Data}{5}{1}{1–13,}
\newblock
\begin{APACrefDOI} \doi{10.1038/sdata.2018.291} \end{APACrefDOI}
\newblock

\newblock

\PrintBackRefs{\CurrentBib}

\bibitem [\protect \citeauthoryear {%
Hollenstein%
, Troendle%
, Zhang%
\BCBL {}\ \BBA {} Langer%
}{%
Hollenstein%
, Troendle%
\BCBL {}\ \protect \BOthers {.}}{%
{\protect \APACyear {2020}}%
}]{%
hollenstein-etal-2020-zuco}
\APACinsertmetastar {%
hollenstein-etal-2020-zuco}%
\begin{APACrefauthors}%
Hollenstein, N.%
, Troendle, M.%
, Zhang, C.%
\BCBL {} Langer, N.%
\end{APACrefauthors}%
\unskip\
\newblock
\APACrefYearMonthDay{2020}{}{}.
\newblock
{\BBOQ}\APACrefatitle {{Z}u{C}o 2.0: A Dataset of Physiological Recordings
  During Natural Reading and Annotation} {{Z}u{C}o 2.0: A dataset of
  physiological recordings during natural reading and annotation}.{\BBCQ}
\newblock
 \APACrefbtitle {{Proceedings of the Twelfth Language Resources and Evaluation
  Conference}} {{Proceedings of the Twelfth Language Resources and Evaluation
  Conference}}\ (\BPGS\ 138--146).
\newblock
\APACaddressPublisher{}{European Language Resources Association}.
\PrintBackRefs{\CurrentBib}

\bibitem [\protect \citeauthoryear {%
Hollenstein%
\ \BBA {} Zhang%
}{%
Hollenstein%
\ \BBA {} Zhang%
}{%
{\protect \APACyear {2019}}%
}]{%
hollenstein-zhang-2019-entity}
\APACinsertmetastar {%
hollenstein-zhang-2019-entity}%
\begin{APACrefauthors}%
Hollenstein, N.%
\BCBT {}\ \BBA {} Zhang, C.%
\end{APACrefauthors}%
\unskip\
\newblock
\APACrefYearMonthDay{2019}{}{}.
\newblock
{\BBOQ}\APACrefatitle {Entity Recognition at First Sight: {I}mproving {NER}
  with Eye Movement Information} {Entity recognition at first sight:
  {I}mproving {NER} with eye movement information}.{\BBCQ}
\newblock
 \APACrefbtitle {{Proceedings of the 2019 Conference of the North {A}merican
  Chapter of the Association for Computational Linguistics: Human Language
  Technologies, Volume 1 (Long and Short Papers)}} {{Proceedings of the 2019
  Conference of the North {A}merican Chapter of the Association for
  Computational Linguistics: Human Language Technologies, Volume 1 (Long and
  Short Papers)}}\ (\BPGS\ 1--10).
\newblock
\APACaddressPublisher{}{Association for Computational Linguistics}.
\PrintBackRefs{\CurrentBib}

\bibitem [\protect \citeauthoryear {%
Husain%
, Vasishth%
\BCBL {}\ \BBA {} Srinivasan%
}{%
Husain%
\ \protect \BOthers {.}}{%
{\protect \APACyear {2014}}%
}]{%
Husain-Vasishth-Srinivasan-2014-hindi}
\APACinsertmetastar {%
Husain-Vasishth-Srinivasan-2014-hindi}%
\begin{APACrefauthors}%
Husain, S.%
, Vasishth, S.%
\BCBL {} Srinivasan, N.%
\end{APACrefauthors}%
\unskip\
\newblock
\APACrefYearMonthDay{2014}{}{}.
\newblock
{\BBOQ}\APACrefatitle {Integration and prediction difficulty in Hindi sentence
  comprehension: Evidence from an eye-tracking corpus} {Integration and
  prediction difficulty in hindi sentence comprehension: Evidence from an
  eye-tracking corpus}.{\BBCQ}
\newblock
\APACjournalVolNumPages{Journal of Eye Movement Research}{8}{2}{1–12,}
\newblock
\begin{APACrefDOI} \doi{10.16910/jemr.8.2.3} \end{APACrefDOI}
\newblock

\newblock

\PrintBackRefs{\CurrentBib}

\bibitem [\protect \citeauthoryear {%
Joshi%
, Mishra%
, Senthamilselvan%
\BCBL {}\ \BBA {} Bhattacharyya%
}{%
Joshi%
\ \protect \BOthers {.}}{%
{\protect \APACyear {2014}}%
}]{%
joshi-etal-2014-measuring}
\APACinsertmetastar {%
joshi-etal-2014-measuring}%
\begin{APACrefauthors}%
Joshi, A.%
, Mishra, A.%
, Senthamilselvan, N.%
\BCBL {} Bhattacharyya, P.%
\end{APACrefauthors}%
\unskip\
\newblock
\APACrefYearMonthDay{2014}{}{}.
\newblock
{\BBOQ}\APACrefatitle {Measuring Sentiment Annotation Complexity of Text}
  {Measuring sentiment annotation complexity of text}.{\BBCQ}
\newblock
 \APACrefbtitle {{Proceedings of the 52nd Annual Meeting of the Association for
  Computational Linguistics (Volume 2: Short Papers)}} {{Proceedings of the
  52nd Annual Meeting of the Association for Computational Linguistics (Volume
  2: Short Papers)}}\ (\BPGS\ 36--41).
\newblock
\APACaddressPublisher{}{Association for Computational Linguistics}.
\PrintBackRefs{\CurrentBib}

\bibitem [\protect \citeauthoryear {%
Keller%
}{%
Keller%
}{%
{\protect \APACyear {2010}}%
}]{%
keller-2010-cogn-plausibility}
\APACinsertmetastar {%
keller-2010-cogn-plausibility}%
\begin{APACrefauthors}%
Keller, F.%
\end{APACrefauthors}%
\unskip\
\newblock
\APACrefYearMonthDay{2010}{}{}.
\newblock
{\BBOQ}\APACrefatitle {Cognitively Plausible Models of Human Language
  Processing} {Cognitively plausible models of human language
  processing}.{\BBCQ}
\newblock
 \APACrefbtitle {{Proceedings of the ACL 2010 Conference Short Papers}}
  {{Proceedings of the ACL 2010 Conference Short Papers}}\ (\BPGS\ 60--67).
\newblock
\APACaddressPublisher{}{Association for Computational Linguistics}.
\PrintBackRefs{\CurrentBib}

\bibitem [\protect \citeauthoryear {%
Kennedy%
, Hill%
\BCBL {}\ \BBA {} Pynte%
}{%
Kennedy%
\ \protect \BOthers {.}}{%
{\protect \APACyear {2003}}%
}]{%
dundee}
\APACinsertmetastar {%
dundee}%
\begin{APACrefauthors}%
Kennedy, A.%
, Hill, R.%
\BCBL {} Pynte, J.%
\end{APACrefauthors}%
\unskip\
\newblock
\APACrefYearMonthDay{2003}{}{}.
\newblock
\APACrefbtitle {{The Dundee Corpus}.} {{The Dundee Corpus}.}
\newblock
\APAChowpublished {Proceedings of the 12th European conference on eye
  movement}.
\PrintBackRefs{\CurrentBib}

\bibitem [\protect \citeauthoryear {%
Kennedy%
, Pynte%
, Murray%
\BCBL {}\ \BBA {} Paul%
}{%
Kennedy%
\ \protect \BOthers {.}}{%
{\protect \APACyear {2013}}%
}]{%
kennedy2013frequency}
\APACinsertmetastar {%
kennedy2013frequency}%
\begin{APACrefauthors}%
Kennedy, A.%
, Pynte, J.%
, Murray, W.S.%
\BCBL {} Paul, S\BHBI A.%
\end{APACrefauthors}%
\unskip\
\newblock
\APACrefYearMonthDay{2013}{}{}.
\newblock
{\BBOQ}\APACrefatitle {Frequency and predictability effects in the {Dundee}
  {Corpus}: An eye movement analysis} {Frequency and predictability effects in
  the {Dundee} {Corpus}: An eye movement analysis}.{\BBCQ}
\newblock
\APACjournalVolNumPages{Quarterly Journal of Experimental
  Psychology}{66}{3}{601--618,}
\newblock
\begin{APACrefDOI} \doi{10.1080/17470218.2012.676054} \end{APACrefDOI}
\newblock

\newblock

\PrintBackRefs{\CurrentBib}

\bibitem [\protect \citeauthoryear {%
Kitaev%
, Cao%
\BCBL {}\ \BBA {} Klein%
}{%
Kitaev%
\ \protect \BOthers {.}}{%
{\protect \APACyear {2019}}%
}]{%
kitaev-etal-2019-multilingual}
\APACinsertmetastar {%
kitaev-etal-2019-multilingual}%
\begin{APACrefauthors}%
Kitaev, N.%
, Cao, S.%
\BCBL {} Klein, D.%
\end{APACrefauthors}%
\unskip\
\newblock
\APACrefYearMonthDay{2019}{}{}.
\newblock
{\BBOQ}\APACrefatitle {Multilingual Constituency Parsing with Self-Attention
  and Pre-Training} {Multilingual constituency parsing with self-attention and
  pre-training}.{\BBCQ}
\newblock
 \APACrefbtitle {{Proceedings of the 57th Annual Meeting of the Association for
  Computational Linguistics}} {{Proceedings of the 57th Annual Meeting of the
  Association for Computational Linguistics}}\ (\BPGS\ 3499--3505).
\newblock
\APACaddressPublisher{}{Association for Computational Linguistics}.
\PrintBackRefs{\CurrentBib}

\bibitem [\protect \citeauthoryear {%
Kitaev%
\ \BBA {} Klein%
}{%
Kitaev%
\ \BBA {} Klein%
}{%
{\protect \APACyear {2018}}%
}]{%
kitaev-klein-2018-constituency}
\APACinsertmetastar {%
kitaev-klein-2018-constituency}%
\begin{APACrefauthors}%
Kitaev, N.%
\BCBT {}\ \BBA {} Klein, D.%
\end{APACrefauthors}%
\unskip\
\newblock
\APACrefYearMonthDay{2018}{}{}.
\newblock
{\BBOQ}\APACrefatitle {Constituency Parsing with a Self-Attentive Encoder}
  {Constituency parsing with a self-attentive encoder}.{\BBCQ}
\newblock
 \APACrefbtitle {{Proceedings of the 56th Annual Meeting of the Association for
  Computational Linguistics (Volume 1: Long Papers)}} {{Proceedings of the 56th
  Annual Meeting of the Association for Computational Linguistics (Volume 1:
  Long Papers)}}\ (\BPGS\ 2676--2686).
\newblock
\APACaddressPublisher{}{Association for Computational Linguistics}.
\PrintBackRefs{\CurrentBib}

\bibitem [\protect \citeauthoryear {%
Kliegl%
, Grabner%
, Rolfs%
\BCBL {}\ \BBA {} Engbert%
}{%
Kliegl%
\ \protect \BOthers {.}}{%
{\protect \APACyear {2004}}%
}]{%
kliegl2004psc}
\APACinsertmetastar {%
kliegl2004psc}%
\begin{APACrefauthors}%
Kliegl, R.%
, Grabner, E.%
, Rolfs, M.%
\BCBL {} Engbert, R.%
\end{APACrefauthors}%
\unskip\
\newblock
\APACrefYearMonthDay{2004}{}{}.
\newblock
{\BBOQ}\APACrefatitle {Length, frequency, and predictability effects of words
  on eye movements in reading} {Length, frequency, and predictability effects
  of words on eye movements in reading}.{\BBCQ}
\newblock
\APACjournalVolNumPages{European Journal of Cognitive
  Psychology}{16}{1-2}{262-284,}
\newblock
\begin{APACrefDOI} \doi{10.1080/09541440340000213} \end{APACrefDOI}
\newblock

\newblock

\PrintBackRefs{\CurrentBib}

\bibitem [\protect \citeauthoryear {%
Kliegl%
, Nuthmann%
\BCBL {}\ \BBA {} Engbert%
}{%
Kliegl%
\ \protect \BOthers {.}}{%
{\protect \APACyear {2006}}%
}]{%
PSC}
\APACinsertmetastar {%
PSC}%
\begin{APACrefauthors}%
Kliegl, R.%
, Nuthmann, A.%
\BCBL {} Engbert, R.%
\end{APACrefauthors}%
\unskip\
\newblock
\APACrefYearMonthDay{2006}{}{}.
\newblock
{\BBOQ}\APACrefatitle {Tracking the mind during reading: {T}he influence of
  past, present, and future words on fixation durations} {Tracking the mind
  during reading: {T}he influence of past, present, and future words on
  fixation durations}.{\BBCQ}
\newblock
\APACjournalVolNumPages{Journal of Experimental Psychology:
  {G}eneral}{135}{1}{12--35,}
\newblock
\begin{APACrefDOI} \doi{10.1037/0096-3445.135.1.12} \end{APACrefDOI}
\newblock

\newblock

\PrintBackRefs{\CurrentBib}

\bibitem [\protect \citeauthoryear {%
Krakowczyk%
, Prasse%
\BCBL {}\ \protect \BOthers {.}}{%
Krakowczyk%
, Prasse%
\BCBL {}\ \protect \BOthers {.}}{%
{\protect \APACyear {2023}}%
}]{%
Krakowczyk2023}
\APACinsertmetastar {%
Krakowczyk2023}%
\begin{APACrefauthors}%
Krakowczyk, D.G.%
, Prasse, P.%
, Reich, D.R.%
, Lapuschkin, S.%
, Scheffer, T.%
\BCBL {} J\"{a}ger, L.A.%
\end{APACrefauthors}%
\unskip\
\newblock
\APACrefYearMonthDay{2023}{}{}.
\newblock
{\BBOQ}\APACrefatitle {Bridging the Gap: Gaze Events as Interpretable Concepts
  to Explain Deep Neural Sequence Models} {Bridging the gap: Gaze events as
  interpretable concepts to explain deep neural sequence models}.{\BBCQ}
\newblock
 \APACrefbtitle {{Proceedings of the 2023 Symposium on Eye Tracking Research
  and Applications}.} {{Proceedings of the 2023 Symposium on Eye Tracking
  Research and Applications}.}
\newblock
\APACaddressPublisher{}{Association for Computing Machinery}.
\PrintBackRefs{\CurrentBib}

\bibitem [\protect \citeauthoryear {%
Krakowczyk%
, Reich%
\BCBL {}\ \protect \BOthers {.}}{%
Krakowczyk%
, Reich%
\BCBL {}\ \protect \BOthers {.}}{%
{\protect \APACyear {2023}}%
}]{%
pymovements}
\APACinsertmetastar {%
pymovements}%
\begin{APACrefauthors}%
Krakowczyk, D.G.%
, Reich, D.R.%
, Chwastek, J.%
, Jakobi, D.N.%
, Prasse, P.%
, Süss, A.%
\BDBL {}Jäger, L.A.%
\end{APACrefauthors}%
\unskip\
\newblock
\APACrefYearMonthDay{2023}{}{}.
\newblock
{\BBOQ}\APACrefatitle {pymovements: A Python Package for Processing Eye
  Movement Data} {pymovements: A python package for processing eye movement
  data}.{\BBCQ}
\newblock
 \APACrefbtitle {{2023 Symposium on Eye Tracking Research and Applications}.}
  {{2023 Symposium on Eye Tracking Research and Applications}.}
\newblock
\APACaddressPublisher{}{Association for Computing Machinery}.
\PrintBackRefs{\CurrentBib}

\bibitem [\protect \citeauthoryear {%
Kuperman%
\ \protect \BOthers {.}}{%
Kuperman%
\ \protect \BOthers {.}}{%
{\protect \APACyear {2023}}%
}]{%
kuperman-2023-mecol2}
\APACinsertmetastar {%
kuperman-2023-mecol2}%
\begin{APACrefauthors}%
Kuperman, V.%
, Siegelman, N.%
, Schroeder, S.%
, Acartürk, C.%
, Alexeeva, S.%
, Amenta, S.%
\BDBL {}et al.%
\end{APACrefauthors}%
\unskip\
\newblock
\APACrefYearMonthDay{2023}{}{}.
\newblock
{\BBOQ}\APACrefatitle {Text reading in English as a second language: Evidence
  from the Multilingual Eye-Movements Corpus} {Text reading in english as a
  second language: Evidence from the multilingual eye-movements corpus}.{\BBCQ}
\newblock
\APACjournalVolNumPages{Studies in Second Language Acquisition}{45}{1}{3–37,}
\newblock
\begin{APACrefDOI} \doi{10.1017/S0272263121000954} \end{APACrefDOI}
\newblock

\newblock

\PrintBackRefs{\CurrentBib}

\bibitem [\protect \citeauthoryear {%
Laurinavichyute%
, Sekerina%
, Alexeeva%
, Bagdasaryan%
\BCBL {}\ \BBA {} Kliegl%
}{%
Laurinavichyute%
\ \protect \BOthers {.}}{%
{\protect \APACyear {2018}}%
}]{%
Laurinavichyute2018RSC}
\APACinsertmetastar {%
Laurinavichyute2018RSC}%
\begin{APACrefauthors}%
Laurinavichyute, A.K.%
, Sekerina, I.A.%
, Alexeeva, S.%
, Bagdasaryan, K.%
\BCBL {} Kliegl, R.%
\end{APACrefauthors}%
\unskip\
\newblock
\APACrefYearMonthDay{2018}{}{}.
\newblock
{\BBOQ}\APACrefatitle {Russian Sentence Corpus: Benchmark measures of eye
  movements in reading in Russian} {Russian sentence corpus: Benchmark measures
  of eye movements in reading in russian}.{\BBCQ}
\newblock
\APACjournalVolNumPages{Behavior Research Methods}{51}{3}{1161–1178,}
\newblock
\begin{APACrefDOI} \doi{10.3758/s13428-018-1051-6} \end{APACrefDOI}
\newblock

\newblock

\PrintBackRefs{\CurrentBib}

\bibitem [\protect \citeauthoryear {%
Leal%
, Lukasova%
, Carthery-Goulart%
\BCBL {}\ \BBA {} Aluísio%
}{%
Leal%
\ \protect \BOthers {.}}{%
{\protect \APACyear {2022}}%
}]{%
Leal2022rastros}
\APACinsertmetastar {%
Leal2022rastros}%
\begin{APACrefauthors}%
Leal, S.E.%
, Lukasova, K.%
, Carthery-Goulart, M.T.%
\BCBL {} Aluísio, S.M.%
\end{APACrefauthors}%
\unskip\
\newblock
\APACrefYearMonthDay{2022}{}{}.
\newblock
{\BBOQ}\APACrefatitle {{RastrOS} Project: Natural Language Processing
  contributions to the development of an eye-tracking corpus with
  predictability norms for {Brazilian} {Portuguese}} {{RastrOS} project:
  Natural language processing contributions to the development of an
  eye-tracking corpus with predictability norms for {Brazilian}
  {Portuguese}}.{\BBCQ}
\newblock
\APACjournalVolNumPages{Language Resources and Evaluation}{56}{4}{1333–1372,}
\newblock
\begin{APACrefDOI} \doi{10.1007/s10579-022-09609-0} \end{APACrefDOI}
\newblock

\newblock

\PrintBackRefs{\CurrentBib}

\bibitem [\protect \citeauthoryear {%
Levenshtein%
}{%
Levenshtein%
}{%
{\protect \APACyear {1966}}%
}]{%
Levenshtein1966}
\APACinsertmetastar {%
Levenshtein1966}%
\begin{APACrefauthors}%
Levenshtein, V.I.%
\end{APACrefauthors}%
\unskip\
\newblock
\APACrefYearMonthDay{1966}{}{}.
\newblock
{\BBOQ}\APACrefatitle {Binary Codes Capable of Correcting Deletions,
  Insertions, and Reversals} {Binary codes capable of correcting deletions,
  insertions, and reversals}.{\BBCQ}
\newblock
\APACjournalVolNumPages{{Soviet Physics-Doklady}}{10}{8}{845--848,}
\newblock

\newblock

\PrintBackRefs{\CurrentBib}

\bibitem [\protect \citeauthoryear {%
Li%
\ \protect \BOthers {.}}{%
Li%
\ \protect \BOthers {.}}{%
{\protect \APACyear {2018}}%
}]{%
li2018reading-attention}
\APACinsertmetastar {%
li2018reading-attention}%
\begin{APACrefauthors}%
Li, X.%
, Liu, Y.%
, Mao, J.%
, He, Z.%
, Zhang, M.%
\BCBL {} Ma, S.%
\end{APACrefauthors}%
\unskip\
\newblock
\APACrefYearMonthDay{2018}{}{}.
\newblock
{\BBOQ}\APACrefatitle {Understanding Reading Attention Distribution during
  Relevance Judgement} {Understanding reading attention distribution during
  relevance judgement}.{\BBCQ}
\newblock
 \APACrefbtitle {{Proceedings of the 27th ACM International Conference on
  Information and Knowledge Management}} {{Proceedings of the 27th ACM
  International Conference on Information and Knowledge Management}}\
  (\BPG~733–742).
\newblock
\APACaddressPublisher{}{Association for Computing Machinery}.
\PrintBackRefs{\CurrentBib}

\bibitem [\protect \citeauthoryear {%
Long%
, Lu%
, Xiang%
, Li%
\BCBL {}\ \BBA {} Huang%
}{%
Long%
\ \protect \BOthers {.}}{%
{\protect \APACyear {2017}}%
}]{%
long-etal-2017-sentiment}
\APACinsertmetastar {%
long-etal-2017-sentiment}%
\begin{APACrefauthors}%
Long, Y.%
, Lu, Q.%
, Xiang, R.%
, Li, M.%
\BCBL {} Huang, C\BHBI R.%
\end{APACrefauthors}%
\unskip\
\newblock
\APACrefYearMonthDay{2017}{}{}.
\newblock
{\BBOQ}\APACrefatitle {A Cognition Based Attention Model for Sentiment
  Analysis} {A cognition based attention model for sentiment analysis}.{\BBCQ}
\newblock
 \APACrefbtitle {{Proceedings of the 2017 Conference on Empirical Methods in
  Natural Language Processing}} {{Proceedings of the 2017 Conference on
  Empirical Methods in Natural Language Processing}}\ (\BPGS\ 462--471).
\newblock
\APACaddressPublisher{}{Association for Computational Linguistics}.
\PrintBackRefs{\CurrentBib}

\bibitem [\protect \citeauthoryear {%
Luke%
\ \BBA {} Christianson%
}{%
Luke%
\ \BBA {} Christianson%
}{%
{\protect \APACyear {2017}}%
}]{%
Luke2017provo}
\APACinsertmetastar {%
Luke2017provo}%
\begin{APACrefauthors}%
Luke, S.G.%
\BCBT {}\ \BBA {} Christianson, K.%
\end{APACrefauthors}%
\unskip\
\newblock
\APACrefYearMonthDay{2017}{}{}.
\newblock
{\BBOQ}\APACrefatitle {The {Provo} Corpus: A large eye-tracking corpus with
  predictability norms} {The {Provo} corpus: A large eye-tracking corpus with
  predictability norms}.{\BBCQ}
\newblock
\APACjournalVolNumPages{Behavior Research Methods}{50}{2}{826–833,}
\newblock
\begin{APACrefDOI} \doi{10.3758/s13428-017-0908-4} \end{APACrefDOI}
\newblock

\newblock

\PrintBackRefs{\CurrentBib}

\bibitem [\protect \citeauthoryear {%
Mak%
\ \BBA {} Willems%
}{%
Mak%
\ \BBA {} Willems%
}{%
{\protect \APACyear {2019}}%
}]{%
mak2019mentalsimulation}
\APACinsertmetastar {%
mak2019mentalsimulation}%
\begin{APACrefauthors}%
Mak, M.%
\BCBT {}\ \BBA {} Willems, R.M.%
\end{APACrefauthors}%
\unskip\
\newblock
\APACrefYearMonthDay{2019}{}{}.
\newblock
{\BBOQ}\APACrefatitle {Mental simulation during literary reading: Individual
  differences revealed with eye-tracking} {Mental simulation during literary
  reading: Individual differences revealed with eye-tracking}.{\BBCQ}
\newblock
\APACjournalVolNumPages{Language, Cognition and Neuroscience}{34}{4}{511-535,}
\newblock
\begin{APACrefDOI} \doi{10.1080/23273798.2018.1552007} \end{APACrefDOI}
\newblock

\newblock

\PrintBackRefs{\CurrentBib}

\bibitem [\protect \citeauthoryear {%
Makowski%
, J\"ager%
, Abdelwahab%
, Landwehr%
\BCBL {}\ \BBA {} Scheffer%
}{%
Makowski%
\ \protect \BOthers {.}}{%
{\protect \APACyear {2019}}%
}]{%
MakowskiECML2018}
\APACinsertmetastar {%
MakowskiECML2018}%
\begin{APACrefauthors}%
Makowski, S.%
, J\"ager, L.A.%
, Abdelwahab, A.%
, Landwehr, N.%
\BCBL {} Scheffer, T.%
\end{APACrefauthors}%
\unskip\
\newblock
\APACrefYearMonthDay{2019}{}{}.
\newblock
{\BBOQ}\APACrefatitle {A discriminative model for identifying readers and
  assessing text comprehension from eye movements} {A discriminative model for
  identifying readers and assessing text comprehension from eye
  movements}.{\BBCQ}
\newblock
 \APACrefbtitle {{Machine Learning and Knowledge Discovery in Databases. ECML
  PKDD 2018}} {{Machine Learning and Knowledge Discovery in Databases. ECML
  PKDD 2018}}\ (\BVOL\ 11051, \BPGS\ 209--225).
\newblock
\APACaddressPublisher{}{Springer Nature}.
\PrintBackRefs{\CurrentBib}

\bibitem [\protect \citeauthoryear {%
Malmaud%
, Levy%
\BCBL {}\ \BBA {} Berzak%
}{%
Malmaud%
\ \protect \BOthers {.}}{%
{\protect \APACyear {2020}}%
}]{%
malmaud-etal-2020-bridging}
\APACinsertmetastar {%
malmaud-etal-2020-bridging}%
\begin{APACrefauthors}%
Malmaud, J.%
, Levy, R.%
\BCBL {} Berzak, Y.%
\end{APACrefauthors}%
\unskip\
\newblock
\APACrefYearMonthDay{2020}{}{}.
\newblock
{\BBOQ}\APACrefatitle {Bridging Information-Seeking Human Gaze and Machine
  Reading Comprehension} {Bridging information-seeking human gaze and machine
  reading comprehension}.{\BBCQ}
\newblock
 \APACrefbtitle {{Proceedings of the 24th Conference on Computational Natural
  Language Learning}} {{Proceedings of the 24th Conference on Computational
  Natural Language Learning}}\ (\BPGS\ 142--152).
\newblock
\APACaddressPublisher{}{Association for Computational Linguistics}.
\PrintBackRefs{\CurrentBib}

\bibitem [\protect \citeauthoryear {%
Mathias%
\ \protect \BOthers {.}}{%
Mathias%
\ \protect \BOthers {.}}{%
{\protect \APACyear {2018}}%
}]{%
mathias-etal-2018-eyes}
\APACinsertmetastar {%
mathias-etal-2018-eyes}%
\begin{APACrefauthors}%
Mathias, S.%
, Kanojia, D.%
, Patel, K.%
, Agrawal, S.%
, Mishra, A.%
\BCBL {} Bhattacharyya, P.%
\end{APACrefauthors}%
\unskip\
\newblock
\APACrefYearMonthDay{2018}{}{}.
\newblock
{\BBOQ}\APACrefatitle {Eyes are the Windows to the Soul: Predicting the Rating
  of Text Quality Using Gaze Behaviour} {Eyes are the windows to the soul:
  Predicting the rating of text quality using gaze behaviour}.{\BBCQ}
\newblock
 \APACrefbtitle {{Proceedings of the 56th Annual Meeting of the Association for
  Computational Linguistics (Volume 1: Long Papers)}} {{Proceedings of the 56th
  Annual Meeting of the Association for Computational Linguistics (Volume 1:
  Long Papers)}}\ (\BPGS\ 2352--2362).
\newblock
\APACaddressPublisher{}{Association for Computational Linguistics}.
\PrintBackRefs{\CurrentBib}

\bibitem [\protect \citeauthoryear {%
Mathias%
, Murthy%
, Kanojia%
, Mishra%
\BCBL {}\ \BBA {} Bhattacharyya%
}{%
Mathias%
\ \protect \BOthers {.}}{%
{\protect \APACyear {2020}}%
}]{%
mathias-etal-2020-happy}
\APACinsertmetastar {%
mathias-etal-2020-happy}%
\begin{APACrefauthors}%
Mathias, S.%
, Murthy, R.%
, Kanojia, D.%
, Mishra, A.%
\BCBL {} Bhattacharyya, P.%
\end{APACrefauthors}%
\unskip\
\newblock
\APACrefYearMonthDay{2020}{}{}.
\newblock
{\BBOQ}\APACrefatitle {Happy Are Those Who Grade without Seeing: A Multi-Task
  Learning Approach to Grade Essays Using Gaze Behaviour} {Happy are those who
  grade without seeing: A multi-task learning approach to grade essays using
  gaze behaviour}.{\BBCQ}
\newblock
 \APACrefbtitle {{Proceedings of the 1st Conference of the Asia-Pacific Chapter
  of the Association for Computational Linguistics and the 10th International
  Joint Conference on Natural Language Processing}} {{Proceedings of the 1st
  Conference of the Asia-Pacific Chapter of the Association for Computational
  Linguistics and the 10th International Joint Conference on Natural Language
  Processing}}\ (\BPGS\ 858--872).
\newblock
\APACaddressPublisher{}{Association for Computational Linguistics}.
\PrintBackRefs{\CurrentBib}

\bibitem [\protect \citeauthoryear {%
Minixhofer%
}{%
Minixhofer%
}{%
{\protect \APACyear {2020}}%
}]{%
Minixhofer_GerPT2_German_large_2020}
\APACinsertmetastar {%
Minixhofer_GerPT2_German_large_2020}%
\begin{APACrefauthors}%
Minixhofer, B.%
\end{APACrefauthors}%
\unskip\
\newblock
\APACrefYearMonthDay{2020}{}{}.
\newblock
\APACrefbtitle {{GerPT2: German large and small versions of GPT2}.} {{GerPT2:
  German large and small versions of GPT2}.}
\newblock
\begin{APACrefURL} {\url{https://github.com/bminixhofer/gerpt2}}
  \end{APACrefURL}
\PrintBackRefs{\CurrentBib}

\bibitem [\protect \citeauthoryear {%
Mishra%
, Kanojia%
\BCBL {}\ \BBA {} Bhattacharyya%
}{%
Mishra%
\ \protect \BOthers {.}}{%
{\protect \APACyear {2016}}%
}]{%
Mishra-Kanojia-Bhattacharyya-2016}
\APACinsertmetastar {%
Mishra-Kanojia-Bhattacharyya-2016}%
\begin{APACrefauthors}%
Mishra, A.%
, Kanojia, D.%
\BCBL {} Bhattacharyya, P.%
\end{APACrefauthors}%
\unskip\
\newblock
\APACrefYearMonthDay{2016}{}{}.
\newblock
{\BBOQ}\APACrefatitle {Predicting Readers’ Sarcasm Understandability by
  Modeling Gaze Behavior} {Predicting readers’ sarcasm understandability by
  modeling gaze behavior}.{\BBCQ}
\newblock
\APACjournalVolNumPages{Proceedings of the AAAI Conference on Artificial
  Intelligence}{30}{1}{1–7,}
\newblock
\begin{APACrefDOI} \doi{10.1609/aaai.v30i1.9884} \end{APACrefDOI}
\newblock

\newblock

\PrintBackRefs{\CurrentBib}

\bibitem [\protect \citeauthoryear {%
Mishra%
, Kanojia%
, Nagar%
, Dey%
\BCBL {}\ \BBA {} Bhattacharyya%
}{%
Mishra%
\ \protect \BOthers {.}}{%
{\protect \APACyear {2017}}%
{\protect \APACexlab {{\protect \BCnt {1}}}}}]{%
mishra-2017-sentiment}
\APACinsertmetastar {%
mishra-2017-sentiment}%
\begin{APACrefauthors}%
Mishra, A.%
, Kanojia, D.%
, Nagar, S.%
, Dey, K.%
\BCBL {} Bhattacharyya, P.%
\end{APACrefauthors}%
\unskip\
\newblock
\APACrefYearMonthDay{2017{\protect \BCnt {1}}}{}{}.
\newblock
{\BBOQ}\APACrefatitle {Leveraging Cognitive Features for Sentiment Analysis}
  {Leveraging cognitive features for sentiment analysis}.{\BBCQ}
\newblock
\APACjournalVolNumPages{CoRR}{abs/1701.05581}{}{156–166,}
\newblock
\begin{APACrefDOI} \doi{10.18653/v1/K16-1016} \end{APACrefDOI}
\newblock
{\href{https://arxiv.org/abs/1701.05581}{{1701.05581}}}
\newblock

\PrintBackRefs{\CurrentBib}

\bibitem [\protect \citeauthoryear {%
Mishra%
, Kanojia%
, Nagar%
, Dey%
\BCBL {}\ \BBA {} Bhattacharyya%
}{%
Mishra%
\ \protect \BOthers {.}}{%
{\protect \APACyear {2017}}%
{\protect \APACexlab {{\protect \BCnt {2}}}}}]{%
mishra2017scanpath}
\APACinsertmetastar {%
mishra2017scanpath}%
\begin{APACrefauthors}%
Mishra, A.%
, Kanojia, D.%
, Nagar, S.%
, Dey, K.%
\BCBL {} Bhattacharyya, P.%
\end{APACrefauthors}%
\unskip\
\newblock
\APACrefYearMonthDay{2017{\protect \BCnt {2}}}{}{}.
\newblock
{\BBOQ}\APACrefatitle {Scanpath Complexity: Modeling Reading Effort Using Gaze
  Information} {Scanpath complexity: Modeling reading effort using gaze
  information}.{\BBCQ}
\newblock
\APACjournalVolNumPages{Proceedings of the AAAI Conference on Artificial
  Intelligence}{31}{1}{1–8,}
\newblock
\begin{APACrefDOI} \doi{10.1609/aaai.v31i1.11159} \end{APACrefDOI}
\newblock

\newblock

\PrintBackRefs{\CurrentBib}

\bibitem [\protect \citeauthoryear {%
Nastase%
, Goldstein%
\BCBL {}\ \BBA {} Hasson%
}{%
Nastase%
\ \protect \BOthers {.}}{%
{\protect \APACyear {2020}}%
}]{%
NASTASE2020117254}
\APACinsertmetastar {%
NASTASE2020117254}%
\begin{APACrefauthors}%
Nastase, S.A.%
, Goldstein, A.%
\BCBL {} Hasson, U.%
\end{APACrefauthors}%
\unskip\
\newblock
\APACrefYearMonthDay{2020}{}{}.
\newblock
{\BBOQ}\APACrefatitle {Keep it real: rethinking the primacy of experimental
  control in cognitive neuroscience} {Keep it real: rethinking the primacy of
  experimental control in cognitive neuroscience}.{\BBCQ}
\newblock
\APACjournalVolNumPages{NeuroImage}{222}{}{117254,}
\newblock
\begin{APACrefDOI} \doi{10.1016/j.neuroimage.2020.117254} \end{APACrefDOI}
\newblock

\newblock

\PrintBackRefs{\CurrentBib}

\bibitem [\protect \citeauthoryear {%
Notaro%
\ \BBA {} Diamond%
}{%
Notaro%
\ \BBA {} Diamond%
}{%
{\protect \APACyear {2018}}%
}]{%
notaro2018duolingo}
\APACinsertmetastar {%
notaro2018duolingo}%
\begin{APACrefauthors}%
Notaro, G.M.%
\BCBT {}\ \BBA {} Diamond, S.G.%
\end{APACrefauthors}%
\unskip\
\newblock
\APACrefYearMonthDay{2018}{}{}.
\newblock
{\BBOQ}\APACrefatitle {Simultaneous {EEG}, eye-tracking, behavioral, and
  screen-capture data during online {German} language learning} {Simultaneous
  {EEG}, eye-tracking, behavioral, and screen-capture data during online
  {German} language learning}.{\BBCQ}
\newblock
\APACjournalVolNumPages{Data in Brief}{21}{}{1937-1943,}
\newblock
\begin{APACrefDOI} \doi{10.1016/j.dib.2018.11.044} \end{APACrefDOI}
\newblock

\newblock

\PrintBackRefs{\CurrentBib}

\bibitem [\protect \citeauthoryear {%
Obaidellah%
, Al~Haek%
\BCBL {}\ \BBA {} Cheng%
}{%
Obaidellah%
\ \protect \BOthers {.}}{%
{\protect \APACyear {2018}}%
}]{%
Obaidellah-2019-code-reading}
\APACinsertmetastar {%
Obaidellah-2019-code-reading}%
\begin{APACrefauthors}%
Obaidellah, U.%
, Al~Haek, M.%
\BCBL {} Cheng, P.C\BHBI H.%
\end{APACrefauthors}%
\unskip\
\newblock
\APACrefYearMonthDay{2018}{}{}.
\newblock
{\BBOQ}\APACrefatitle {A Survey on the Usage of Eye-Tracking in Computer
  Programming} {A survey on the usage of eye-tracking in computer
  programming}.{\BBCQ}
\newblock
\APACjournalVolNumPages{ACM Computing Surveys}{51}{1}{1–58,}
\newblock
\begin{APACrefDOI} \doi{10.1145/3145904} \end{APACrefDOI}
\newblock

\newblock

\PrintBackRefs{\CurrentBib}

\bibitem [\protect \citeauthoryear {%
Pan%
, Yan%
, Richter%
, Shu%
\BCBL {}\ \BBA {} Kliegl%
}{%
Pan%
\ \protect \BOthers {.}}{%
{\protect \APACyear {2021}}%
}]{%
Pan2021BSC}
\APACinsertmetastar {%
Pan2021BSC}%
\begin{APACrefauthors}%
Pan, J.%
, Yan, M.%
, Richter, E.M.%
, Shu, H.%
\BCBL {} Kliegl, R.%
\end{APACrefauthors}%
\unskip\
\newblock
\APACrefYearMonthDay{2021}{}{}.
\newblock
{\BBOQ}\APACrefatitle {The {Beijing Sentence Corpus}: A {Chinese} sentence
  corpus with eye movement data and predictability norms} {The {Beijing
  Sentence Corpus}: A {Chinese} sentence corpus with eye movement data and
  predictability norms}.{\BBCQ}
\newblock
\APACjournalVolNumPages{Behavior Research Methods}{54}{4}{1989–2000,}
\newblock
\begin{APACrefDOI} \doi{10.3758/s13428-021-01730-2} \end{APACrefDOI}
\newblock

\newblock

\PrintBackRefs{\CurrentBib}

\bibitem [\protect \citeauthoryear {%
Parker%
, Kirkby%
\BCBL {}\ \BBA {} Slattery%
}{%
Parker%
\ \protect \BOthers {.}}{%
{\protect \APACyear {2017}}%
}]{%
parker2017passage-reading}
\APACinsertmetastar {%
parker2017passage-reading}%
\begin{APACrefauthors}%
Parker, A.J.%
, Kirkby, J.A.%
\BCBL {} Slattery, T.J.%
\end{APACrefauthors}%
\unskip\
\newblock
\APACrefYearMonthDay{2017}{}{}.
\newblock
{\BBOQ}\APACrefatitle {Predictability effects during reading in the absence of
  parafoveal preview} {Predictability effects during reading in the absence of
  parafoveal preview}.{\BBCQ}
\newblock
\APACjournalVolNumPages{Journal of Cognitive Psychology}{29}{8}{902-911,}
\newblock
\begin{APACrefDOI} \doi{10.1080/20445911.2017.1340303} \end{APACrefDOI}
\newblock

\newblock

\PrintBackRefs{\CurrentBib}

\bibitem [\protect \citeauthoryear {%
Prasse%
, Reich%
, Makowski%
, Scheffer%
\BCBL {}\ \BBA {} J\"ager%
}{%
Prasse%
\ \protect \BOthers {.}}{%
{\protect \APACyear {2023}}%
}]{%
prasse2023spygan-review}
\APACinsertmetastar {%
prasse2023spygan-review}%
\begin{APACrefauthors}%
Prasse, P.%
, Reich, D.R.%
, Makowski, S.%
, Scheffer, T.%
\BCBL {} J\"ager, L.A.%
\end{APACrefauthors}%
\unskip\
\newblock
\APACrefYearMonthDay{2023}{}{}.
\newblock
\APACrefbtitle {Evaluating the utility of synthetic eye movement data for
  pre-training neural networks.} {Evaluating the utility of synthetic eye
  movement data for pre-training neural networks.}
\newblock
\APACrefnote{Under review}
\PrintBackRefs{\CurrentBib}

\bibitem [\protect \citeauthoryear {%
Radford%
\ \protect \BOthers {.}}{%
Radford%
\ \protect \BOthers {.}}{%
{\protect \APACyear {2019}}%
}]{%
radford2019language}
\APACinsertmetastar {%
radford2019language}%
\begin{APACrefauthors}%
Radford, A.%
, Wu, J.%
, Child, R.%
, Luan, D.%
, Amodei, D.%
\BCBL {} Sutskever, I.%
\end{APACrefauthors}%
\unskip\
\newblock
\APACrefYearMonthDay{2019}{}{}.
\newblock
\APACrefbtitle {Language Models are Unsupervised Multitask Learners.} {Language
  models are unsupervised multitask learners.}
\newblock
\begin{APACrefURL} {\url{https://api.semanticscholar.org/CorpusID:160025533}}
  \end{APACrefURL}
\PrintBackRefs{\CurrentBib}

\bibitem [\protect \citeauthoryear {%
Rayner%
}{%
Rayner%
}{%
{\protect \APACyear {1998}}%
}]{%
Rayner1998}
\APACinsertmetastar {%
Rayner1998}%
\begin{APACrefauthors}%
Rayner, K.%
\end{APACrefauthors}%
\unskip\
\newblock
\APACrefYearMonthDay{1998}{}{}.
\newblock
{\BBOQ}\APACrefatitle {Eye movements in reading and information processing: 20
  years of research} {Eye movements in reading and information processing: 20
  years of research}.{\BBCQ}
\newblock
\APACjournalVolNumPages{Psychological Bulletin}{124}{3}{372--422,}
\newblock
\begin{APACrefDOI} \doi{10.1037/0033-2909.124.3.372} \end{APACrefDOI}
\newblock

\newblock

\PrintBackRefs{\CurrentBib}

\bibitem [\protect \citeauthoryear {%
Rayner%
\ \BBA {} Carroll%
}{%
Rayner%
\ \BBA {} Carroll%
}{%
{\protect \APACyear {2018}}%
}]{%
raynerCarroll2018}
\APACinsertmetastar {%
raynerCarroll2018}%
\begin{APACrefauthors}%
Rayner, K.%
\BCBT {}\ \BBA {} Carroll, P.J.%
\end{APACrefauthors}%
\unskip\
\newblock
\APACrefYearMonthDay{2018}{}{}.
\newblock
{\BBOQ}\APACrefatitle {Eye movements and reading comprehension} {Eye movements
  and reading comprehension}.{\BBCQ}
\newblock
 \APACrefbtitle {{New Methods in Reading Comprehension Research}} {{New Methods
  in Reading Comprehension Research}}\ (\BPGS\ 129--150).
\newblock
\APACaddressPublisher{}{Routledge}.
\PrintBackRefs{\CurrentBib}

\bibitem [\protect \citeauthoryear {%
Ribeiro%
, Brandl%
, Søgaard%
\BCBL {}\ \BBA {} Hollenstein%
}{%
Ribeiro%
\ \protect \BOthers {.}}{%
{\protect \APACyear {2023}}%
}]{%
ribeiro2023webqamgaze}
\APACinsertmetastar {%
ribeiro2023webqamgaze}%
\begin{APACrefauthors}%
Ribeiro, T.%
, Brandl, S.%
, Søgaard, A.%
\BCBL {} Hollenstein, N.%
\end{APACrefauthors}%
\unskip\
\newblock
\APACrefYearMonthDay{2023}{}{}.
\newblock
\APACrefbtitle {{WebQAmGaze}: A Multilingual Webcam Eye-Tracking-While-Reading
  Dataset.} {{WebQAmGaze}: A multilingual webcam eye-tracking-while-reading
  dataset.}
\newblock
\APACrefnote{arXiv preprint}
\PrintBackRefs{\CurrentBib}

\bibitem [\protect \citeauthoryear {%
Schiller%
, Teufel%
\BCBL {}\ \BBA {} St\"ockert%
}{%
Schiller%
\ \protect \BOthers {.}}{%
{\protect \APACyear {1999}}%
}]{%
STTSguidelines}
\APACinsertmetastar {%
STTSguidelines}%
\begin{APACrefauthors}%
Schiller, A.%
, Teufel, S.%
\BCBL {} St\"ockert, C.%
\end{APACrefauthors}%
\unskip\
\newblock
\APACrefYearMonthDay{1999}{}{}.
\newblock
\APACrefbtitle {Guidelines f\"ur das {T}agging deutscher {T}extcorpora mit
  {STTS} ({K}leines und gro\ss es {T}agset).} {Guidelines f\"ur das {T}agging
  deutscher {T}extcorpora mit {STTS} ({K}leines und gro\ss es {T}agset).}
\newblock
\APAChowpublished {\url{www.sfs.uni-tuebingen.de/resources/stts-1999.pdf}}.
\PrintBackRefs{\CurrentBib}

\bibitem [\protect \citeauthoryear {%
Siegelman%
\ \protect \BOthers {.}}{%
Siegelman%
\ \protect \BOthers {.}}{%
{\protect \APACyear {2022}}%
}]{%
Siegelman2022}
\APACinsertmetastar {%
Siegelman2022}%
\begin{APACrefauthors}%
Siegelman, N.%
, Schroeder, S.%
, Acart\"{u}rk, C.%
, Ahn, H\BHBI D.%
, Alexeeva, S.%
, Amenta, S.%
\BDBL {}Kuperman, V.%
\end{APACrefauthors}%
\unskip\
\newblock
\APACrefYearMonthDay{2022}{}{}.
\newblock
{\BBOQ}\APACrefatitle {Expanding horizons of cross-linguistic research on
  reading: {The Multilingual Eye-movement Corpus} ({MECO})} {Expanding horizons
  of cross-linguistic research on reading: {The Multilingual Eye-movement
  Corpus} ({MECO})}.{\BBCQ}
\newblock
\APACjournalVolNumPages{Behavior Research Methods}{54}{6}{2843--2863,}
\newblock
\begin{APACrefDOI} \doi{10.3758/s13428-021-01772-6} \end{APACrefDOI}
\newblock

\newblock

\PrintBackRefs{\CurrentBib}

\bibitem [\protect \citeauthoryear {%
Sood%
, Tannert%
, Frassinelli%
, Bulling%
\BCBL {}\ \BBA {} Vu%
}{%
Sood%
\ \protect \BOthers {.}}{%
{\protect \APACyear {2020}}%
}]{%
sood2020interpreting}
\APACinsertmetastar {%
sood2020interpreting}%
\begin{APACrefauthors}%
Sood, E.%
, Tannert, S.%
, Frassinelli, D.%
, Bulling, A.%
\BCBL {} Vu, N.T.%
\end{APACrefauthors}%
\unskip\
\newblock
\APACrefYearMonthDay{2020}{}{}.
\newblock
{\BBOQ}\APACrefatitle {Interpreting Attention Models with Human Visual
  Attention in Machine Reading Comprehension} {Interpreting attention models
  with human visual attention in machine reading comprehension}.{\BBCQ}
\newblock
 \APACrefbtitle {Proceedings of the 24th Conference on Computational Natural
  Language Learning} {Proceedings of the 24th conference on computational
  natural language learning}\ (\BPGS\ 12--25).
\PrintBackRefs{\CurrentBib}

\bibitem [\protect \citeauthoryear {%
{SR Research Ltd}%
}{%
{SR Research Ltd}%
}{%
{\protect \APACyear {2010}}%
}]{%
eyelink-install-guide}
\APACinsertmetastar {%
eyelink-install-guide}%
\begin{APACrefauthors}%
{SR Research Ltd}%
\end{APACrefauthors}%
\unskip\
\newblock
\APACrefYearMonthDay{2010}{}{}.
\newblock
{\BBOQ}\APACrefatitle {{EyeLink} 1000 {Installation Guide}} {{EyeLink} 1000
  {Installation Guide}}{\BBCQ}\ [\bibcomputersoftwaremanual].
\newblock
\APACaddressPublisher{Mississauga, Canada}{}.
\newblock
\begin{APACrefURL}
  {\url{http://sr-research.jp/support/files/EyeLinkCLsetupguide.pdf}}
  \end{APACrefURL}
\newblock
\APACrefnote{Document version 1.5.2}
\PrintBackRefs{\CurrentBib}

\bibitem [\protect \citeauthoryear {%
{SR Research Ltd.}%
}{%
{SR Research Ltd.}%
}{%
{\protect \APACyear {2011}}%
}]{%
dataviewer}
\APACinsertmetastar {%
dataviewer}%
\begin{APACrefauthors}%
{SR Research Ltd.}%
\end{APACrefauthors}%
\unskip\
\newblock
\APACrefYearMonthDay{2011}{}{}.
\newblock
{\BBOQ}\APACrefatitle {Eye{Link} {D}ata {V}iewer user's manual} {Eye{Link}
  {D}ata {V}iewer user's manual}{\BBCQ}\ [\bibcomputersoftwaremanual].
\newblock
\APACaddressPublisher{Mississauga, Canada}{}.
\newblock
\APACrefnote{Document version 1.11.1}
\PrintBackRefs{\CurrentBib}

\bibitem [\protect \citeauthoryear {%
Sui%
, Dirix%
, Woumans%
\BCBL {}\ \BBA {} Duyck%
}{%
Sui%
\ \protect \BOthers {.}}{%
{\protect \APACyear {2022}}%
}]{%
Sui2022gecocn}
\APACinsertmetastar {%
Sui2022gecocn}%
\begin{APACrefauthors}%
Sui, L.%
, Dirix, N.%
, Woumans, E.%
\BCBL {} Duyck, W.%
\end{APACrefauthors}%
\unskip\
\newblock
\APACrefYearMonthDay{2022}{}{}.
\newblock
{\BBOQ}\APACrefatitle {{GECO-CN}: Ghent Eye-tracking {COrpus} of sentence
  reading for {Chinese-English} bilinguals} {{GECO-CN}: Ghent eye-tracking
  {COrpus} of sentence reading for {Chinese-English} bilinguals}.{\BBCQ}
\newblock
\APACjournalVolNumPages{Behavior Research Methods}{55}{6}{2743–2763,}
\newblock
\begin{APACrefDOI} \doi{10.3758/s13428-022-01931-3} \end{APACrefDOI}
\newblock

\newblock

\PrintBackRefs{\CurrentBib}

\bibitem [\protect \citeauthoryear {%
Takmaz%
, Pezzelle%
, Beinborn%
\BCBL {}\ \BBA {} Fern{\'a}ndez%
}{%
Takmaz%
\ \protect \BOthers {.}}{%
{\protect \APACyear {2020}}%
}]{%
takmaz-etal-2020-img-caption}
\APACinsertmetastar {%
takmaz-etal-2020-img-caption}%
\begin{APACrefauthors}%
Takmaz, E.%
, Pezzelle, S.%
, Beinborn, L.%
\BCBL {} Fern{\'a}ndez, R.%
\end{APACrefauthors}%
\unskip\
\newblock
\APACrefYearMonthDay{2020}{}{}.
\newblock
{\BBOQ}\APACrefatitle {Generating Image Descriptions via Sequential Cross-Modal
  Alignment Guided by Human Gaze} {Generating image descriptions via sequential
  cross-modal alignment guided by human gaze}.{\BBCQ}
\newblock
 \APACrefbtitle {{Proceedings of the 2020 Conference on Empirical Methods in
  Natural Language Processing (EMNLP)}} {{Proceedings of the 2020 Conference on
  Empirical Methods in Natural Language Processing (EMNLP)}}\ (\BPGS\
  4664--4677).
\newblock
\APACaddressPublisher{}{Association for Computational Linguistics}.
\PrintBackRefs{\CurrentBib}

\bibitem [\protect \citeauthoryear {%
{TIGER Project}%
}{%
{TIGER Project}%
}{%
{\protect \APACyear {2003}}%
}]{%
tiger2003schema}
\APACinsertmetastar {%
tiger2003schema}%
\begin{APACrefauthors}%
{TIGER Project}%
\end{APACrefauthors}%
\unskip\
\newblock
\APACrefYearMonthDay{2003}{}{}.
\newblock
{\BBOQ}\APACrefatitle {{TIGER} Annotationsschema} {{TIGER}
  annotationsschema}{\BBCQ}\ [\bibcomputersoftwaremanual].
\newblock
\APACaddressPublisher{}{Universität des Saarlands, Universität Stuttgart,
  Universität Potsdam}.
\PrintBackRefs{\CurrentBib}

\bibitem [\protect \citeauthoryear {%
Touvron%
\ \protect \BOthers {.}}{%
Touvron%
\ \protect \BOthers {.}}{%
{\protect \APACyear {2023}}%
}]{%
touvron2023llama}
\APACinsertmetastar {%
touvron2023llama}%
\begin{APACrefauthors}%
Touvron, H.%
, Martin, L.%
, Stone, K.%
, Albert, P.%
, Almahairi, A.%
, Babaei, Y.%
\BDBL {}Scialom, T.%
\end{APACrefauthors}%
\unskip\
\newblock
\APACrefYearMonthDay{2023}{}{}.
\newblock
\APACrefbtitle {Llama 2: Open Foundation and Fine-Tuned Chat Models.} {Llama 2:
  Open foundation and fine-tuned chat models.}
\newblock
\APACrefnote{arXiv preprint arXiv:2307.09288}
\PrintBackRefs{\CurrentBib}

\bibitem [\protect \citeauthoryear {%
Townsend%
, Begon%
\BCBL {}\ \BBA {} Harper%
}{%
Townsend%
\ \protect \BOthers {.}}{%
{\protect \APACyear {2003}}%
}]{%
Oekologie}
\APACinsertmetastar {%
Oekologie}%
\begin{APACrefauthors}%
Townsend, C.R.%
, Begon, M.%
\BCBL {} Harper, J.L.%
\end{APACrefauthors}%
\unskip\
\newblock
\APACrefYear{2003}.
\newblock
\APACrefbtitle {{\"Okologie}} {{\"Okologie}}\ (J.~Steidle, F.~Thomas,
  B.~Stadler, U.~Hoffmeister\BCBL {}\ \BBA {} T.~Hoffmeister, \BTRANSS{}).
\newblock
\APACaddressPublisher{Berlin}{Springer}.
\PrintBackRefs{\CurrentBib}

\bibitem [\protect \citeauthoryear {%
Vieira%
}{%
Vieira%
}{%
{\protect \APACyear {2020}}%
}]{%
vieira2020rastro}
\APACinsertmetastar {%
vieira2020rastro}%
\begin{APACrefauthors}%
Vieira, J.M.M.%
\end{APACrefauthors}%
\unskip\
\newblock
\APACrefYear{2020}.
\unskip\
\newblock
\APACrefbtitle {The {B}razilian {P}ortuguese eye tracking corpus with a
  predictability study focusing on lexical and partial prediction} {The
  {B}razilian {P}ortuguese eye tracking corpus with a predictability study
  focusing on lexical and partial prediction}\ \APACtypeAddressSchool
  {\BPhD}{}{Universidade Federal do Cear\'{a}}.
\unskip\
\newblock
\begin{APACrefURL} {\url{http://repositorio.ufc.br/handle/riufc/55798}}
  \end{APACrefURL}
\PrintBackRefs{\CurrentBib}

\bibitem [\protect \citeauthoryear {%
E.~Wilcox%
, Meister%
, Cotterell%
\BCBL {}\ \BBA {} Pimentel%
}{%
E.~Wilcox%
\ \protect \BOthers {.}}{%
{\protect \APACyear {2023}}%
}]{%
wilcox-etal-2023-language}
\APACinsertmetastar {%
wilcox-etal-2023-language}%
\begin{APACrefauthors}%
Wilcox, E.%
, Meister, C.%
, Cotterell, R.%
\BCBL {} Pimentel, T.%
\end{APACrefauthors}%
\unskip\
\newblock
\APACrefYearMonthDay{2023}{}{}.
\newblock
{\BBOQ}\APACrefatitle {Language Model Quality Correlates with Psychometric
  Predictive Power in Multiple Languages} {Language model quality correlates
  with psychometric predictive power in multiple languages}.{\BBCQ}
\newblock
 \APACrefbtitle {{Proceedings of the 2023 Conference on Empirical Methods in
  Natural Language Processing}} {{Proceedings of the 2023 Conference on
  Empirical Methods in Natural Language Processing}}\ (\BPGS\ 7503--7511).
\newblock
\APACaddressPublisher{}{Association for Computational Linguistics}.
\PrintBackRefs{\CurrentBib}

\bibitem [\protect \citeauthoryear {%
E.G.~Wilcox%
, Gauthier%
, Hu%
, Qian%
\BCBL {}\ \BBA {} Levy%
}{%
E.G.~Wilcox%
\ \protect \BOthers {.}}{%
{\protect \APACyear {2020}}%
}]{%
wilcox-etal:2020-on-the-predictive-power}
\APACinsertmetastar {%
wilcox-etal:2020-on-the-predictive-power}%
\begin{APACrefauthors}%
Wilcox, E.G.%
, Gauthier, J.%
, Hu, J.%
, Qian, P.%
\BCBL {} Levy, R.P.%
\end{APACrefauthors}%
\unskip\
\newblock
\APACrefYearMonthDay{2020}{}{}.
\newblock
{\BBOQ}\APACrefatitle {On the Predictive Power of Neural Language Models for
  Human Real-Time Comprehension Behavior} {On the predictive power of neural
  language models for human real-time comprehension behavior}.{\BBCQ}
\newblock
 \APACrefbtitle {{Proceedings of the 42nd Annual Meeting of the Cognitive
  Science Society}} {{Proceedings of the 42nd Annual Meeting of the Cognitive
  Science Society}}\ (\BPG~1707–1713).
\PrintBackRefs{\CurrentBib}

\bibitem [\protect \citeauthoryear {%
Wolfer%
\ \protect \BOthers {.}}{%
Wolfer%
\ \protect \BOthers {.}}{%
{\protect \APACyear {2013}}%
}]{%
popsci}
\APACinsertmetastar {%
popsci}%
\begin{APACrefauthors}%
Wolfer, S.%
, Müller-Feldmeth, D.%
, Konieczny, L.%
, Held, U.%
, Maksymski, K.%
, Hansen-Schirra, S.%
\BDBL {}Auer, P.%
\end{APACrefauthors}%
\unskip\
\newblock
\APACrefYearMonthDay{2013}{}{}.
\newblock
{\BBOQ}\APACrefatitle {{PopSci}: A reading corpus of popular science texts with
  rich multi-level annotations. {A} case study} {{PopSci}: A reading corpus of
  popular science texts with rich multi-level annotations. {A} case
  study}.{\BBCQ}
\newblock
 \APACrefbtitle {{Book of Abstracts of the 17th European Conference on Eye
  Movements.}} {{Book of Abstracts of the 17th European Conference on Eye
  Movements.}}
\PrintBackRefs{\CurrentBib}

\bibitem [\protect \citeauthoryear {%
{World Medical Association}%
}{%
{World Medical Association}%
}{%
{\protect \APACyear {2013}}%
}]{%
helsinki-declaration2013}
\APACinsertmetastar {%
helsinki-declaration2013}%
\begin{APACrefauthors}%
{World Medical Association}%
\end{APACrefauthors}%
\unskip\
\newblock
\APACrefYearMonthDay{2013}{}{}.
\newblock
\APACrefbtitle {{World Medical Association Declaration of Helsinki}: {E}thical
  principles for medical research involving human subjects.} {{World Medical
  Association Declaration of Helsinki}: {E}thical principles for medical
  research involving human subjects.}
\newblock
\begin{APACrefURL}
  {\url{https://www.wma.net/policies-post/wma-declaration-of-helsinki-ethical-principles-for-medical-research-involving-human-subjects/}}
  \end{APACrefURL}
\PrintBackRefs{\CurrentBib}

\bibitem [\protect \citeauthoryear {%
Wu%
\ \BBA {} Kit%
}{%
Wu%
\ \BBA {} Kit%
}{%
{\protect \APACyear {2023}}%
}]{%
Wu2023HKC}
\APACinsertmetastar {%
Wu2023HKC}%
\begin{APACrefauthors}%
Wu, Y.%
\BCBT {}\ \BBA {} Kit, C.%
\end{APACrefauthors}%
\unskip\
\newblock
\APACrefYearMonthDay{2023}{}{}.
\newblock
{\BBOQ}\APACrefatitle {{Hong Kong }Corpus of {Chinese} Sentence and Passage
  Reading} {{Hong Kong }corpus of {Chinese} sentence and passage
  reading}.{\BBCQ}
\newblock
\APACjournalVolNumPages{Scientific Data}{10}{1}{1–13,}
\newblock
\begin{APACrefDOI} \doi{10.1038/s41597-023-02813-9} \end{APACrefDOI}
\newblock

\newblock

\PrintBackRefs{\CurrentBib}

\bibitem [\protect \citeauthoryear {%
Yaneva%
}{%
Yaneva%
}{%
{\protect \APACyear {2016}}%
}]{%
yaneve2016asd}
\APACinsertmetastar {%
yaneve2016asd}%
\begin{APACrefauthors}%
Yaneva, V.%
\end{APACrefauthors}%
\unskip\
\newblock
\APACrefYear{2016}.
\unskip\
\newblock
\APACrefbtitle {{Assessing Text and Web Accessibility for People with Autism
  Spectrum Disorder}} {{Assessing Text and Web Accessibility for People with
  Autism Spectrum Disorder}}\ \APACtypeAddressSchool {\BUPhD}{}{}.
\unskip\
\newblock
\APACaddressSchool {}{University of Wolverhampton}.
\PrintBackRefs{\CurrentBib}

\bibitem [\protect \citeauthoryear {%
Yang%
\ \BBA {} Hollenstein%
}{%
Yang%
\ \BBA {} Hollenstein%
}{%
{\protect \APACyear {2023}}%
}]{%
yang2023plm}
\APACinsertmetastar {%
yang2023plm}%
\begin{APACrefauthors}%
Yang, D.%
\BCBT {}\ \BBA {} Hollenstein, N.%
\end{APACrefauthors}%
\unskip\
\newblock
\APACrefYearMonthDay{2023}{}{}.
\newblock
{\BBOQ}\APACrefatitle {{PLM-AS}: Pre-trained Language Models Augmented with
  Scanpaths for Sentiment Classification} {{PLM-AS}: Pre-trained language
  models augmented with scanpaths for sentiment classification}.{\BBCQ}
\newblock
 \APACrefbtitle {{Proceedings of the Northern Lights Deep Learning Workshop}}
  {{Proceedings of the Northern Lights Deep Learning Workshop}}\ (\BVOL~4).
\PrintBackRefs{\CurrentBib}

\bibitem [\protect \citeauthoryear {%
Yi%
, Guo%
, Jiang%
, Wang%
\BCBL {}\ \BBA {} Sun%
}{%
Yi%
\ \protect \BOthers {.}}{%
{\protect \APACyear {2020}}%
}]{%
yi2020gaze-sum}
\APACinsertmetastar {%
yi2020gaze-sum}%
\begin{APACrefauthors}%
Yi, K.%
, Guo, Y.%
, Jiang, W.%
, Wang, Z.%
\BCBL {} Sun, L.%
\end{APACrefauthors}%
\unskip\
\newblock
\APACrefYearMonthDay{2020}{}{}.
\newblock
{\BBOQ}\APACrefatitle {A Dataset for Exploring Gaze Behaviors in Text
  Summarization} {A dataset for exploring gaze behaviors in text
  summarization}.{\BBCQ}
\newblock
 \APACrefbtitle {{Proceedings of the 11th ACM Multimedia Systems Conference}}
  {{Proceedings of the 11th ACM Multimedia Systems Conference}}\
  (\BPG~243–248).
\newblock
\APACaddressPublisher{}{Association for Computing Machinery}.
\PrintBackRefs{\CurrentBib}

\bibitem [\protect \citeauthoryear {%
Zang%
, Fu%
, Bai%
, Yan%
\BCBL {}\ \BBA {} Liversedge%
}{%
Zang%
\ \protect \BOthers {.}}{%
{\protect \APACyear {2018}}%
}]{%
Zang2018zh-word-len}
\APACinsertmetastar {%
Zang2018zh-word-len}%
\begin{APACrefauthors}%
Zang, C.%
, Fu, Y.%
, Bai, X.%
, Yan, G.%
\BCBL {} Liversedge, S.P.%
\end{APACrefauthors}%
\unskip\
\newblock
\APACrefYearMonthDay{2018}{}{}.
\newblock
{\BBOQ}\APACrefatitle {Investigating word length effects in {Chinese} reading.}
  {Investigating word length effects in {Chinese} reading.}{\BBCQ}
\newblock
\APACjournalVolNumPages{Journal of Experimental Psychology: Human Perception
  and Performance}{44}{12}{1831–1841,}
\newblock
\begin{APACrefDOI} \doi{10.1037/xhp0000589} \end{APACrefDOI}
\newblock

\newblock

\PrintBackRefs{\CurrentBib}

\bibitem [\protect \citeauthoryear {%
Zhang%
\ \protect \BOthers {.}}{%
Zhang%
\ \protect \BOthers {.}}{%
{\protect \APACyear {2022}}%
}]{%
Zhang2022chinese}
\APACinsertmetastar {%
Zhang2022chinese}%
\begin{APACrefauthors}%
Zhang, G.%
, Yao, P.%
, Ma, G.%
, Wang, J.%
, Zhou, J.%
, Huang, L.%
\BDBL {}Li, X.%
\end{APACrefauthors}%
\unskip\
\newblock
\APACrefYearMonthDay{2022}{}{}.
\newblock
{\BBOQ}\APACrefatitle {The database of eye-movement measures on words in
  {Chinese} reading} {The database of eye-movement measures on words in
  {Chinese} reading}.{\BBCQ}
\newblock
\APACjournalVolNumPages{Scientific Data}{9}{1}{1-8,}
\newblock
\begin{APACrefDOI} \doi{10.1038/s41597-022-01464-6} \end{APACrefDOI}
\newblock

\newblock

\PrintBackRefs{\CurrentBib}

\end{thebibliography}


\end{document}